\documentclass[10pt,twocolumn,letterpaper]{article}

\ifdefined\BUILDMODE\else\def\BUILDMODE{arxiv}\fi

\newif\ifDRAFT
\newif\ifSUBMISSION
\newif\ifCAMREADY
\newif\ifARXIV
\newif\ifSUPPLEMENTAL

\makeatletter
\def\@bm@dra{draft}\def\@bm@sub{submission}
\def\@bm@cam{camready}\def\@bm@arx{arxiv}\def\@bm@spl{supplemental}
\ifx\BUILDMODE\@bm@dra \DRAFTtrue       \fi
\ifx\BUILDMODE\@bm@sub \SUBMISSIONtrue  \fi
\ifx\BUILDMODE\@bm@cam \CAMREADYtrue    \fi
\ifx\BUILDMODE\@bm@arx \ARXIVtrue       \fi
\ifx\BUILDMODE\@bm@spl \SUPPLEMENTALtrue\fi
\makeatother

\newif\ifSHOWNOTES
\newif\ifSHOWURL
\newif\ifSHOWSUPP
\newif\ifNONANON

\ifDRAFT
  \SHOWNOTEStrue
\fi
\ifCAMREADY
  \NONANONtrue
  \SHOWURLtrue
\fi
\ifARXIV
  \NONANONtrue
  \SHOWURLtrue
  \SHOWSUPPtrue
\fi

\ifARXIV
  \usepackage[pagenumbers]{cvpr}
\else
  \ifCAMREADY
    \usepackage{cvpr}
  \else
    \usepackage[review]{cvpr}
  \fi
\fi
\def\confName{CVPR}
\def\confYear{2027}

\ifdefined\PAPERVARIANT\else\def\PAPERVARIANT{extended}\fi
\newif\ifEXTENDED
\makeatletter
\def\@pv@ext{extended}
\ifx\PAPERVARIANT\@pv@ext \EXTENDEDtrue \fi
\makeatother

\makeatletter
\@namedef{ver@everyshi.sty}{2001/05/15}
\renewcommand\LenToUnit[1]{\strip@pt\dimexpr#1\relax}
\makeatother

\usepackage{graphicx}
\usepackage{booktabs}
\usepackage{pifont}
\usepackage{wrapfig}
\usepackage{placeins}
\usepackage{amsmath}
\usepackage{amssymb}
\usepackage{mathtools}
\usepackage{xcolor}
\usepackage{colortbl}
\usepackage{tabularx}
\usepackage{algorithm}
\usepackage{algpseudocode}
\floatstyle{plain}
\newfloat{promptbox}{tbp}{lop}
\usepackage{tikz}
\usetikzlibrary{positioning, arrows.meta, fit, calc, backgrounds, shapes.geometric}

\usepackage{microtype}

\usepackage{siunitx}
\usepackage{subcaption}

\patchcmd{\abstract}{\vspace*{12pt}}{\vspace*{4pt}}{}{%
  \PackageError{dash2sim}{Could not adjust abstract spacing}{Check the CVPR abstract definition.}}

\usepackage{comment}
\ifSHOWSUPP
  \includecomment{supponly}
\else
  \excludecomment{supponly}
\fi

\ifEXTENDED
  \includecomment{webonly}\excludecomment{confonly}
\else
  \excludecomment{webonly}\includecomment{confonly}
\fi

\usepackage{atbegshi}
\newif\ifDISCARDPAGES
\AtBeginShipout{\ifDISCARDPAGES\AtBeginShipoutDiscard\fi}
\newcounter{bodylastpage}
\makeatletter
\let\deferredprotectedwrite\protected@write
\newcommand\deferredwrites{\let\protected@write\deferredprotectedwrite}
\newcommand\immediatewrites{%
  \long\def\protected@write##1##2##3{%
    \begingroup
      ##2\let\protect\@unexpandable@protect
      \edef\reserved@a{\immediate\write##1{##3}}\reserved@a
    \endgroup}}
\makeatother

\definecolor{cvprblue}{rgb}{0.21,0.49,0.74}
\usepackage[pagebackref,breaklinks,colorlinks,allcolors=cvprblue]{hyperref}

\definecolor{textgreen}{rgb}{0.4980392156862745, 0.788235294117647, 0.4980392156862745}
\definecolor{ourmediumblue}{rgb}{0.21568627450980393,0.49411764705882355,0.7215686274509804}

\definecolor{darkspringgreen}{rgb}{0.09, 0.45, 0.27}
\definecolor{cornellred}{rgb}{0.7, 0.11, 0.11}
\definecolor{darkcerulean}{rgb}{0.03, 0.27, 0.49}
\definecolor{amaranth}{rgb}{0.9, 0.17, 0.31}
\definecolor{flame}{rgb}{0.89, 0.35, 0.13}

\newcommand{\cmark}{\ding{51}}
\newcommand{\xmark}{\ding{55}}
\newcommand{\gcheck}{{\color{green!60!black}\cmark}}
\newcommand{\rcross}{{\color{red!75!black}\xmark}}

\newcommand{\txtdeepblue}[1]{\textcolor{darkcerulean}{#1}}

\newcommand{\projectnamecolor}{\textbf{\txtdeepblue{Dash}}2Sim}

\newcommand{\projectname}{Dash2Sim}
\newcommand{\benchmarkname}{ROADWork4D}
\newcommand{\clbenchmarkname}{ROADWork4D-CL}

\newcommand{\eebenchmarkname}{ROADWork4D-E2E}

\newcommand{\confweb}[2]{\ifEXTENDED #2\else #1\fi}

\newcommand{\tablesize}{\confweb{\small}{\normalsize}}

\newcommand{\secref}[1]{\S\ref{sec:#1}}
\newcommand{\figref}[1]{Fig.~\ref{fig:#1}}
\newcommand{\Figref}[1]{Figure~\ref{fig:#1}}
\newcommand{\tabref}[1]{Tab.~\ref{tab:#1}}
\newcommand{\Tabref}[1]{Table~\ref{tab:#1}}

\newcommand{\Algoref}[1]{Algorithm~\ref{alg:#1}}

\newcommand{\suppnumber}[1]{\ifSUBMISSION\ref*{#1}\else\ref{#1}\fi}
\newcommand{\appref}[1]{\suppref{app:#1}}
\newcommand{\suppref}[1]{\ifEXTENDED\S\else Supp.~Sec.~\fi\suppnumber{#1}}
\newcommand{\supptabref}[1]{\ifEXTENDED\else Supp.~\fi Tab.~\suppnumber{#1}}

\newcommand{\suppfigref}[1]{\ifEXTENDED\else Supp.~\fi Fig.~\suppnumber{#1}}

\urldef{\projecturl}\url{https://www.cs.cmu.edu/~dash2sim/}

\definecolor{tablehighlight}{HTML}{E8F0F5}
\definecolor{tablehighlightgreen}{HTML}{E8F3E8}
\definecolor{tablebest}{HTML}{A9DBB6}
\definecolor{tablesecond}{HTML}{D4E8B9}
\definecolor{tableworst}{HTML}{FDD2D3}
\newcommand{\bestcell}[1]{\cellcolor{tablebest}#1}
\newcommand{\secondcell}[1]{\cellcolor{tablesecond}#1}
\newcommand{\worstcell}[1]{\cellcolor{tableworst}#1}

\newif\ifstatsverified
\statsverifiedtrue            

\newcommand{\statFullScenarios}{4{,}244}
\newcommand{\statFullCities}{17}
\newcommand{\statFullFrames}{633{,}404}
\newcommand{\statFullHoursNum}{35.2}
\newcommand{\statFullHours}{\statFullHoursNum\,h}
\newcommand{\statFullSfMPoints}{276.3\,M}
\newcommand{\statFullTracks}{142{,}863}
\newcommand{\statFullDyn}{65.0k}
\newcommand{\statFullStat}{77.9k}
\newcommand{\statFullBoxes}{2.74\,M}
\newcommand{\statFullKm}{724.5\,km}

\newcommand{\statVerScenarios}{2{,}201}

\title{\projectnamecolor: Driving \txtdeepblue{Simulation} from in-the-wild \txtdeepblue{Dashcam} Videos in Work Zones}

\author{
  Anurag Ghosh\textsuperscript{1}\quad
  Francesco Pittaluga\textsuperscript{2}\quad
  Khiem Vuong\textsuperscript{1}\quad
  Angela Chen\textsuperscript{1}\\
  Juan Alvarez-Padilla\textsuperscript{3}\quad
  Manmohan Chandraker\textsuperscript{2,4}\quad
  Srinivasa Narasimhan\textsuperscript{1}\normalfont\\[4pt]
  \small\textsuperscript{1}Carnegie Mellon University\quad
  \textsuperscript{2}NEC Labs America\\
  \small
  \textsuperscript{3}MIT\thanks{Work done at Carnegie Mellon University}\quad
  \textsuperscript{4}UC San Diego%
  \ifSHOWURL\\[6pt]\projecturl\fi
}

\hypersetup{
  pdftitle={\projectname: Driving Simulation from in-the-wild Dashcam Videos in Work Zones},
  pdfkeywords={autonomous driving, end-to-end planning, closed-loop planning, dashcam, long-tail planning, data-driven simulation},
}
\ifNONANON
  \hypersetup{pdfauthor={Anurag Ghosh, Francesco Pittaluga, Khiem Vuong, Angela Chen, Juan Rodolfo Alvarez Padilla, Manmohan Chandraker, Srinivasa Narasimhan}}
\else
  \hypersetup{pdfauthor={Anonymous}}
\fi

\begin{document}

\ifSUPPLEMENTAL
  \nolinenumbers
  \immediatewrites
  \DISCARDPAGEStrue
\fi

\makeatletter
\g@addto@macro\@maketitle{
\noindent\begin{minipage}{\textwidth}
    \centering
    \captionsetup{type=figure,position=bottom,skip=4pt}
    \includegraphics[width=0.97\linewidth]{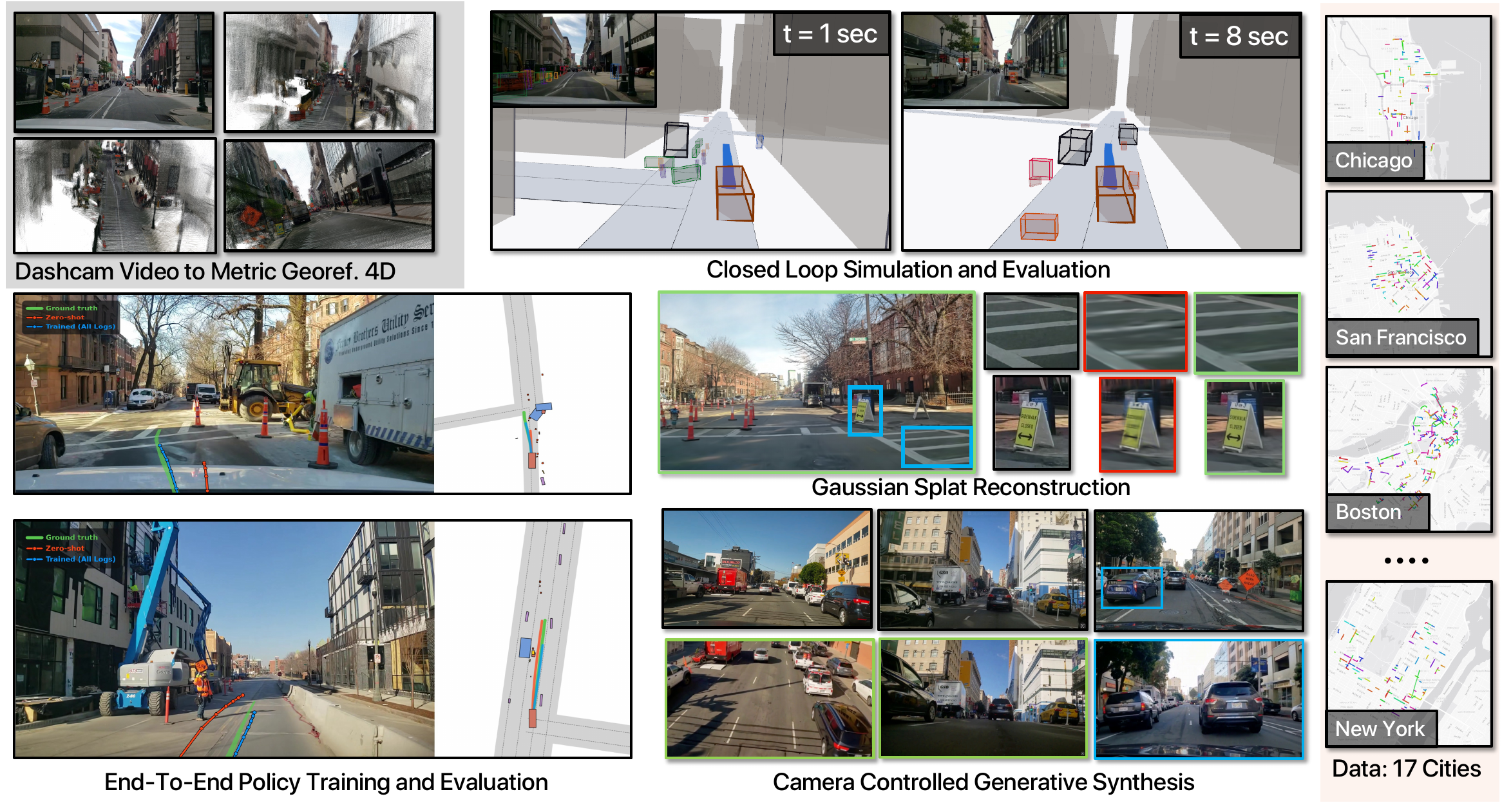}\par
\caption{{\textbf{Top:} \projectname{} recovers metric, geo-referenced 4D work-zone scenes from in-the-wild monocular dashcam videos. The recovered scenes are suitable for training and evaluating driving policies in privileged closed-loop simulation.
\textbf{Middle and bottom left:} Fine-tuning on recovered logs brings end-to-end trajectory predictions (blue) closer to recovered reference trajectories (green) than zero-shot predictions (orange) are.
\textbf{Middle right:} Recovered depth improves Gaussian splat rendering of road markings and temporary signs. Crops show the reference image (black) and OmniRe renderings without (red) and with (green) our depth supervision.
\textbf{Bottom right:} Recovered geometry conditions generative synthesis along \textcolor[HTML]{92D050}{modified camera trajectories} and from the \textcolor[HTML]{00B0F0}{highlighted vehicle}'s perspective, with source views above generated views.
\textbf{Far right:} Recovered routes show \benchmarkname{}'s coverage across \statFullCities{} US cities.}}
\label{fig:teaser}
\end{minipage}\par
\vspace{10pt}
}
\raggedbottom
\AtBeginShipoutNext{\global\let\@textbottom\relax}
\makeatother
\maketitle

\begin{abstract}
Self-driving simulations rely on fleet data collected in a few cities or on hand-crafted synthetic scenes. Dashcam videos cover a far broader range of locations and situations, including long-tailed scenarios such as work zones. \projectname{} constructs metric, geo-referenced 4D logs from monocular dashcam videos, pairing observations with trajectories, depth, and tracked objects. We introduce \benchmarkname{}, a benchmark for autonomous driving in work zones across \statFullCities{} US cities, built by augmenting ROADWork dashcam videos with recovered 4D logs. \benchmarkname{} provides \statFullScenarios{} scenes for end-to-end planning and \statVerScenarios{} scenes for privileged closed-loop planning, supporting both training and evaluation. On closed-loop planning, rule-based and hybrid planners outperform learned planners, while temporary work-zone channels remain challenging across planner families. For end-to-end planning, fine-tuning addresses the primary work-zone challenge of lateral placement, reducing lateral trajectory error by 65\%. Recovered depth improves novel-view synthesis, including a 19\% reduction in CLIP-FID, providing supervision for sensor simulation.
\end{abstract}

\section{Introduction}
\label{sec:intro}

\begin{table*}[t]
\centering
\small
\setlength{\tabcolsep}{1pt}
\renewcommand{\arraystretch}{1.12}
\caption{\textbf{Comparison with existing driving datasets and benchmarks.} We compare data sources, scale, scene representations, and support for End-to-End and Closed-Loop planning. Argoverse 2 counts refer to its sensor collection.}
\vspace{-1em}
\label{tab:benchmarks}
\begin{tabularx}{\textwidth}{@{}l l >{\raggedright\arraybackslash\leavevmode}X c c r r c c c@{}}
\toprule
& & & & & \multicolumn{2}{c}{\textbf{Scale}} & \multicolumn{3}{c}{\textbf{Planning Capabilities}} \\
\cmidrule(lr){6-7} \cmidrule(lr){8-10}
\textbf{Benchmark} & \textbf{Source} & \textbf{Available Data} & \begin{tabular}[t]{@{}c@{}}\textbf{Long}\\\textbf{Tail}\end{tabular} & \textbf{Maps} & \textbf{Hours} & \textbf{Scenes} & \begin{tabular}[t]{@{}c@{}}\textbf{End-to-}\\\textbf{End}\end{tabular} & \begin{tabular}[t]{@{}c@{}}\textbf{Closed}\\\textbf{Loop}\end{tabular} & \begin{tabular}[t]{@{}c@{}}\textbf{Reactive}\\\textbf{agents}\end{tabular} \\
\midrule
KITTI~\cite{kitti} & Fleet & Cameras, LiDAR, Tracks & \rcross & \rcross & 6 & --- & \gcheck & \rcross & \rcross \\
nuScenes~\cite{nuscenes} & Fleet & Cameras, LiDAR, Radar, Maps, Tracks & \rcross & \gcheck & 5.6 & 1,000 & \gcheck & \rcross & \rcross \\
Argoverse 2~\cite{argoverse2} & Fleet & Cameras, LiDAR, Maps, Tracks & \rcross & \gcheck & 4.2 & 1,000 & \gcheck & \rcross & \rcross \\
WOMD~\cite{waymo_motion,waymax} & Fleet & LiDAR, Embeddings, Maps, Tracks & \rcross & \gcheck & 574 & 100k & \rcross & Privileged & \gcheck \\
NAVSIM v2~\cite{navsim,cao2025pseudo} & Fleet & Cameras, LiDAR, Maps, Tracks & \rcross & \gcheck & --- & 104k & \gcheck & Pseudo-sim & \gcheck \\
nuPlan~\cite{nuplan} & Fleet & Cameras, LiDAR, Maps, Tracks & \rcross & \gcheck & 1,282 & --- & \gcheck & Privileged & \gcheck \\
WOD-E2E~\cite{wod_e2e} & Fleet & Cameras & \gcheck & \rcross & 12 & 4,021 & \gcheck & \rcross & \rcross \\
KITScenes-LT~\cite{wagner2026longtail} & Fleet & Cameras & \gcheck & \rcross & 2.5 & 1,000 & \gcheck & \rcross & \rcross \\
\midrule
CARLA~\cite{carla} & Synthetic & Cameras, LiDAR, Maps, Tracks & \rcross & \gcheck & --- & --- & \gcheck & Rendered sensors & \gcheck \\
Bench2Drive~\cite{bench2drive} & Synthetic & Cameras, LiDAR, Maps, Tracks & \gcheck & \gcheck & --- & 220 & \gcheck & Rendered sensors & \gcheck \\
Fail2Drive~\cite{fail2drive} & Synthetic & Cameras, LiDAR, Maps, Tracks & \gcheck & \gcheck & --- & 200 & \gcheck & Rendered sensors & \gcheck \\
InterPlan~\cite{interplan} & Fleet-Synth. & Maps, Tracks & \gcheck & \gcheck & --- & 335 & \rcross & Privileged & \gcheck \\
\midrule
\rowcolor{tablehighlightgreen}
\benchmarkname{} & Dashcam & Camera, Depth, Maps, Tracks & \gcheck & \gcheck & \statFullHoursNum & \statFullScenarios{} & \gcheck & Privileged & \gcheck \\
\bottomrule
\end{tabularx}
\vspace{-1em}
\end{table*}

Dashboard cameras have become standard equipment on consumer vehicles~\cite{autoinsurance2026dashcams}. The footage they collectively produce, even just the fraction uploaded to public video platforms, spans more cities and unusual situations than any publicly available autonomous vehicle dataset. However, exploiting this data for learning how to drive autonomously is underexplored. Work zones specifically are a persistent failure mode for commercial driving systems in 2026~\cite{sfgate2023cruiseconcrete, techcrunch2026waymoworkzone, cnbc2026waymorecall}. Instrumented fleets are expensive to scale, making it difficult to collect rare driving situations such as work zones in sufficient numbers (\tabref{benchmarks}). \noindent\textit{This paper asks: Can ordinary dashcam footage give us useful training and evaluation data for driving situations that existing datasets cover poorly?}

\noindent We introduce \benchmarkname{}, a benchmark for autonomous driving in work zones with \statFullScenarios{} scenes across \statFullCities{} US cities. Our system, \projectname{}, constructs the benchmark by recovering physically grounded 4D logs from monocular dashcam videos in the ROADWork~\cite{ghosh2025roadwork} dataset. By recovering structured logs from existing dashcam footage, we make it possible to study rare driving situations at scale without collecting each scene with an instrumented fleet.

Accurate 4D scene recovery from monocular imagery is difficult and considered brittle~\cite{allshire2025visual, xiangli2026honey}. For end-to-end planning, the 4D driving log at a minimum needs metric trajectories. The monocular setting is especially hard
since forward driving motion does not provide significant parallax. Our {\it first observation} is that {\it street imagery can help recover scene geometry} by acting as reliable anchor references. This recovery itself enables training and evaluation of end-to-end planners (\eg VLA models), while the recovered depth supports novel-view synthesis and generative conditioning.

However, driving is an active task with multi-object interactions~\cite{codevilla2018offline, dauner2023pdmclosed} requiring closed-loop simulation. Closed-loop simulation (with privileged planning~\cite{nuplan}) sets a higher bar for 4D recovery: it requires metric and geo-referenced trajectories, localized objects and agents with consistent identities, and mutual consistency among trajectories, agents, objects, and the map for planner rollouts. Our {\it second observation} is that the recorded driving provides a mechanism for checking this consistency. We use {\it non-reactive replay} with an independent map source to assess compatibility of recovered logs with the closed-loop simulation setup.

We make the following contributions toward enabling autonomous driving on long-tailed driving scenes:

\begin{enumerate}
    \item \textbf{A New Benchmark for Long-Tail Driving.} To our knowledge, \benchmarkname{} is the first large-scale benchmark comprised of metric, geo-referenced scenes recovered from in-the-wild dashcam videos to support both end-to-end learning and closed-loop driving evaluation. Our results support the feasibility of scaling driving data in rare situations with abundant dashcam videos.
    \item \textbf{Closed-loop Planning.} We show that Hybrid Pluto~\cite{cheng2024pluto} and rule-based PDM-Closed~\cite{dauner2023pdmclosed} outperform all learned planners by up to 40 points (\secref{cl-planning}) but still perform poorly overall in work-zone driving. Work-zone layouts emerge as the primary generalization challenge, showing the need for dashcam-derived datasets.
    \item \textbf{End-to-end Planning.} Fine-tuning an end-to-end planner addresses the primary challenge of lateral placement in channel-sensitive work zones, reducing lateral trajectory prediction error by 65\% (\secref{e2e-planning}). We observe that adding data from more cities improves both closed-loop and end-to-end planning performance.
    \item \textbf{Novel-View Synthesis and Generative Conditioning.} The recovered depth improves novel-view synthesis by up to 19\% on perceptual metrics (\secref{nvs}), opening the door for sensor simulation from in-the-wild videos. We also explore generative synthesis conditioned on our recovered depth, showing results for ego camera control and agent view synthesis from another vehicle's perspective (\secref{generative-simulation}).
\end{enumerate}

\vspace{-0.5em}
\section{Related Work}
\label{sec:related}

\paragraph{Learning from In-the-wild Data.} Advances in vision and robotics capabilities increasingly come from new, in-the-wild data sources rather than more in-domain collection: pooling heterogeneous robot logs across embodiments~\cite{o2024openxembodiment}, distilling web-scale video into manipulation policies~\cite{cheang2024gr}, curating web imagery for multimodal learning~\cite{gadre2023datacomp}, and using internet-scale data for end-to-end planning~\cite{hwang2024emma}. Many in-the-wild dashcam datasets~\citep{ghosh2025roadwork,zendel2018wilddash,zendel2022unifying} provide long-tail data primarily for 2D perception. Earlier efforts~\citep{deepcrashtest, miao2024dashcam} replay extracted vehicle trajectories as scripted scenarios inside a synthetic engine rather than recovering the full scene. Building on work that recovers depth and view-synthesis supervision from internet imagery~\cite{li2018megadepth,tung2024megascenes,xiangli2026honey} and metric geometry from aerial and street imagery~\cite{vuong2025aerialmegadepth,klingner2013street,vuong2024toward}, \benchmarkname{} extends this approach to 4D logs for end-to-end planning and privileged closed-loop evaluation.

\vspace{-1em}
\paragraph{Autonomous Driving and Simulation.} Autonomous driving simulation falls into two categories. The first builds on data collected by instrumented fleets driving through a small number of cities~\citep{kitti,nuscenes, waymo_motion, nuplan}, with rare-scenario variants~\citep{wod_e2e, wagner2026longtail} also drawn from fleets. The second relies on synthetic simulators~\citep{carla}, with rare and adversarial scenarios~\citep{bench2drive, fail2drive} created by hand. We compare benchmark coverage and simulation protocols in \appref{benchmark-comparison}. Another direction of work addresses sensor simulation, either via generative models~\citep{gaia, cosmos, ren2025cosmos, ren2025gen3c, wang2026sensor2sensor} or via neural rendering~\citep{yang2023unisim, zhou2025hugsim, cao2025pseudo, chen2025omnire}. Both directions require rich conditioning signals and largely rely on fleet data sources. \projectname{} provides physically grounded 4D logs at scale, along with conditioning for sensor simulation, layout~\cite{chitta2024sledge}, and behavior generation from monocular video.

\vspace{-1em}
\paragraph{Planning for Autonomous Driving.} Deployable planners are developed on closed-loop simulators~\citep{nuplan, waymax, zhou2025hugsim, alpasim2025}, with rule-based~\citep{dauner2023pdmclosed}, learned, and hybrid planners~\citep{li2024hydramdp, zheng2025diffusion, cheng2024pluto, ghosh2026rad} all actively explored in the privileged planning setting. End-to-end planners~\citep{sima2024drivelm,zhou2025opendrivevla,xu2025drivegpt4,diffusiondrive,lu2026onevl} are also being heavily explored; they primarily target image-to-trajectory evaluation, are generally evaluated using fleet-derived benchmarks~\citep{wod_e2e, navsim}, and assume multi-modal sensor inputs. In both settings, all of these methods are trained and evaluated on fleet logs, which have limited long-tail coverage. We propose the \clbenchmarkname{} benchmark for training and evaluation on closed-loop long-tail scenarios. \eebenchmarkname{} additionally supports end-to-end training and evaluation from dashcams.    


\section{Constructing \benchmarkname{}}
\label{sec:framework}

\begin{figure}[!t]
    \centering
    \includegraphics[width=\confweb{1.0}{1.0}\linewidth]{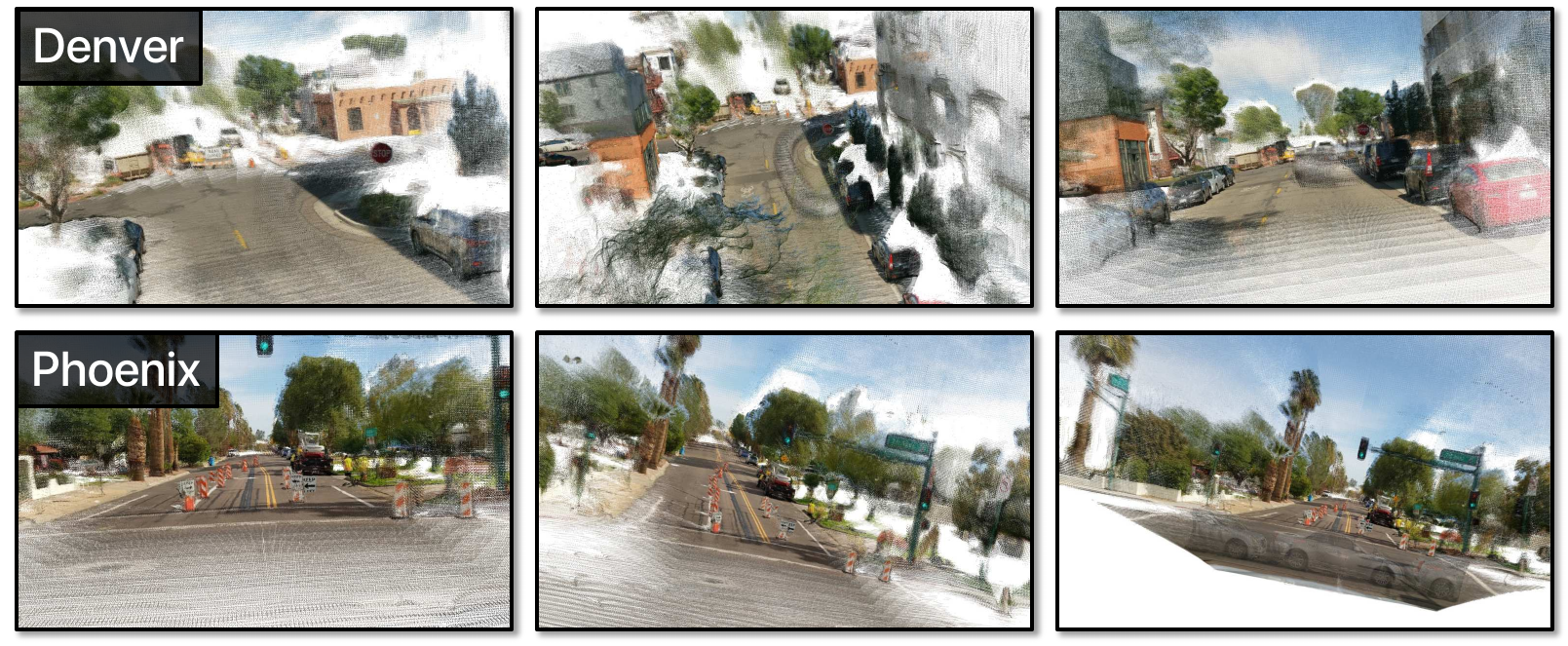}
    \vspace{-1em}
\caption{{\textbf{Metric and geo-referenced point clouds for \benchmarkname{} scenes recovered by \projectname{}.} Roads, lane markings, sidewalks, vegetation, and small long-tailed work-zone objects (cones, tubular markers, signs) are reconstructed from in-the-wild monocular videos. \textbf{Please zoom in.} \textit{Supplemental material contains additional media and visualizations.}}}
    \vspace{-1em}
\label{fig:pointclouds}
\end{figure}

\benchmarkname{} pairs dashcam videos with metric ego trajectories, dense depth, and tracked 3D agents and objects supporting many applications. Metric trajectories provide supervision for end-to-end planning, tracked agents and objects support privileged closed-loop evaluation, and dense depth supports novel-view synthesis and generative conditioning.

We construct these attributes with \projectname{}. We anchor reconstruction to street imagery, detect and track agents and objects in 2D~\cite{ghosh2025roadwork,carion2025sam}, lift them to 3D~\cite{huang2026wilddet3d} conditioned on the recovered depth, and produce logs compatible with existing simulators. \appref{framework-details} details the pipeline, including how we detect and track long-tailed objects.

\vspace{-1em}
\paragraph{Metric Monocular Reconstruction with Street Imagery.} Monocular reconstruction is a critical element of 4D driving log recovery \label{sec:reconstruction} but is scale-ambiguous~\cite{hartley2003multiple} and forward motion in driving does not offer enough parallax for reliable reconstruction. 
But street imagery provides dense, GPS-tagged anchors, building on prior work on localization~\cite{zamir2016generic, vuong2024toward, vuong2025aerialmegadepth}. We retrieve street images, and run SfM over the dashcam and street images together. A similarity transform from the SfM-recovered street imagery poses to their known GPS locations then transfers scale and geo-references the dashcam trajectory. Including street imagery provides high reconstruction accuracy for monocular videos. A detailed description of our approach is provided in \appref{geometry-blocks} and we verify our pose estimates on nuScenes-mini~\cite{nuscenes} in \secref{validation} (\Tabref{pose-eval}). Finally, we use dashcam GPS only to narrow down retrieval of street imagery, and visual place recognition~\cite{berton2025megaloc} can potentially replace it when no GPS is available (\appref{vpr}).



\vspace{-1em}
\paragraph{Constructing the Privileged Closed-Loop Benchmark.}
\label{sec:simulation}

Passive sensorimotor (open-loop) evaluation does not necessarily reflect true driving performance~\cite{codevilla2018offline, dauner2023pdmclosed} compared to active sensorimotor (closed-loop) evaluation~\cite{nuplan, waymax, zhou2025hugsim, alpasim2025}. Closed-loop evaluation lets the planner control the simulated vehicle and respond to the consequences of its decisions. We consider the privileged planning setup~\cite{nuplan}, where the planner receives agent states and a map directly. We first check whether the recovered entities are consistent enough to support simulation.

\begin{figure}[!t]
    \centering
    \includegraphics[width=\linewidth]{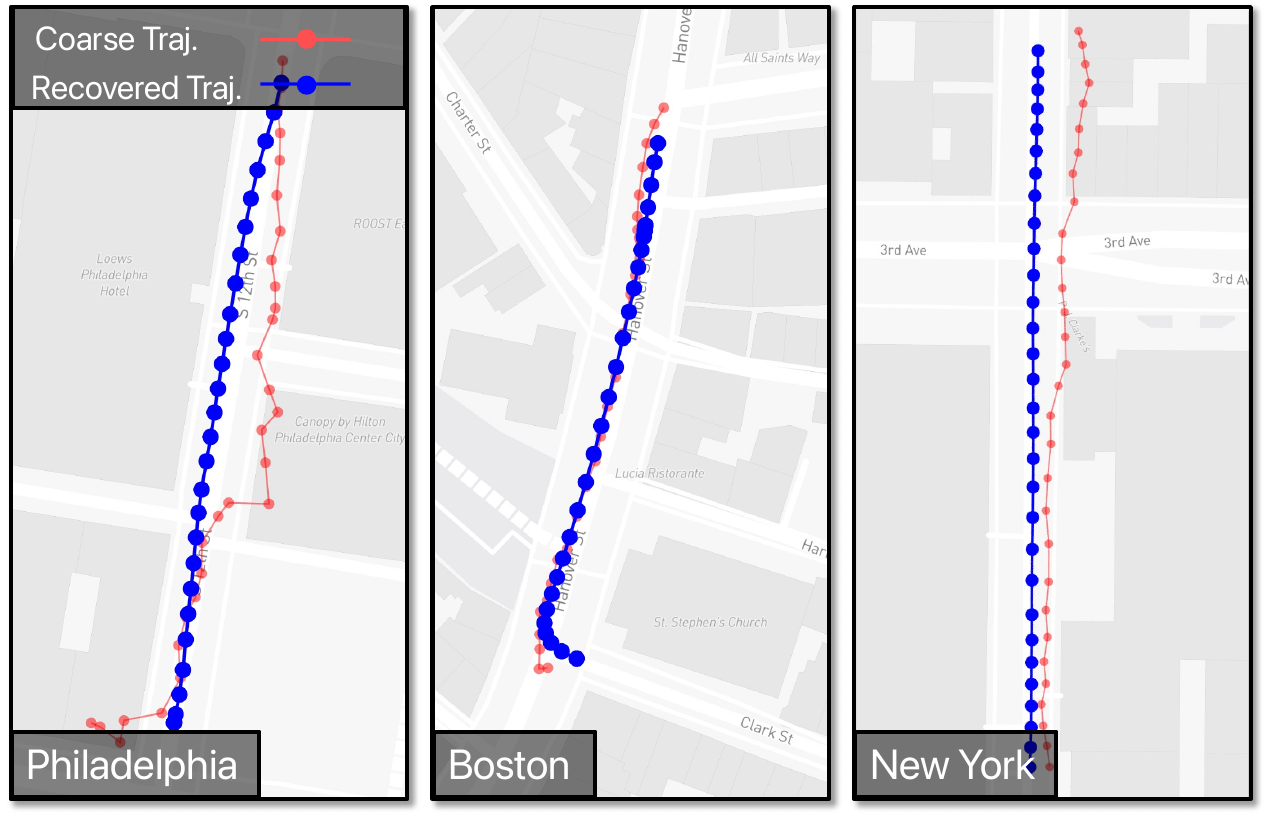}
    \caption{{\textbf{Recovered ego trajectories.} Red: coarse GPS tracks from the dashcam video. Blue: metric 6-DoF poses recovered by \projectname{}. Coarse GPS drifts into adjacent blocks, while recovered poses align with the driven lanes.}}
    \label{fig:ego-trajectories}
    \vspace{-1em}
\end{figure}
%

All our source videos correspond to drivers completing their routes without any incidents.
So, good driving in the real world must be reflected in the simulator's score. We replay the recorded ego trajectory as the \textit{planner's} output and roll out with non-reactive agents on an independently maintained map. The simulator scores the recovered trajectory against the recovered objects and agents and the map. We select \clbenchmarkname{} using nuPlan~\cite{nuplan} and an OpenStreetMap-derived map~\cite{openstreetmap}, with scenarios scoring above 90 under channel-insensitive replay. The score measures compatibility with the privileged closed-loop simulation. We discuss how map errors and the simulator's driving criteria affect this selection in \appref{discussion}.


\vspace{-1em}
\noindent\paragraph{Data source.} 
\label{sec:data-source}
We use the ROADWork~\cite{ghosh2025roadwork} dataset as our primary data source, as it provides dashcam video recordings of navigating through work zones across the U.S. We recover \statFullScenarios{} scenarios from 4,375 source videos to form \benchmarkname{}. Our dataset spans \textbf{\statFullCities{} US cities} and \textbf{\statFullKm} of recovered ego trajectory, with \textbf{\statFullTracks} instance tracks providing \textbf{\statFullBoxes} per-frame 3D boxes across both common dynamic agents and long-tailed road objects.

\vspace{-1em}
\noindent\paragraph{Dataset Splits.} 
\label{sec:dataset-splits}
\eebenchmarkname{} uses all \statFullScenarios{} recovered scenes, with 3,732 for training and 512 held out for evaluation. \clbenchmarkname{} comprises \statVerScenarios{} selected scenes, with 1,689 for training and the same evaluation set. Both benchmarks cover \statFullCities{} cities. For the training-data comparison, \textbf{CL} denotes the 1,689 shared training scenes, \textbf{E2E-Only} the 2,043 additional training scenes, and \textbf{E2E} their union. All models use the same held-out evaluation set.

\begin{table}[t]
\centering
\caption{\textbf{Street imagery improves sparse reconstruction on nuScenes-mini.} Street imagery raises reconstruction success (Recon.) from 43\% to 100\% and recovers near-unit scale. Coarse GPS retains street imagery but uses noisy position anchors, increasing scale and relative translation errors. We report median scale ratios and pose errors across daytime scenes. $^{*}$Results over successfully reconstructed scenes only.}
\label{tab:pose-eval}
\footnotesize
\setlength{\tabcolsep}{2.5pt}
\renewcommand{\arraystretch}{1.12}
\begin{tabularx}{\linewidth}{l*{5}{>{\centering\arraybackslash}X}}
\toprule
Condition & Recon. & Scale & \shortstack{ATE\\(m)} & \shortstack{RPE$_t$\\(m)} & \shortstack{RPE$_r$\\(\textdegree)} \\
\midrule
No street imagery & 43\% & 1.5708$^{*}$ & 0.070$^{*}$ & 3.311$^{*}$ & 0.48$^{*}$ \\
Coarse GPS & 100\% & 1.6102 & 0.084 & 4.010 & 0.10 \\
\rowcolor{tablehighlightgreen} Street imagery (Ours) & 100\% & 1.0024 & 0.084 & 0.107 & 0.10 \\
\bottomrule
\end{tabularx}
\end{table}

\vspace{-1em}
\paragraph{Annotation Validation.}
\label{sec:validation}
Our dashcam videos lack ground-truth ego poses and LiDAR, so we use nuScenes-mini~\cite{nuscenes} for pose and sparse geometry evaluation. Anchoring with street imagery recovers metric ego trajectories with median translation error within 10\,cm and rotation error within $1^{\circ}$ (\Tabref{pose-eval}). Anchoring with street imagery also introduces a 3--9\,m global alignment error (\appref{pose-eval}), which matters for map-based simulation and is not captured by trajectory metrics after alignment. We validate the value of our recovered depth via novel-view synthesis (\secref{nvs}). We also show that dashcam GPS, while useful, is not necessary for street imagery retrieval (\appref{vpr}). We validate 3D object lifting on a long-tailed dataset, WorkZone3D~\cite{sural2026workzone3d} (\appref{wilddet-eval}).

\begin{table}[t]
\centering
\caption{\textbf{Dense depth agreement with sparse geometry.} \projectname{} achieves the strongest agreement with COLMAP on the 512-scene evaluation set. The gap to feedforward models grows at the tighter $\delta_{0.5}$ and $\delta_{0.25}$ thresholds. $\delta_t$ measures agreement within a depth ratio of $1.25^t$ (\appref{depth-agreement}).}
\label{tab:colmap_agreement}
\footnotesize
\setlength{\tabcolsep}{2.5pt}
\renewcommand{\arraystretch}{1.12}
\begin{tabularx}{\linewidth}{@{}>{\raggedright\arraybackslash}Xccccc@{}}
\toprule
Method & AbsRel $\downarrow$ & \shortstack{RMSE $\downarrow$\\(m)}
       & \shortstack{$\delta_1\,\uparrow$\\(\%)}
       & \shortstack{$\delta_{0.5}\,\uparrow$\\(\%)}
       & \shortstack{$\delta_{0.25}\,\uparrow$\\(\%)} \\
\midrule
MapAnything~\cite{keethamapanything} & 0.162 & 7.74 & 82.98 & 61.45 & 37.83 \\
VGGT~\cite{wang2025vggt} & 0.148 & 7.86 & 82.77 & 62.15 & 39.20 \\
VGGT-$\Omega$~\cite{vggtomega} & 0.113 & 6.66 & 88.69 & 73.00 & 51.18 \\
\rowcolor{tablehighlightgreen} Dash2Sim (MP-SfM~\cite{pataki2025mpsfm}) & 0.072 & 5.83 & 93.71 & 88.53 & 79.18 \\
\bottomrule
\end{tabularx}
\end{table}
\paragraph{Role of Street Imagery for Sparse Geometry Recovery.} We test whether street imagery enables the metric recovery needed to construct driving benchmarks from monocular dashcam videos (\Tabref{pose-eval}). Replacing panorama anchors with simulated consumer-grade GPS increases translation RPE from $0.11$\,m to $4.01$\,m and the recovered scale ratio from $1.002$ to $1.610$. Removing street imagery causes reconstruction to fail on more than half the scenes due to insufficient monocular parallax. Street imagery thus provides both metric anchors and the cross-view diversity needed for SfM convergence for monocular driving videos. We additionally assess the accuracy of recovered sparse geometry against LiDAR points of nuScenes-mini (\supptabref{tab:sparse-lidar}).

\begin{table*}[t]
\centering
\begin{minipage}[t]{0.48\textwidth}
\centering
\vspace{0pt}
\captionof{table}{{\textbf{Closed-loop Evaluation on \clbenchmarkname{}.} Learned planners~\cite{zheng2025diffusion, cheng2024pluto, plantf} struggle to generalize, while the hybrid planner~\cite{cheng2024pluto} and the rule-based planner~\cite{dauner2023pdmclosed} perform much better.}}
\label{tab:main_results}
\footnotesize
\setlength{\tabcolsep}{2pt}
\renewcommand{\arraystretch}{1.12}
\begin{tabularx}{\linewidth}{l*{4}{>{\centering\arraybackslash}X}}
\toprule
& \multicolumn{2}{c}{\shortstack{\textbf{CLS-NR}\\\strut}} & \multicolumn{2}{c}{\shortstack{\textbf{CLS-R}\\\strut}} \\
\cmidrule(lr){2-3} \cmidrule(lr){4-5}
\textbf{Planner} & \shortstack{Score$\uparrow$\\\strut} & \shortstack{Coll.\%$\downarrow$\\\strut} & \shortstack{Score$\uparrow$\\\strut} & \shortstack{Coll.\%$\downarrow$\\\strut} \\
\midrule
\multicolumn{5}{l}{\textit{Learned}} \\
PlanTF~\cite{plantf}         & \worstcell{17.7} & \secondcell{4.3} & \worstcell{18.0} & 4.3 \\
Diffusion Planner~\cite{zheng2025diffusion}      & 24.0 & \worstcell{51.8} & 31.3 & \worstcell{39.1} \\
Pluto~\cite{cheng2024pluto} & 32.7 & 10.5 & 40.3 & 6.8 \\
\midrule
\multicolumn{5}{l}{\textit{Rule-Based}} \\
PDM-Closed     & \secondcell{53.0} & \bestcell{2.2} & \secondcell{53.7} & \secondcell{1.8} \\
\midrule
\multicolumn{5}{l}{\textit{Hybrid}} \\
Pluto~\cite{cheng2024pluto}   & \bestcell{57.5} & 4.7 & \bestcell{59.7} & \bestcell{1.6} \\
\bottomrule
\end{tabularx}
\end{minipage}%
\hfill
\begin{minipage}[t]{0.48\textwidth}
\centering
\vspace{0pt}
\captionof{table}{{\textbf{Effect of channel sensitivity (CLS-NR).} Channel-sensitive evaluation penalizes drivable-area deviations from GT.}}
\label{tab:channel_sensitivity}
\footnotesize
\setlength{\tabcolsep}{2pt}
\renewcommand{\arraystretch}{1.12}
\begin{tabularx}{\linewidth}{l*{4}{>{\centering\arraybackslash}X}}
\toprule
& \multicolumn{2}{c}{\shortstack{\textbf{Channel-}\\\textbf{Sensitive}}} & \multicolumn{2}{c}{\shortstack{\textbf{Channel-}\\\textbf{Insensitive}}} \\
\cmidrule(lr){2-3} \cmidrule(lr){4-5}
\textbf{Planner} & \shortstack{Score$\uparrow$\\\strut} & \shortstack{DAC\\Fail\%$\downarrow$} & \shortstack{Score$\uparrow$\\\strut} & \shortstack{DAC\\Fail\%$\downarrow$} \\
\midrule
\multicolumn{5}{l}{\textit{Learned}} \\
PlanTF~\cite{plantf}         & \worstcell{17.7} & \secondcell{18.0} & \worstcell{22.9} & 5.3 \\
Diffusion Planner~\cite{zheng2025diffusion}      & 24.0 & \worstcell{30.9} & 30.9 & 9.2 \\
Pluto~\cite{cheng2024pluto} & 32.7 & 23.4 & 43.0 & \secondcell{4.3} \\
\midrule
\multicolumn{5}{l}{\textit{Rule-Based}} \\
PDM-Closed     & \secondcell{53.0} & 27.3 & \secondcell{57.4} & \worstcell{9.4} \\
\midrule
\multicolumn{5}{l}{\textit{Hybrid}} \\
Pluto~\cite{cheng2024pluto}   & \bestcell{57.5} & \bestcell{17.8} & \bestcell{68.8} & \bestcell{1.2} \\
\bottomrule
\end{tabularx}
\end{minipage}
\end{table*}
\noindent\paragraph{Dense Depth Agreement with Sparse Geometry.}
\label{sec:colmap_agreement}

\begin{figure}[t]
    \vspace{-1em}
    \centering
    \includegraphics[width=1\linewidth]{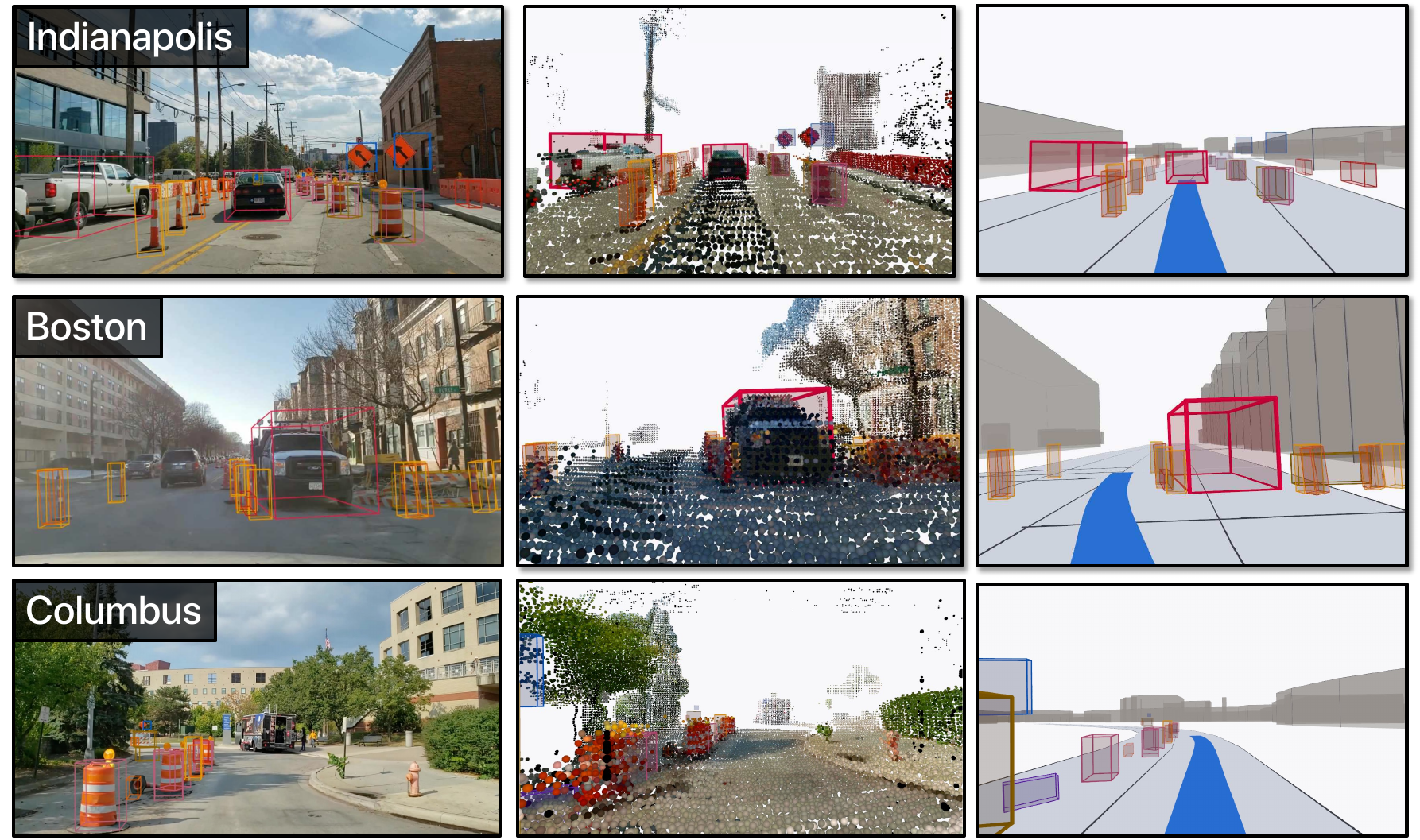}
    \vspace{-1em}
    \caption{{\textbf{Driving logs from \benchmarkname{} recovered by 
    \projectname{}.} Our logs include long-tailed work zone scenarios with rare objects, layouts and complex lane change behaviors such as crossing the double yellow line in a work zone (\textbf{top row}). \textit{Supplemental material contains additional media and visualizations.}}}
    \label{fig:cl-logs-cities-main}
    \vspace{-2em}
\end{figure}

Having validated sparse geometry recovery, we use the recovered SfM points as a reference for our dashcam videos, where LiDAR is unavailable. Similar to MegaDepth~\cite{li2018megadepth}, we evaluate dense depth~\cite{pataki2025mpsfm} through agreement with sparse SfM geometry~\cite{schonberger2016structure, pan2024global} on our 512-scene held-out evaluation set. We compare with VGGT~\cite{wang2025vggt}, VGGT-$\Omega$~\cite{vggtomega}, and MapAnything~\cite{keethamapanything}, excluding sky regions and aligning scale per sequence (\appref{depth-agreement}).

\Tabref{colmap_agreement} shows that depth recovered by Dash2Sim using MP-SfM~\cite{pataki2025mpsfm} achieves the strongest agreement across all metrics, with VGGT-$\Omega$ the closest feedforward alternative. As the depth estimates improve, standard thresholds become less sensitive to the remaining errors. The advantage is especially pronounced at tighter thresholds, e.g. $\delta_{0.25}$: 79.18\% versus 51.18\%, a gain of 28.00 points. These results suggest that feedforward models~\cite{vggtomega, wang2025vggt, keethamapanything} leave room for further gains in geometric precision compared to optimization based reconstruction~\cite{pataki2025mpsfm} on the in-the-wild dashcam videos needed to expand driving data beyond fleet collection.

\section{Applications}


\subsection{Privileged Closed-Loop Planning}
\label{sec:cl-planning}

Temporary work-zone channels can require lane changes and departures from mapped lane centerlines. We use \clbenchmarkname{} to study how privileged planners handle these long-tailed layouts in closed-loop simulation.

\vspace{-1em}
\paragraph{Setup.} We evaluate PlanTF~\cite{plantf}, Diffusion Planner~\cite{zheng2025diffusion}, Pluto~\cite{cheng2024pluto} (learned and hybrid variants), and PDM-Closed~\cite{dauner2023pdmclosed}  . The learned planners are trained on 1 million nuPlan scenario tokens without fine-tuning on \clbenchmarkname{}, except for when we train PlanTF~\cite{plantf} to check learnability and generalization across cities. Each planner receives the inputs according to the nuPlan protocol, and we do not modify any baseline. We report non-reactive (CLS-NR) and IDM-reactive (CLS-R) closed-loop scores with one change in the definition of drivable area to include work-zone-specific channel sensitivity (\appref{cl-planning-details}).

\vspace{-1em}
\paragraph{Results and Insights.} Our results show that a counterintuitive finding from prior work~\cite{dauner2023pdmclosed}, shown on regular driving scenarios, extends to work-zone scenarios: \textit{rule-based and hybrid planners generalize better than learned planners}. The hybrid planner, Pluto~\cite{cheng2024pluto}, achieves the highest overall score by combining a learned planner with a rule-based safety mechanism. A rule-based planner is not far behind, outperforming all purely learned planners by 20 to 35 points in non-reactive evaluation (\Tabref{main_results}).

\begin{figure}[t]
    \vspace{-1em}
    \centering
    \includegraphics[width=\linewidth]{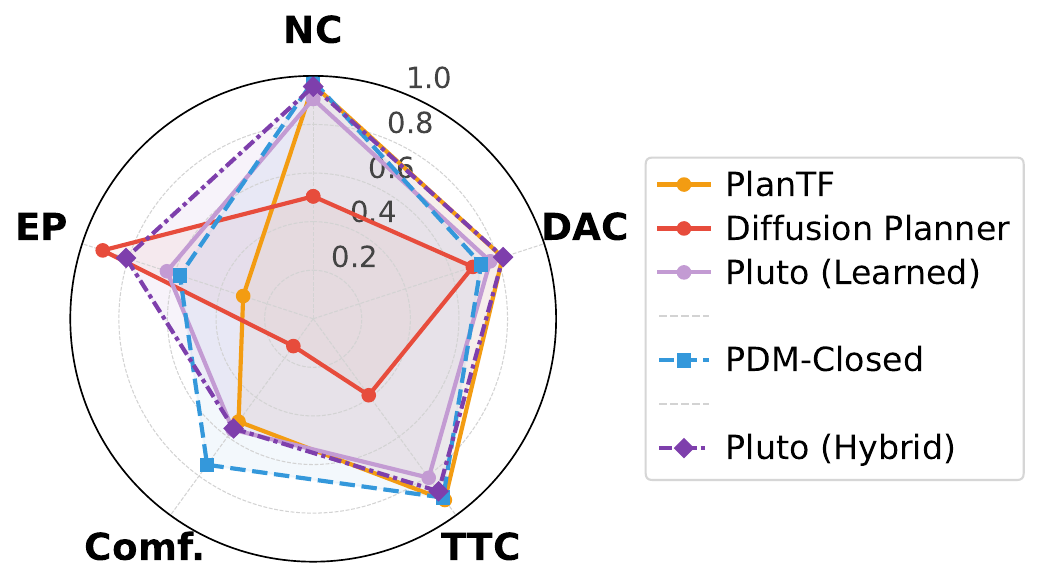}
    \caption{{\textbf{Closed-Loop Planner Analysis.} PlanTF shows low progress, Diffusion Planner is aggressive with a high crash rate, and Pluto Learned trades comfort for progress. PDM-Closed is safe but very conservative. Pluto Hybrid balances all metrics.}}
    \label{fig:cl-comparison-nonreactive-radar}
    \vspace{-1em}
\end{figure}

\noindent\textbf{Generalization Analysis of Planners.} Among the learned planners, Pluto~\cite{cheng2024pluto} outperforms PlanTF~\cite{plantf} and Diffusion Planner~\cite{zheng2025diffusion} by a large margin, and each learned planner fails differently (\figref{cl-comparison-nonreactive-radar}).

\noindent\textbf{Role of Channel Sensitivity in Work Zones.} In channel-sensitive evaluation, the drivable area is restricted to only the lane segments along the ground-truth route the human driver followed. In channel-insensitive evaluation, all mapped lane segments are considered drivable (details in \appref{cl-planning-details}). Lane closures, merges, and detours in work zones frequently require crossing into adjacent lanes or taking alternate paths, a failure mode autonomous vehicles also exhibit~\cite{missionlocal2024waymosfo, techcrunch2026waymoworkzone}.

All planners improve in channel-insensitive mode (\Tabref{channel_sensitivity}). Pluto~\cite{cheng2024pluto} shows the largest score improvement and a reduction in DAC failure rate. We observe that compared to PlanTF~\cite{plantf} and Diffusion Planner~\cite{zheng2025diffusion}, Pluto's~\cite{cheng2024pluto} input representation includes reference lines that encode lane centerlines, which likely provide additional structure for planning in normal driving but reduce channel sensitivity when work-zone closures require deviation from lane centerlines. PDM-Closed~\cite{dauner2023pdmclosed} shows a similar pattern, indicating that centerline following, while helpful in normal driving, is likely suboptimal for navigating work zones.

\noindent\textbf{Additional Insights.} In \appref{cl-planning-details}, we further analyze the effect of \textit{static work zone objects} on planner performance. Removing dynamic agents has little effect on overall scores, indicating that work zone layouts are the primary generalization challenge for privileged motion planners. Also, to verify if motion planning is learnable from \benchmarkname{}, we train PlanTF~\cite{plantf} on increasing amounts of data, and observe that data from even one city improves performance on the full 17-city set. We observe that it nearly matches zero-shot Diffusion Planner~\cite{zheng2025diffusion} in performance when trained on all cities. The supplemental analysis also examines ego progress and drivable-area compliance, and includes qualitative planner rollouts.

\subsection{End-to-End Planning}
\label{sec:e2e-planning}

End-to-end planners map camera input directly to trajectories, but they are trained and evaluated on fleet logs with limited long-tail coverage (\secref{related}). \eebenchmarkname{} pairs dashcam observations with recovered metric trajectories for learning to predict future waypoints from recorded driving. We consider the state-of-the-art Vision-Language-Action Model (VLA) planner, OneVL~\cite{lu2026onevl}, and evaluate its performance on \eebenchmarkname{}, our end-to-end planning benchmark derived from \benchmarkname{}.

\begin{table}[t]
\centering
\caption{\textbf{End-to-end planning on \eebenchmarkname{}.} (a) OneVL~\cite{lu2026onevl} reasoning variants before and after fine-tuning on 3,732 scenes. (b) Answer-Only fine-tuned with different training data. CL denotes 1,689 training scenes shared with the closed-loop benchmark; E2E-Only denotes 2,043 scenes used exclusively for end-to-end planning. E2E combines both groups (3,732 scenes). All models use the same held-out test set.}
\label{tab:vla-e2e}
\small
\setlength{\tabcolsep}{3pt}
\renewcommand{\arraystretch}{1.12}
\begin{tabularx}{\linewidth}{l*{4}{>{\raggedleft\arraybackslash}X}}
\toprule
& \multicolumn{2}{c}{\textbf{Zero-shot}} & \multicolumn{2}{c}{\textbf{Fine-tuned}} \\
\cmidrule(lr){2-3} \cmidrule(l){4-5}
\textbf{(a) Variant} & ADE & FDE & ADE & FDE \\
\midrule
Answer-Only & \bestcell{1.390} & \bestcell{3.326} & \secondcell{0.808} & \secondcell{2.034} \\
Text-CoT & \worstcell{1.547} & \worstcell{3.646} & \worstcell{0.824} & \worstcell{2.046} \\
Latent-CoT & \secondcell{1.525} & \secondcell{3.597} & \bestcell{0.804} & \bestcell{2.028} \\
\bottomrule
\end{tabularx}

\medskip
\begin{tabularx}{\linewidth}{l*{3}{>{\raggedleft\arraybackslash}X}}
\toprule
\textbf{(b) Training data} & Scenes & ADE & FDE \\
\midrule
CL & 1{,}689 & \secondcell{0.827} & \secondcell{2.090} \\
E2E-Only & 2{,}043 & \worstcell{0.847} & \worstcell{2.146} \\
E2E & 3{,}732 & \bestcell{0.808} & \bestcell{2.034} \\
\bottomrule
\end{tabularx}
\end{table}
\begin{figure}[t]
    \centering
    \begin{minipage}[t]{0.49\linewidth}
        \centering
        \includegraphics[width=\linewidth]{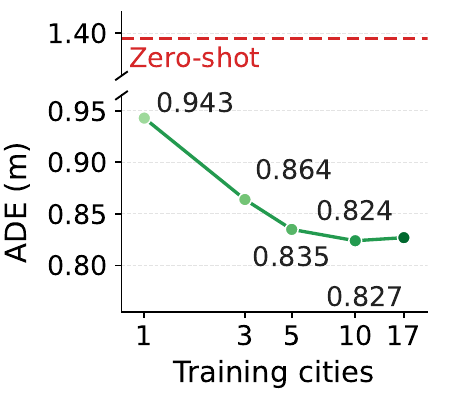}
    \end{minipage}\hfill%
    \begin{minipage}[t]{0.49\linewidth}
        \centering
        \includegraphics[width=\linewidth]{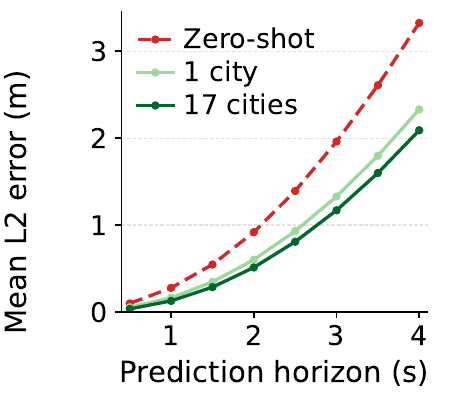}
    \end{minipage}
    \caption{\textbf{End-to-end planning is learnable.} \textbf{(Left)} Training on recovered logs from even one city improves over zero-shot, with further gains from additional cities. \textbf{(Right)} Additional cities help most at longer prediction horizons, which are important for anticipating upcoming lane shifts in work zones.}
    \label{fig:vla-scaling}
\end{figure}
\begin{figure}[!t]
\centering
\includegraphics[width=\linewidth]{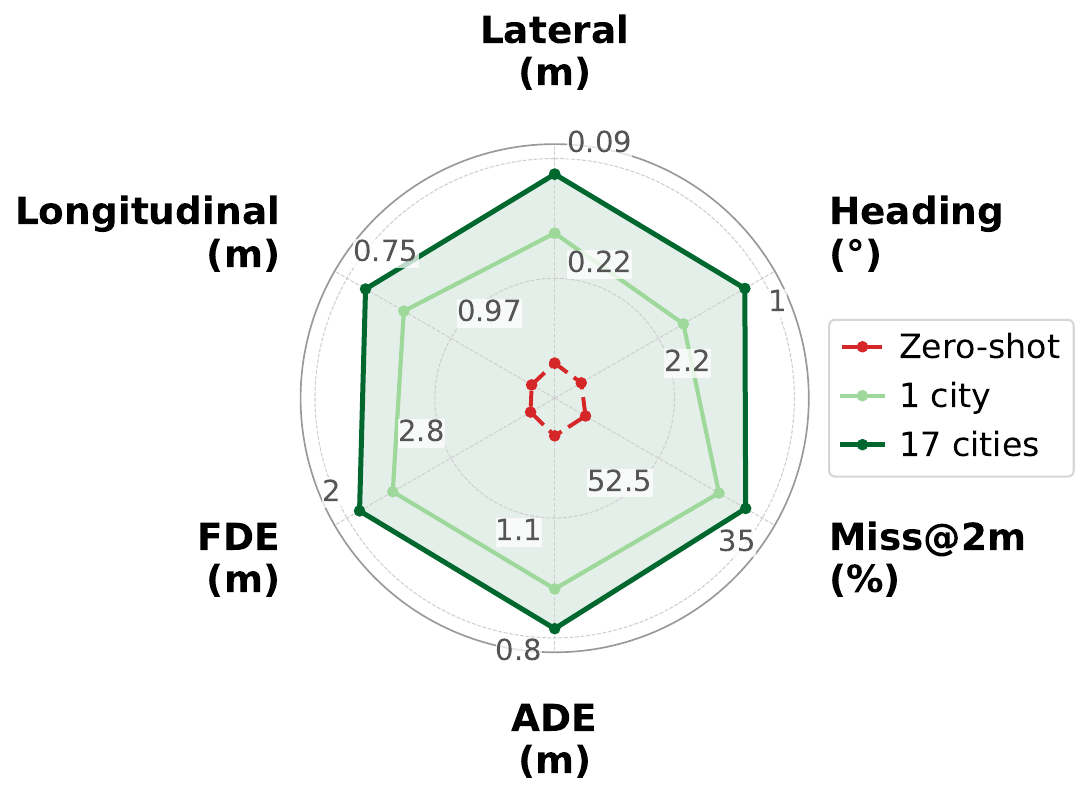}
\caption{\textbf{End-to-End Planner Analysis.} Fine-tuning on recovered logs from 1 and 17 cities improves all metrics over zero-shot, with the largest relative reductions in lateral and heading error. Each axis has its own scale, with lower errors farther from the center.}
\label{fig:vla-error-radar}
\end{figure}
\vspace{-1em}
\paragraph{Setup.} OneVL~\cite{lu2026onevl} is trained on the NAVSIM~\cite{navsim} benchmark and predicts 8 future waypoints over a 4\,s horizon from a given image and ego state, without privileged perception or a map as input. We report performance on the 512-scene held-out set shared with \clbenchmarkname{} (\appref{e2e-planning}). For fine-tuning the VLA, we use 3{,}732 training scenes from all 17 cities, and employ LoRA fine-tuning following~\cite{lu2026onevl}.

\begin{table*}[t]
\centering
\small
\caption{\textbf{Held-out Novel-View Gaussian Splat Reconstruction Results.} Adding depth supervision from \projectname{} improves both low-level image processing metrics and high-level perceptual metrics while performing novel-view synthesis.
}
\vspace{-1em}
\label{tab:nvs-test}
\resizebox{0.8\linewidth}{!}{%
\begin{tabular}{lccccccc}
\toprule
Method
  & PSNR$\uparrow$
  & SSIM$\uparrow$
  & LPIPS$\downarrow$
  & FID$\downarrow$
  & cFID$\downarrow$
  & DINO-S$\downarrow$
  & DSim$\downarrow$ \\
\midrule
OmniRe~\cite{chen2025omnire}
  & \worstcell{22.21} & \worstcell{0.769} & \worstcell{0.269} & \worstcell{104.1} & \worstcell{18.5} & \worstcell{0.090} & \worstcell{0.120} \\
OmniRe~\cite{chen2025omnire}\,+\,Ours
  & \bestcell{22.42\,{\scriptsize(+0.9\%)}} & \bestcell{0.777\,{\scriptsize(+1.0\%)}} & \bestcell{0.232\,{\scriptsize(-14\%)}} & \bestcell{89.7\,{\scriptsize(-14\%)}} & \bestcell{14.9\,{\scriptsize(-19\%)}} & \bestcell{0.081\,{\scriptsize(-10\%)}} & \bestcell{0.101\,{\scriptsize(-16\%)}} \\
\bottomrule
\end{tabular}
}
\vspace{-1em}
\end{table*}
\vspace{-1em}
\paragraph{Results and Insights.} Zero-shot VLAs perform poorly on work zones regardless of their reasoning paradigm (\Tabref{vla-e2e}), consistent with the poor generalization in closed-loop planning (\secref{cl-planning}). Fine-tuning on all 3{,}732 training scenes reduces ADE from 1.390\,m to 0.808\,m, a gain much larger than the differences across reasoning variants. After fine-tuning, the reasoning variants perform similarly to answer-only, with Latent-CoT slightly improving ADE from 0.808\,m to 0.804\,m (\Tabref{vla-e2e}).

\noindent\textbf{Planner Error Analysis.} Fine-tuning improves the lateral motion error (0.31\,m to 0.11\,m) and the heading error by roughly $3\times$ (\figref{vla-error-radar}). Lateral placement also remains challenging in privileged closed-loop evaluation, where planners receive recovered agent states and a map directly (\Tabref{channel_sensitivity}).

\noindent\textbf{Training Data Comparison.} For Answer-Only, the 1,689 CL training scenes give 0.827\,m ADE, versus 0.847\,m for the 2,043 E2E-Only scenes. Training on all 3,732 scenes gives the lowest ADE (0.808\,m) and FDE (2.034\,m) in this comparison (\Tabref{vla-e2e}). Scenes excluded by our closed-loop selection criterion can therefore still provide useful supervision for end-to-end planning (\appref{discussion}).

\noindent\textbf{Scaling Training Across Cities.} We use the same CL training scenes and city groups for both planning tasks to compare how training across cities affects closed-loop and end-to-end performance. Training on a single city (Los Angeles) reduces mean ADE from 1.390\,m to 0.943\,m and improves prediction in 15 of the other 16 cities (\appref{e2e-planning}). Adding cities improves accuracy monotonically up to 10 cities (0.824\,m), after which the marginal utility plateaus (\figref{vla-scaling}, left). We observe similar trends in closed-loop planning (\suppfigref{fig:plantf-data-scaling}), showing that scaling training across cities benefits both.

\noindent\textbf{Analysis Across Temporal Horizons.} Fine-tuned models outperform zero-shot at every prediction horizon (\figref{vla-scaling}, right). The models trained on 1 and 17 cities are nearly indistinguishable for the first 1.5\,s and differ mainly at prediction horizons beyond 1.5\,s. Note that better long-horizon planning is necessary in work zones, as the model must anticipate the work-zone layout, such as an upcoming lane shift, rather than extrapolating its current motion~\cite{li2024ego}.

\noindent\textbf{Additional Insights.} In \appref{e2e-planning}, we further analyze per-city generalization, reasoning supervision from recovered 3D annotations, and trace augmentation for Text-CoT. We show qualitative examples of predicted trajectories and reasoning after fine-tuning.

\begin{figure}[t!]
    \centering
    \includegraphics[width=\confweb{1.0}{1.0}\linewidth]{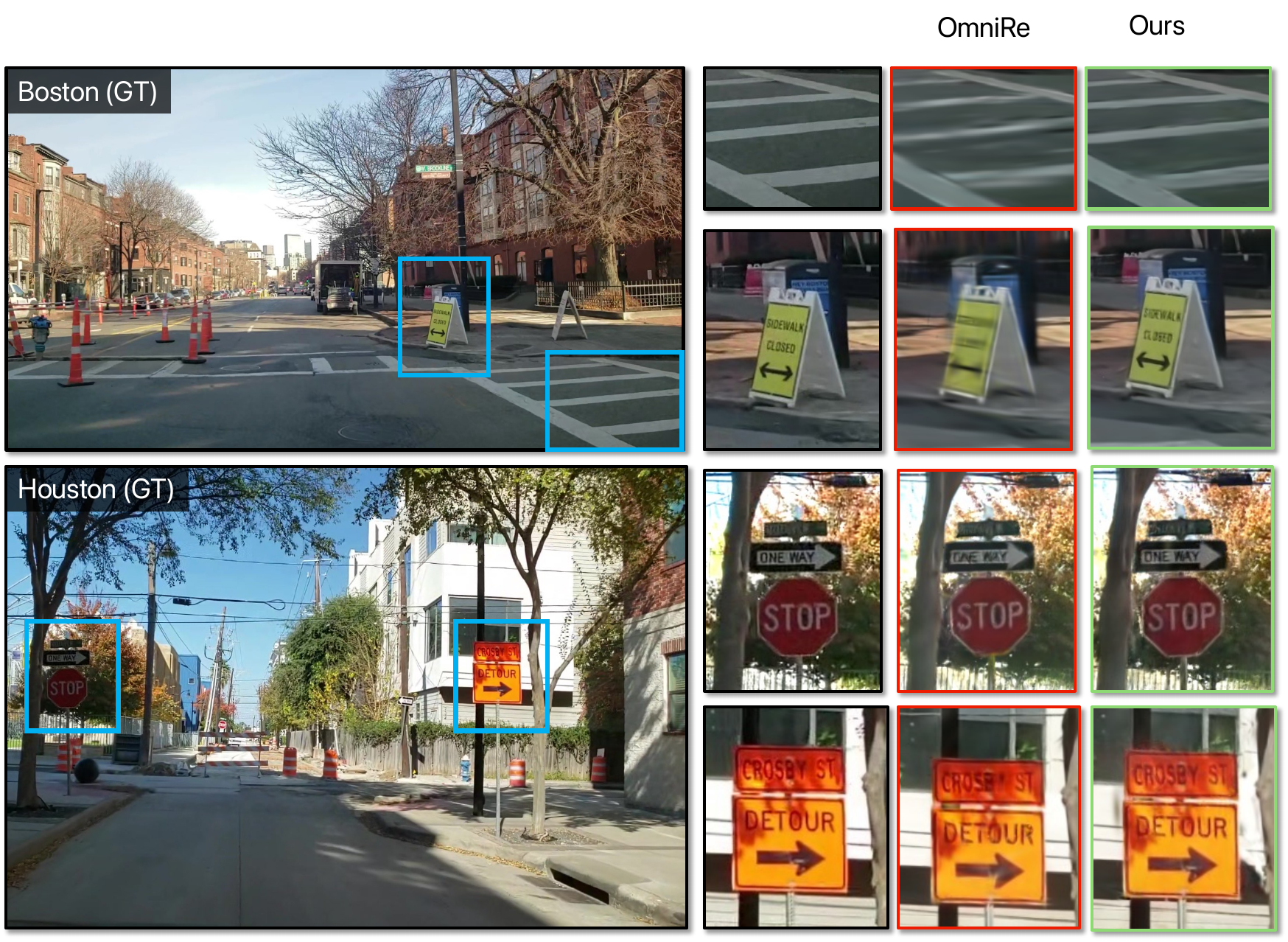}
    \vspace{-2em}
    \caption{{\textbf{Gaussian Splat Reconstruction with our Depth Supervision.} Dense depth from \projectname{} improves novel-view synthesis, especially for small objects such as vertical panels and temporary traffic control signs that contain textual details relevant for navigating work zones~\cite{ghosh2025roadwork}. \textbf{Best viewed zoomed in.}}}
    \label{fig:nvs-compare-details}
    \vspace{-1em}
\end{figure}
%

\begin{figure*}[t]
    \centering
    \includegraphics[width=\linewidth]{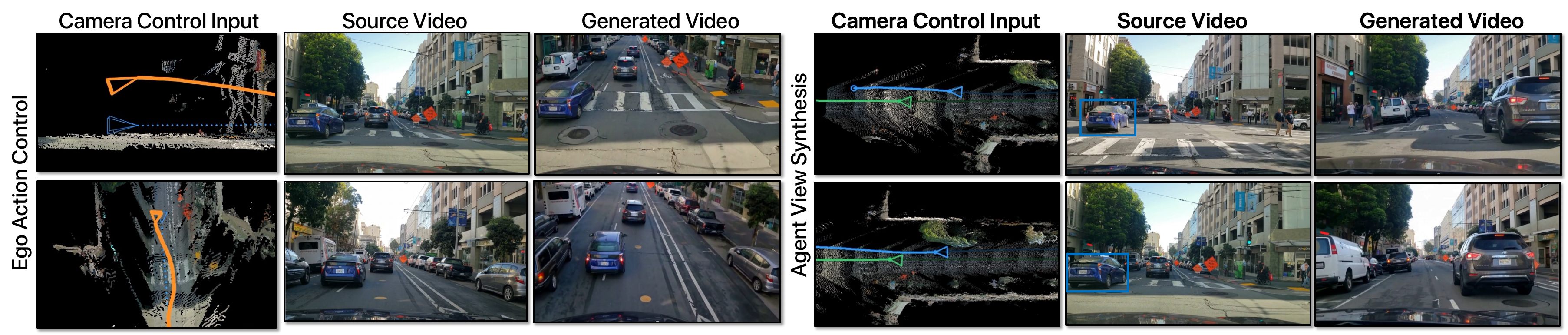}
    \caption{{\textbf{Qualitative results of Camera-Controlled Generative Synthesis.} We show generative view synthesis under large viewpoint changes, including elevated camera trajectories and views from another vehicle. \textbf{(Left)} \textbf{Ego action control} generates views along a \textcolor{flame}{\textbf{modified camera trajectory}}, with the \textcolor{ourmediumblue}{original ego trajectory} (dotted). \textbf{(Right)} \textbf{Agent view synthesis} generates views from the perspective of \textcolor{ourmediumblue}{\textbf{another vehicle}}. The camera control input shows the \textcolor{darkspringgreen}{ego trajectory} and the \textcolor{ourmediumblue}{\textbf{highlighted agent's trajectory}}. \textbf{Best viewed zoomed in.} \textit{Supplemental material contains additional media and visualizations.} }}
    \label{fig:generative-conditioning-control}
\end{figure*}

\subsection{Gaussian Splat Reconstruction}
\label{sec:nvs}

Directly evaluating depth recovered from in-the-wild imagery is difficult because ground-truth depth is unavailable. Following prior work~\cite{vuong2025aerialmegadepth, chen2025omnire}, we use novel-view synthesis as a practical downstream evaluation: whether the depth recovered by \projectname{} improves rendering of held-out views. This also showcases one necessary ingredient for sensor simulation, which can test perception-dependent failures by rendering sensor streams during simulation rollouts~\cite{yang2023unisim, zhou2025hugsim}.

\vspace{-1em}
\paragraph{Setup.}
We reconstruct 100 geographically diverse \benchmarkname{} scenes (due to computational limitations) with the widely adopted OmniRe~\cite{chen2025omnire} framework for novel-view synthesis using Gaussian Splatting~\cite{kerbl2023gaussian}. The baseline uses random splat initialization and photometric optimization. OmniRe\,+\,Ours uses our recovered depth (\secref{reconstruction}) for splat initialization and an inverse-depth $\ell_1$ loss. Every 10th frame is held out for novel-view evaluation following~\cite{chen2025omnire}.

\vspace{-1em}
\paragraph{Reconstruction quality.}  Depth supervision improves every metric on the held-out test split
(\Tabref{nvs-test}). LPIPS improves by 14\%, CLIP-FID by 19\%, DreamSim by 16\%, and PSNR improves by 0.9\%.

Since small long-tail objects occupy few pixels, we also evaluate rendering quality on object crops across work zone categories. Depth supervision improves LPIPS across all nine categories, with the largest gain on cones ($20\%$). \Figref{nvs-compare-details} shows visual comparisons: depth improves perceptual details of both common and long-tailed objects, including legibility of temporary signage that end-to-end models must interpret to navigate work zones~\cite{ghosh2025roadwork}.

\noindent\textbf{Additional Insights.} Full-split results (\appref{nvs-details}) also show improved rendering of training views, implying that depth regularizes the underlying Gaussian Splat representations. We also show initial results of rendering additional camera viewpoints.


\subsection{Camera-Controlled Generative Synthesis}
\label{sec:generative-simulation}

Generative world models~\cite{gaia, ren2025cosmos} show  potential for generating controllable driving scenarios for training and evaluation. Depth-conditioned video models~\cite{ren2025gen3c, vuong2026fixanything} use geometry rendered along a camera trajectory to guide generation and complete regions unseen from the source viewpoints.

To explore the potential of generative world model conditioning, we treat the camera control problem as an inpainting problem over point-cloud renderings as conditioning input to a generative video model. We render our recovered point clouds along the requested camera trajectory and condition MiniMax H3~\cite{minimax2026h3} on these renderings and the source video to complete missing regions. Our recovered depth is good enough (see \S \ref{sec:colmap_agreement}) to directly condition generative models for camera control under large viewpoint changes.

\Figref{generative-conditioning-control} shows qualitative results using our recovered scenes: ego action control changes the camera trajectory, while agent view synthesis generates the scene from another vehicle's perspective. This requires completing views that were not captured by the dashcam, showing how our recovered geometry can condition generative view synthesis beyond the recorded ego viewpoint. We describe the conditioning setup in \appref{generative-conditioning}.


\section{Conclusion}
\label{sec:conclusion}

\projectname{} improves accessibility by enabling learning from in-the-wild videos, a widely available data source, extending the long-tail coverage established by ROADWork~\cite{ghosh2025roadwork} from the 2D to the 4D regime. \benchmarkname{} provides \statFullScenarios{} recovered work-zone driving scenes across \statFullCities{} cities for studying autonomous driving from dashcams. Qualitative results on other dashcam video datasets~\cite{yu2020bdd100k, wang2024dc} (\suppfigref{fig:zero-shot-dcvr-bdd100k}) show Dash2Sim's broader applicability.

We show that planners trained and evaluated on fleet data do not generalize reliably to long-tailed work zone scenarios. We identify channel sensitivity as the major bottleneck of planning in work zones. Training on our data significantly reduces lateral trajectory error, and adding data from more cities improves planning performance. Our recovered depth improves rendering of small objects like temporary traffic control signs and conditions generative models for camera control, including views from another vehicle.

\noindent\textbf{Limitations and Future Work.} A single camera may leave important aspects of the scene unobserved and be insufficient for learning some driving behaviors. Errors in reconstruction, and object and agent recovery and mismatched maps can affect recovered logs. Our closed-loop benchmark is constructed for nuPlan and extending it to other simulators will add requirements. We discuss these limitations, agent behavior, and deployment validation in \appref{limitations}. Improving 4D reconstruction~\cite{pan2024global, pan2026gluemap, wang2025vggt, keethamapanything, vggtomega}, fleet-like sensor generation from dashcams~\cite{ren2025cosmos, gaia, wang2026sensor2sensor}, learned traffic models~\cite{hagedorn2025planners}, and real-world validation~\cite{luo2025sim2val} are promising future research directions.


\ifARXIV
  \paragraph{LLM Usage Statement}
We have used AI assistance from OpenAI Codex and Claude Code for writing code, running experiments, and polishing the text in the manuscript. We have reviewed all the outputs from these LLMs (including any code artifacts and experimental scripts), and have manually reviewed and referenced the text, references, code and the associated data. We take full responsibility for all results and content included in the manuscript.

  {\small\bibliography{refs}}
\fi

\clearpage
\ifSUBMISSION
  \nolinenumbers
  \immediatewrites
  \setcounter{bodylastpage}{\value{page}}
  \DISCARDPAGEStrue
\fi
\ifSUPPLEMENTAL
  \setcounter{page}{1}
  \DISCARDPAGESfalse
  \deferredwrites
  \linenumbers[1]
  \maketitlesupplementary
\fi
\appendix
\raggedbottom
\makeatletter
\setlength{\@fptop}{0pt}
\setlength{\@dblfptop}{0pt}
\setlength{\@fpsep}{12pt}
\setlength{\@dblfpsep}{12pt}
\makeatother
\renewcommand{\thetable}{\thesection.\arabic{table}}
\renewcommand{\thefigure}{\thesection.\arabic{figure}}
\setcounter{table}{0}
\setcounter{figure}{0}
\makeatletter
\@addtoreset{table}{section}
\@addtoreset{figure}{section}
\makeatother


\section{Discussion}
\label{app:discussion}


\begin{table}[!htbp]
\centering
\small
\setlength{\tabcolsep}{4pt}
\renewcommand{\arraystretch}{1.05}
\caption{{\textbf{\benchmarkname{} at a glance.}}}
\label{tab:stats-full}
\begin{tabular}{@{}lr@{}}
\toprule
\rowcolor{tablehighlight}[0pt][0pt]
\multicolumn{2}{@{}l}{\emph{Scale}} \\
Scenarios            & \statFullScenarios \\
US cities            & \statFullCities \\
Frames @ 5\,Hz       & \statFullFrames \\
Driving time         & \statFullHours \\
\midrule
\rowcolor{tablehighlight}[0pt][0pt]
\multicolumn{2}{@{}l}{\emph{Reconstruction}} \\
SfM 3D points        & \statFullSfMPoints \\
\midrule
\rowcolor{tablehighlight}[0pt][0pt]
\multicolumn{2}{@{}l}{\emph{3D annotations}} \\
Instance tracks      & \statFullTracks \\
\quad dynamic / static & \statFullDyn{} / \statFullStat \\
Per-frame 3D boxes   & \statFullBoxes \\
\bottomrule
\end{tabular}
\end{table}

\noindent\textbf{Annotation Accuracy and Closed-Loop Eligibility.} Recovery accuracy measures how closely the recovered scene matches the reference, including pose, 3D box, and depth errors. Closed-loop eligibility concerns whether the recovered scene supports the chosen simulation protocol. We report recovery accuracy wherever reference data are available. Replay-based selection addresses the additional scene and map requirements for closed-loop use, rather than providing a universal measure of annotation quality.

\noindent\textbf{Scope of Replay-Based Selection.} We select \clbenchmarkname{} using non-reactive replay of the recorded ego trajectory, with channel-insensitive scoring and a predefined threshold. The score combines driving criteria such as drivable-area compliance and collision avoidance against an independently maintained map. 

This mechanism can measure certain inconsistencies (e.g. map errors, errors in the ego pathway due to 3D reconstruction, phantom objects appearing in the driven pathway), but does not test every aspect of the scene. For example, semantic errors do not affect the log-replay score for the most part (nuPlan does distinguish between collisions with static objects and dynamic agents). It also requires that source videos contain no at-fault driving events, so extending it to, say, videos of accidents requires distinguishing real collisions from reconstruction artifacts. Lastly, and more fundamentally, replay-based selection inherits the simulator's notion of ``good driving'', so a faithfully recovered log can still be penalized when the real driving departs from such a notion. The comfort, drivable-area, and time-to-collision criteria used here can penalize maneuvers observed in work zones.

Concretely, even when the recovered 4D log is correct, its score can be low because \textit{(a)} comfort terms penalize the hard braking and sharp steering that are routine in dense or evasive driving; \textit{(b)} drivable-area and driving-direction checks penalize crossing into a closed or oncoming lane, even when cones, a flagger, or a contraflow channel make it the only legal path (during replay with nuPlan~\cite{nuplan}, we use the proposed \textit{channel-insensitive} mode for this reason); \textit{(c)} a static map does not represent temporary work zones, so a correct trajectory may appear to leave the road entirely; and \textit{(d)} near-misses and recoveries from another agent's mistake can receive low time-to-collision (TTC) scores even when accurately reconstructed. The replay threshold therefore specifies compatibility with this simulation protocol.

The selection rule defines the closed-loop benchmark for the chosen simulator and metrics. End-to-end training uses recorded observations and trajectory targets. Our training-data comparison shows that the CL training scenes and E2E-Only scenes each provide useful supervision, with the full E2E training set achieving the best held-out trajectory prediction (\secref{e2e-planning}). The dataset therefore provides a shared source of driving observations with task-specific evaluation protocols.

For context, the non-reactive log-replay score reported in our analysis of nuPlan Test14-Hard~\cite{plantf} is 85. That result concerns fleet-derived long-tailed scenes and does not establish the validity of our threshold. Comparisons across datasets also depend on their maps, driving situations, and scoring conditions.

\noindent\textbf{Publicly Available Reconstruction and Simulation Resources.} Replay-based selection uses publicly available, independently maintained references. Street imagery provides metric anchors for reconstruction and geo-references the trajectory. OpenStreetMap supplies the map against which the recovered log is replayed. Together with the simulator interface, these resources let us construct a closed-loop benchmark from dashcam videos without collecting new fleet logs. The value of recovered logs lies in expanding the driving situations available for learning and evaluation. Their use requires validation appropriate to the task, while establishing the reliability of a deployed driving system requires independent testing beyond any individual data source.

\subsection{Frequently Asked Questions}
\label{app:faq}

\noindent\textbf{Why not evaluate data quality with component-level metrics such as 3D box error, tracking error, depth error, and map alignment error?}  Component metrics need ground truth, which is unavailable for in-the-wild videos. Nevertheless, we report component metrics wherever ground truth exists. We evaluate pose and depth against nuScenes-mini ground truth, 3D object lifting on WorkZone3D~\cite{sural2026workzone3d}, and depth through its effect on novel-view synthesis. Component metrics also miss errors that matter, as each metric is computed in isolation. For example, a 3--9\,m global alignment error (\appref{pose-eval}) is invisible to the corresponding component-level metric yet is important for accurate simulation.

\noindent\textbf{Why use an additional selection criterion for closed-loop evaluation?} End-to-end prediction uses recorded observations and future trajectories. Privileged closed-loop evaluation also needs recovered agents, objects, and a routable map that support planner rollouts. Replay-based selection checks compatibility with our simulation setup, which is critical for ascertaining simulatability of our scenarios as being ``solvable''.

\noindent\textbf{What will be released, and under what license?} The source dataset~\cite{ghosh2025roadwork} is released under ODC-1.0. We use these videos as released, without further processing. We will release the videos, the generated 4D annotations, and the simulation code under a permissive open-source license. We do not redistribute any licensed data, following prior work~\cite{gadre2023datacomp, vuong2025aerialmegadepth}. The recovered annotations and task-specific protocols complement instrumented and synthetic self-driving datasets. They do not provide survey-grade ground truth.

\begin{figure}[!htbp]
    \centering
    \includegraphics[width=\linewidth]{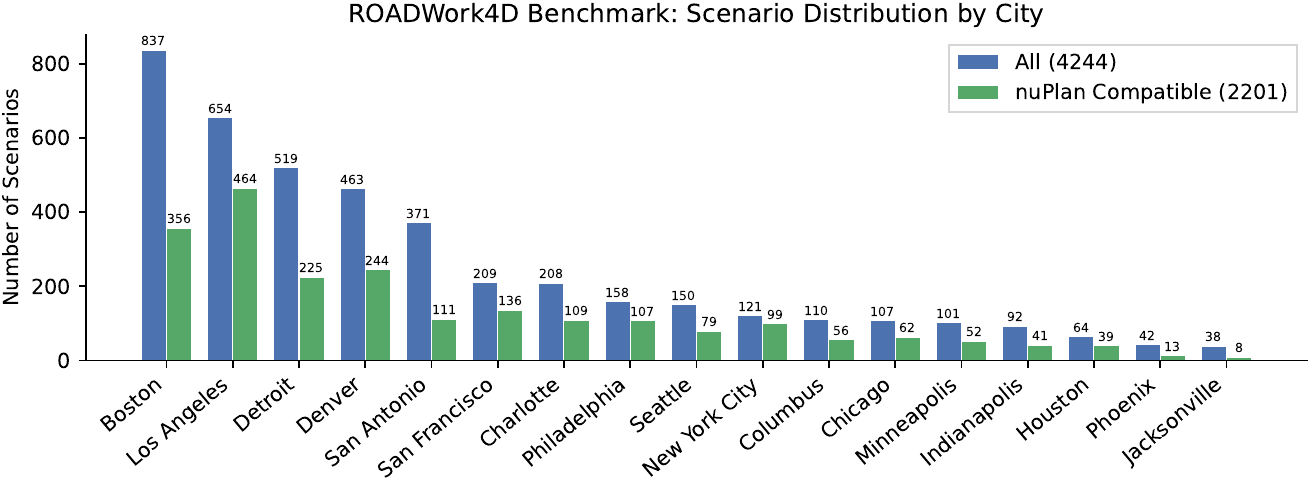}
    \caption{\textbf{\benchmarkname{} scenario distribution by city.} The 4{,}244 end-to-end scenes (blue) and 2{,}201 privileged closed-loop scenes (green), across 17 US cities. The two benchmarks have similar city distributions.}
    \label{fig:city-distribution}
\end{figure}
%


\FloatBarrier
\section{Additional Evaluation and Analysis}
\label{app:additional-eval}

\subsection{Privileged Closed-Loop Planning}
\label{app:cl-planning-details}

\begin{figure}[!htbp]
    \centering
    \includegraphics[width=0.85\linewidth]{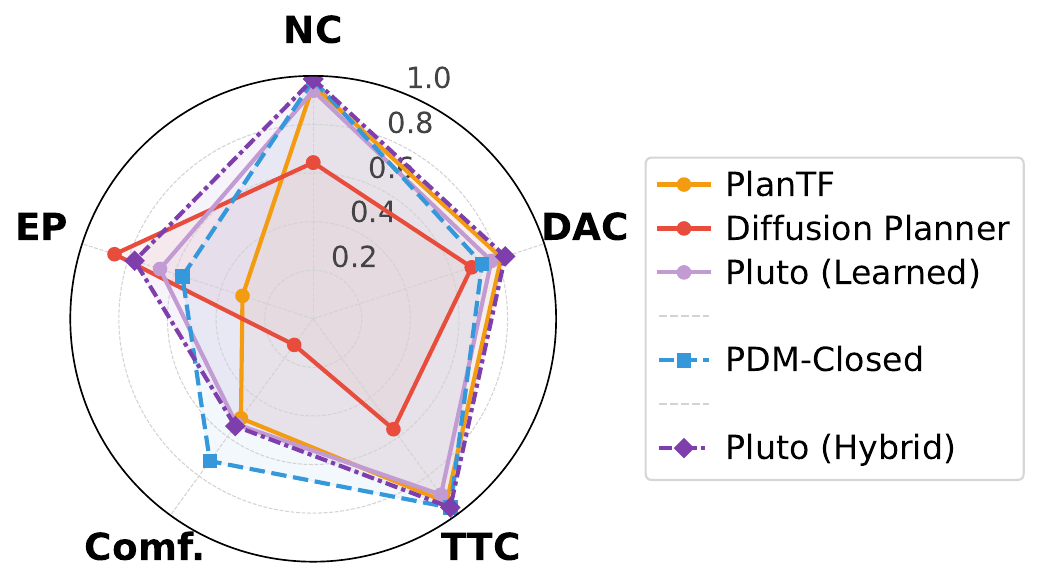}
    \caption{\textbf{CLS-R Breakdown.} Results in the reactive case closely match those in the non-reactive case.}
    \label{fig:cl-comparison-reactive-radar}
\end{figure}

\begin{table*}[t]
\centering
\begin{minipage}[t]{0.48\textwidth}
\vspace{0pt}
\centering
\captionof{table}{\textbf{Effect of channel sensitivity (CLS-NR).} Channel-sensitive evaluation penalizes drivable-area deviations from the GT route. \textit{Restated to aid the discussion here.}}
\label{tab:channel_sensitivity_supp}
\footnotesize
\setlength{\tabcolsep}{2pt}
\renewcommand{\arraystretch}{1.1}
\begin{tabularx}{\linewidth}{l*{4}{>{\centering\arraybackslash}X}}
\toprule
& \multicolumn{2}{c}{\shortstack{\textbf{Channel-}\\\textbf{Sensitive}}} & \multicolumn{2}{c}{\shortstack{\textbf{Channel-}\\\textbf{Insensitive}}} \\
\cmidrule(lr){2-3} \cmidrule(lr){4-5}
\textbf{Planner} & \shortstack{Score$\uparrow$\\\strut} & \shortstack{DAC\\Fail\%$\downarrow$} & \shortstack{Score$\uparrow$\\\strut} & \shortstack{DAC\\Fail\%$\downarrow$} \\
\midrule
\multicolumn{5}{l}{\textit{Learned}} \\
PlanTF~\cite{plantf}         & \worstcell{17.7} & \secondcell{18.0} & \worstcell{22.9} & 5.3 \\
Diffusion Planner~\cite{zheng2025diffusion}      & 24.0 & \worstcell{30.9} & 30.9 & 9.2 \\
Pluto~\cite{cheng2024pluto} & 32.7 & 23.4 & 43.0 & \secondcell{4.3} \\
\midrule
\multicolumn{5}{l}{\textit{Rule-Based}} \\
PDM-Closed     & \secondcell{53.0} & 27.3 & \secondcell{57.4} & \worstcell{9.4} \\
\midrule
\multicolumn{5}{l}{\textit{Hybrid}} \\
Pluto~\cite{cheng2024pluto}   & \bestcell{57.5} & \bestcell{17.8} & \bestcell{68.8} & \bestcell{1.2} \\
\bottomrule
\end{tabularx}
\end{minipage}%
\hfill
\begin{minipage}[t]{0.48\textwidth}
\vspace{0pt}
\centering
\captionof{table}{\textbf{Effect of Static objects (CLS-NR).} Scores are low even without traffic, which shows that construction zone layouts are the primary generalization challenge.}
\label{tab:dynamic_objects}
\footnotesize
\setlength{\tabcolsep}{2pt}
\renewcommand{\arraystretch}{1.1}
\begin{tabularx}{\linewidth}{l*{4}{>{\centering\arraybackslash}X}}
\toprule
& \multicolumn{2}{c}{\shortstack{\textbf{With}\\\textbf{Dynamics}}} & \multicolumn{2}{c}{\shortstack{\textbf{Objects}\\\textbf{Only}}} \\
\cmidrule(lr){2-3} \cmidrule(lr){4-5}
\textbf{Planner} & \shortstack{Score$\uparrow$\\\strut} & \shortstack{Coll.\%$\downarrow$\\\strut} & \shortstack{Score$\uparrow$\\\strut} & \shortstack{Coll.\%$\downarrow$\\\strut} \\
\midrule
\multicolumn{5}{l}{\textit{Learned}} \\
PlanTF~\cite{plantf}         & \worstcell{17.7} & \secondcell{4.3} & \worstcell{18.3} & 4.9 \\
Diffusion Planner~\cite{zheng2025diffusion}      & 24.0 & \worstcell{51.8} & 32.2 & \worstcell{35.7} \\
Pluto~\cite{cheng2024pluto} & 32.7 & 10.5 & 31.0 & 9.6 \\
\midrule
\multicolumn{5}{l}{\textit{Rule-Based}} \\
PDM-Closed     & \secondcell{53.0} & \bestcell{2.2} & \secondcell{56.4} & \bestcell{0.6} \\
\midrule
\multicolumn{5}{l}{\textit{Hybrid}} \\
Pluto~\cite{cheng2024pluto}   & \bestcell{57.5} & 4.7 & \bestcell{62.2} & \secondcell{0.8} \\
\bottomrule
\end{tabularx}
\end{minipage}
\end{table*}

\begin{figure*}[!tp]
    \centering
    \includegraphics[width=1\linewidth]{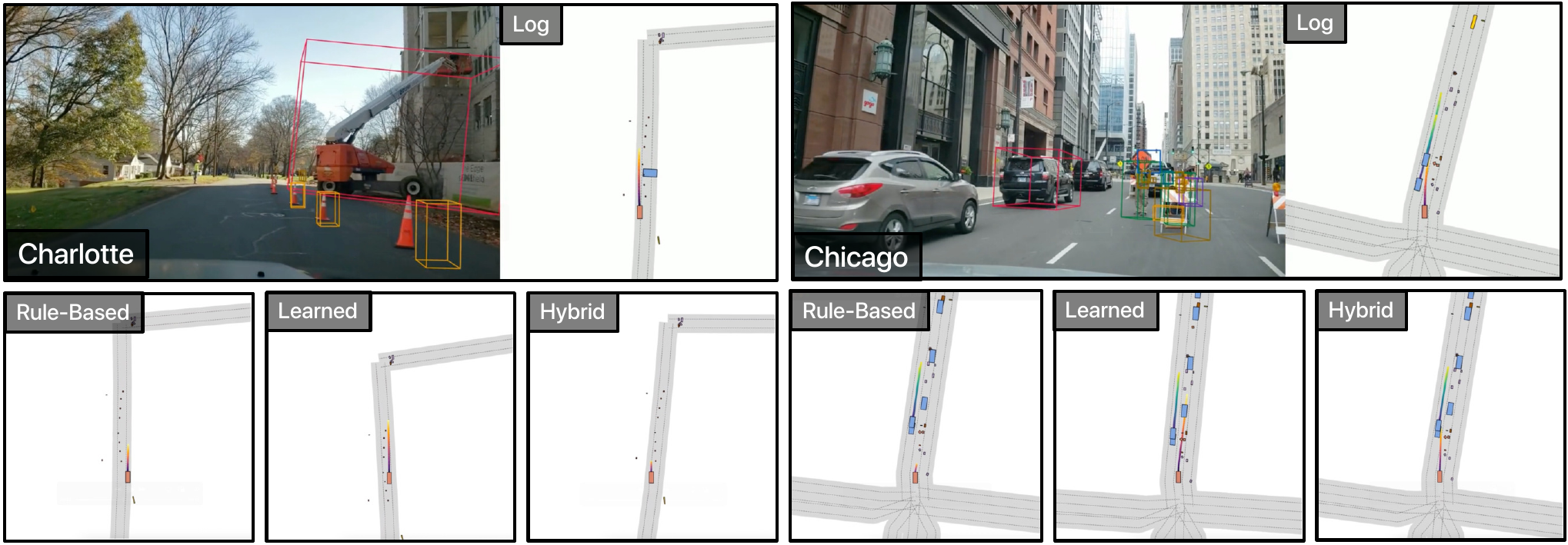}
    \caption{{\textbf{Qualitative Results on \clbenchmarkname{}.} We show the recovered log (top) and planner rollouts (bottom; trajectories colored by time) for rule-based PDM-Closed~\cite{dauner2023pdmclosed}, learned Diffusion Planner~\cite{zheng2025diffusion}, and hybrid Pluto~\cite{cheng2024pluto}. Lane channel sensitivity is required for safe work zone navigation.    In Charlotte, passing the boom lift requires a human-like excursion into the opposing lane: Diffusion Planner and Pluto go into the wrong lane, while PDM-Closed stalls behind the cones rather than leaving the lane. In Chicago, Pluto tracks the human route between queued vehicles and the work zone, PDM-Closed barely advances, and Diffusion Planner veers into the coned-off area toward a parked truck. \ifSHOWURL More examples can be viewed on our \href{https://www.cs.cmu.edu/~dash2sim/}{website}.\fi}}
    \label{fig:closed-loop-examples}
\end{figure*}

\begin{figure}[tbp]
    \centering
    \begin{minipage}{0.9\linewidth}
        \centering
        \includegraphics[width=\linewidth]{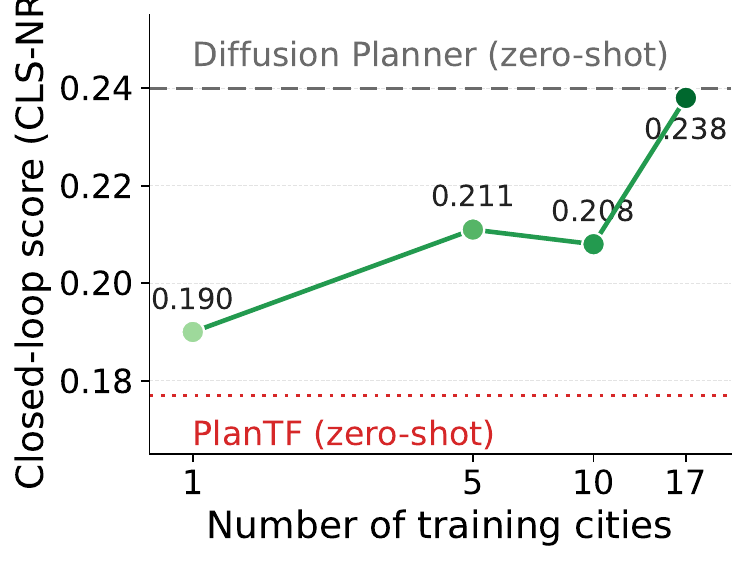}
    \end{minipage}\par\medskip
    \begin{minipage}{0.9\linewidth}
        \centering
        \includegraphics[width=\linewidth]{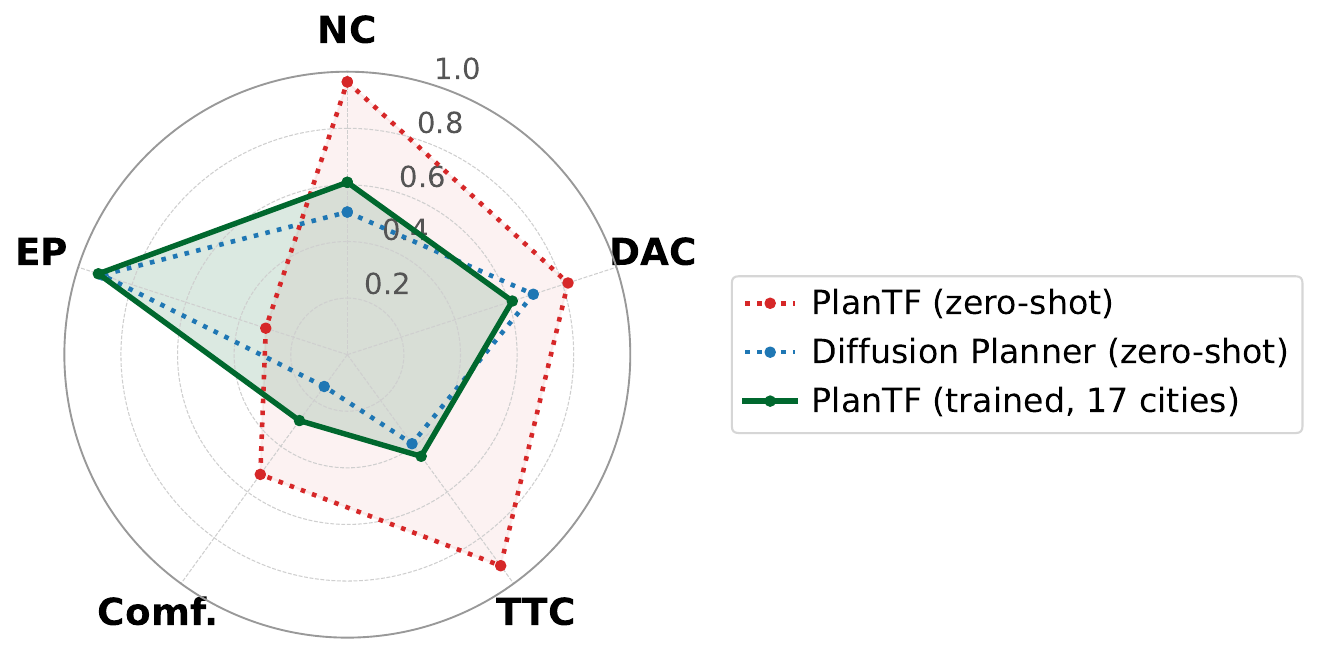}
    \end{minipage}
    \caption{\textbf{Closed-Loop Planning is Learnable from \clbenchmarkname{}.} We train PlanTF~\cite{plantf} on city-wise subsets of \clbenchmarkname{} and evaluate every model on the test set containing scenarios from all the cities. \textbf{(Top)} Closed-loop score increases with the number of training cities, with even data from one city improving performance over a zero-shot model and nearly reaching zero-shot Diffusion Planner~\cite{zheng2025diffusion} when trained on all cities. \textbf{(Bottom)} Zero-shot PlanTF (red) is conservative, with high no-collision (NC), drivable-area compliance (DAC), and time-to-collision (TTC) but little ego progress (EP), which is ultimately the point of driving. Training on all 17 cities (green) substantially improves ego progress at the expense of the other metrics, reaching the zero-shot Diffusion Planner~\cite{zheng2025diffusion} (blue).}
    \label{fig:plantf-data-scaling}
\end{figure}

\paragraph{Evaluation protocol.} We randomly select evaluation scenes from the closed-loop subset, with a sampling bias toward covering every city. We use the nuPlan simulator~\cite{nuplan} with the same evaluation metrics and aggregation mechanism as nuPlan~\cite{dauner2023pdmclosed}. The metrics are no-collision (NC), drivable-area compliance (DAC), time-to-collision (TTC), comfort, ego progress (EP), driving direction, and speed limit compliance. Each scenario uses a 15-second simulation window, matching nuPlan's closed-loop horizon. Our scenes are 30\,seconds long, we choose scenes starting at 5 seconds and ending at 20 seconds. One could create additional scenarios by sliding the temporal window, an exploration we leave to future work. Traffic light status is absent from our logs, and we map all objects and agents to appropriate nuPlan categories for compatibility. The routable map is derived from OpenStreetMap~\cite{openstreetmap} lane centerlines.

\noindent\textbf{Channel-sensitive scoring (definition).} We construct the drivable area as the union of polygons for lane segments matched to the recorded ego trajectory using spatial proximity and heading alignment, together with connecting lane segments. Channel-insensitive scoring treats all mapped lane segments as drivable. This distinction captures the role of temporary traffic control objects (such as cones, barricades, and vertical panels), which direct traffic into specific channels that may not align with mapped lanes. This helps us understand whether planners can perform the lane changes in work zone scenarios, rather than simply following any valid lane to the goal point. In the main paper, we report channel-sensitive scores and base our discussion on them, since they are more relevant for work zones, but we also report channel-insensitive scores to isolate the effect of channel sensitivity on planner performance. Channel-sensitive scoring has a limitation that is difficult to resolve: the human's route is the only route we can easily identify as safe, so an alternate route that respected the construction zone would still be penalized. We report channel-insensitive scores alongside the channel-sensitive ones partly for this reason. A critic judging whether a rollout respects the observed scene could provide an additional signal, which we leave to future work. \Figref{closed-loop-examples} shows qualitative rollouts where the rule-based, learned, and hybrid planners fail in complementary ways when a work zone requires deviating from, or keeping to, the human's channel.

\noindent\textbf{Role of Static Work Zone Objects.} We consider an additional simulation mode, where we analyze the effect of static work-zone objects by removing all dynamic agents from the simulation, leaving only the ego trajectory, static work-zone objects, and the map. This ablation primarily isolates the \emph{layout} axis of the long-tail challenges work zones present (\appref{workzones}). The results show that work-zone layouts and channel sensitivity, not dynamic agents, drive the difficulty: removing dynamic agents entirely yields only modest improvements in scores and collision rates (\Tabref{dynamic_objects}). Rule-based PDM-Closed~\cite{dauner2023pdmclosed} has dramatically fewer collisions but struggles to make progress, consistent with its known inability to change lanes~\cite{dauner2023pdmclosed, ghosh2026rad}.

\noindent\textbf{Closed-Loop Planning is Learnable from \benchmarkname{}.} Training data from even one city improves closed-loop performance on the full 17-city test set, which shows that planners can learn long-tailed driving behavior from 4D logs recovered from dashcams. We train PlanTF~\cite{plantf} on subsets of city logs of \benchmarkname{}, starting from a subset that contains logs from a single city up to the one that contains all of them. We evaluate every trained model on the same 17-city test set as earlier. Zero-shot PlanTF has low ego progress ($30$) despite high drivable-area compliance (\figref{plantf-data-scaling}). Training on a single city considerably increases ego progress at the expense of drivable-area compliance, and adding data from more cities then improves drivable-area compliance, reducing drivable-area failures by 16.2\%. The score matches zero-shot Diffusion Planner~\cite{zheng2025diffusion} (\figref{plantf-data-scaling}), which has shown much better performance on existing long-tailed fleet-derived planning datasets~\cite{zheng2025diffusion} when compared to PlanTF~\cite{plantf}.

\begin{figure}[tp]
    \centering
    \includegraphics[width=\linewidth,height=0.82\textheight,keepaspectratio]{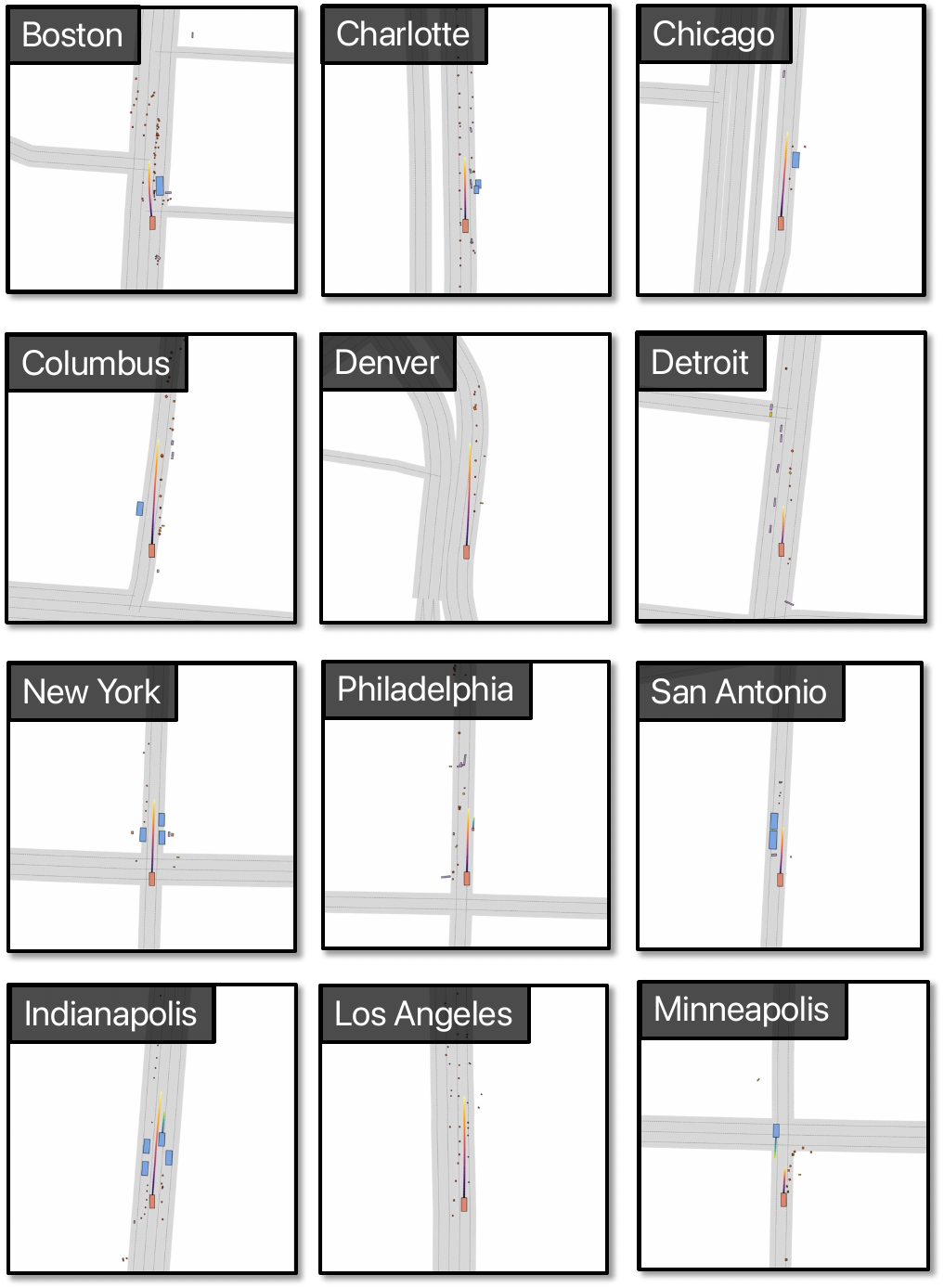}
\caption{\textbf{\projectname{} recovers a geographically diverse dataset of work-zone driving logs.} Bird's-eye views of logs reconstructed from dashcam videos, one each from twelve cities across the U.S. These logs can be used to train and evaluate motion planners in closed-loop simulation.}
\label{fig:cl-logs-cities}
\end{figure}
%


\FloatBarrier
\subsection{End-to-End Planning}
\label{app:e2e-planning}
\suppressfloats[t]

\begin{figure*}[tp]
\centering
\begin{minipage}[t]{0.48\textwidth}
\vspace{0pt}
\centering
    \includegraphics[width=\linewidth]{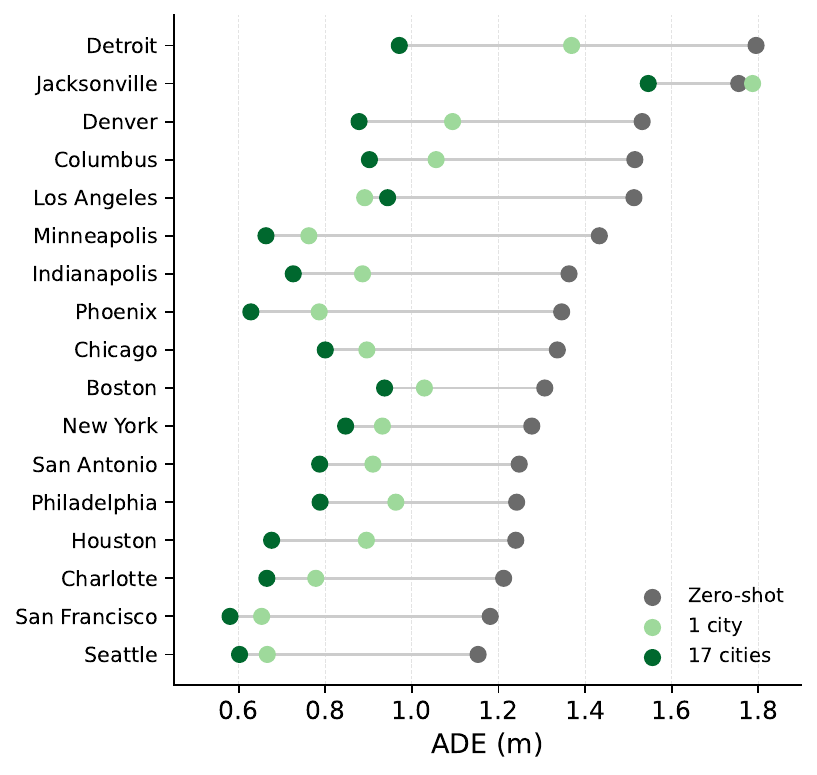}
    \caption{\textbf{Per-city ADE with increasing city diversity.} Each row is one test city, comparing zero-shot OneVL~\cite{lu2026onevl} with models fine-tuned on one city (Los Angeles) and on all 17 cities. Training on a single city improves performance in most test cities, including the unseen ones. The model fine-tuned on Los Angeles performs slightly better in Los Angeles than the model trained on all cities.}
    \label{fig:vla-city-dumbbell-cities}
\end{minipage}%
\hfill
\begin{minipage}[t]{0.48\textwidth}
\vspace{0pt}
\centering
    \resizebox{\linewidth}{!}{%
    \begin{tikzpicture}[x=1bp,y=1bp]
        \node[anchor=south west,inner sep=0] at (0,0)
            {\includegraphics{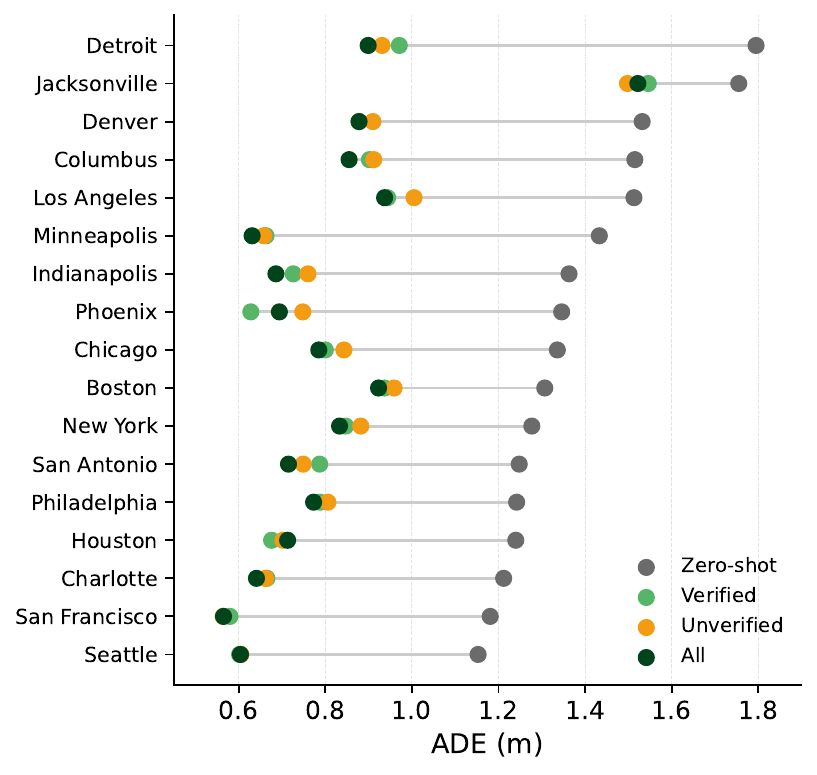}};
        \fill[white] (270,46) rectangle (390,112);
        \definecolor{clsubset}{rgb}{0.337254902,0.709803922,0.403921569}
        \definecolor{additionale2e}{rgb}{0.952941176,0.611764706,0.070588235}
        \definecolor{fulle2e}{rgb}{0,0.266666667,0.105882353}
        \foreach \y/\col/\txt in {100/gray/Zero-shot,85.6/clsubset/CL,71.2/additionale2e/E2E-Only,56.8/fulle2e/E2E} {
            \fill[\col] (277,\y) circle (4bp);
            \node[anchor=west,inner sep=0,font=\sffamily\fontsize{9}{11}\selectfont]
                at (287,\y) {\txt};
        }
    \end{tikzpicture}%
    }
    \caption{\textbf{Per-city ADE across training configurations.} Each row is one test city, comparing zero-shot OneVL~\cite{lu2026onevl} with models fine-tuned on CL, E2E-Only, and E2E training data (\Tabref{vla-e2e}). Training on E2E-Only improves over zero-shot in every city, while E2E performs best.}
    \label{fig:vla-city-dumbbell}
\end{minipage}
\end{figure*}

\noindent\textbf{Overview.} \secref{e2e-planning} showed that end-to-end planning is learnable from \benchmarkname{} and that the 3,732-scene E2E training set achieves the best prediction in our training-data comparison. This section details the setup and expands the analysis with per-city generalization and the behavior of zero-shot reasoning paradigms. Our end-to-end results correlate with the closed-loop findings of \secref{cl-planning}, while also showing the utility of our dataset and the complementarity of the two evaluations.

\paragraph{Details of our setup.} We compare the answer-only, Text-CoT, and Latent-CoT variants of OneVL~\cite{lu2026onevl}, pretrained on NAVSIM~\cite{navsim}, before and after LoRA fine-tuning (rank 16 for 2 epochs) on \eebenchmarkname{}. Each model predicts a 4\,s trajectory from the dashcam image and ego state. To test the value of scenes outside the closed-loop subset, we train the answer-only model on CL, E2E-Only, and their union (\Tabref{vla-e2e}). We evaluate all models on the same 512 held-out scenes shared with the closed-loop evaluation, sampling 16 frames per scene at 1\,fps. For geographic generalization, we use the same CL training subsets as the closed-loop experiments and evaluate across all 17 cities.

\paragraph{Generating Text-CoT Supervision.} We use the recovered 3D annotations to generate reasoning traces about work-zone navigation. For each training frame, we prompt Qwen3-VL-30B-A3B-Instruct with the image, ego state, nearby object descriptions, and a maneuver derived from the recovered future trajectory. We ask it to explain the maneuver using the work-zone layout, and use the resulting traces to fine-tune OneVL's reasoning variants~\cite{lu2026onevl}. The \hyperlink{cot-prompt}{prompt box} shows the annotation instructions and per-frame inputs.

\begin{promptbox*}[t]
\centering
\begingroup
\setlength{\fboxsep}{8pt}
\fbox{%
\begin{minipage}{\dimexpr\textwidth-2\fboxsep-2\fboxrule\relax}
\normalsize
\hypertarget{cot-prompt}{\textbf{Text-CoT annotation prompt}}
\par\medskip
\textbf{Instructions}
\par\smallskip
{\ttfamily\raggedright
You are an expert driving instructor annotating work-zone driving demonstrations.
You are given a front-camera image, a verified list of scene facts, the ego
state, and the expert driver's chosen plan. Treat the supplied plan as correct.
\par}
\medskip
\textbf{Reasoning (\texttt{think\_steps})}
\par\smallskip
{\ttfamily\raggedright
Write two to four first-person sentences justifying the plan.
\begin{itemize}
\setlength{\itemsep}{2pt}
\setlength{\parsep}{0pt}
\item Ground the reasoning only in the listed facts, and use the exact category names.
\item If any listed objects are within roughly fifteen meters, address the closest ones.
Do not claim that there is nothing to attend to.
\item Describe each object's position (left, right, or ahead) consistently with its listed bearing.
Explain how cones, drums, or barricades channel the drivable corridor when relevant.
\item Use qualitative descriptions, such as "close ahead on the right," without digits or brackets.
\item Use the provided closing sentence exactly as the final sentence.
\end{itemize}
}
\smallskip
\textbf{Additional observations (\texttt{extra\_observations})}
\par\smallskip
{\ttfamily\raggedright
Briefly note any safety-relevant objects or conditions that are clearly visible
in the image but absent from the listed facts, such as workers, flaggers,
machinery, or closed lanes. Most frames have nothing noteworthy, so leave this
field as an empty string unless something safety-relevant is clearly visible.
Do not repeat listed objects or describe scenery or generic road furniture.
\par}
\medskip
\textbf{Per-frame inputs}
\par\smallskip
{\ttfamily\raggedright
image: \{dashcam frame\}\par
scene\_facts: \{object counts, categories, distances, bearings, and motion\}\par
ego\_and\_command: speed \{speed\} m/s, \{motion trend\}; command \{command\}\par
expert\_plan: Over the next four seconds, the ego should \{speed\_action\} and \{lateral\_action\}.\par
closing\_sentence: Based on the understanding of the driving scene and the navigation information, the ego should \{speed\_action\} and \{lateral\_action\}.
\par}
\end{minipage}%
}
\endgroup
\par\smallskip
\begin{minipage}{\textwidth}
\small
The teacher returns \texttt{think\_steps} and \texttt{extra\_observations}. Braced fields are filled from each frame's recovered annotations. The speed and lateral actions summarize the recovered future ego trajectory.
\end{minipage}
\end{promptbox*}

\paragraph{Metrics.} We evaluate one predicted trajectory per input against the reference trajectory recovered by \projectname{}. Let $\hat{\mathbf{p}}_{i,t}$ and $\mathbf{p}_{i,t}\in\mathbb{R}^2$ denote predicted and reference positions for sample $i$ at time $0.5t$\,s, with $T=8$ waypoints over a 4\,s horizon. Positions use the current ego frame, with $x$ forward and $y$ left. We define $\mathbf{e}_{i,t}=\hat{\mathbf{p}}_{i,t}-\mathbf{p}_{i,t}$, and let $N$ be the number of valid samples.

\noindent\textbf{Average and final displacement errors (ADE and FDE).} ADE averages the Euclidean position error across all eight waypoints, while FDE measures the error at the final, 4\,s waypoint. We report both in meters,
\begin{align*}
    \mathrm{ADE} &= \frac{1}{NT}\sum_{i=1}^{N}\sum_{t=1}^{T}\lVert\mathbf{e}_{i,t}\rVert_2, \\
    \mathrm{FDE} &= \frac{1}{N}\sum_{i=1}^{N}\lVert\mathbf{e}_{i,T}\rVert_2.
\end{align*}
The per-horizon curves in \figref{vla-scaling} report $\frac{1}{N}\sum_i\lVert\mathbf{e}_{i,t}\rVert_2$ at each prediction time.

\noindent\textbf{Lateral and longitudinal errors.} We project each position error perpendicular and parallel to the reference travel direction at that waypoint. Let $\mathbf{u}_{i,t}$ be the normalized reference segment $\mathbf{p}_{i,t}-\mathbf{p}_{i,t-1}$, with $\mathbf{p}_{i,0}=(0,0)$ at the current ego position, and let $\mathbf{n}_{i,t}=(-u_{i,t,y},u_{i,t,x})$. The reported errors are mean absolute projections in meters,
\begin{align*}
    E_{\mathrm{lat}} &= \frac{1}{NT}\sum_{i=1}^{N}\sum_{t=1}^{T}|\mathbf{e}_{i,t}^{\mathsf{T}}\mathbf{n}_{i,t}|, \\
    E_{\mathrm{lon}} &= \frac{1}{NT}\sum_{i=1}^{N}\sum_{t=1}^{T}|\mathbf{e}_{i,t}^{\mathsf{T}}\mathbf{u}_{i,t}|.
\end{align*}

\noindent\textbf{Heading error.} We compare the predicted and reference headings $\hat{\theta}_{i,T}$ and $\theta_{i,T}$ at the final waypoint. The metric averages the absolute angular difference, wrapped to $[-\pi,\pi)$ and converted from radians to degrees:
\[
    E_{\mathrm{heading}} = \frac{180}{\pi N}\sum_{i=1}^{N}\bigl|\operatorname{wrap}_{[-\pi,\pi)}(\hat{\theta}_{i,T}-\theta_{i,T})\bigr|.
\]

\noindent\textbf{Miss rate at 2\,m (Miss@2m).} A prediction is a miss when its final displacement error exceeds 2\,m. We report the percentage of valid samples that satisfy this condition:
\[
    \mathrm{Miss@2m} = \frac{100}{N}\sum_{i=1}^{N}\mathbf{1}\!\left[\lVert\mathbf{e}_{i,T}\rVert_2 > 2\,\mathrm{m}\right].
\]
Here $\mathbf{1}[\cdot]$ is the indicator function.

\noindent\textbf{Zero-Shot Comparison of OneVL Variants.} The three OneVL variants all perform similarly, scoring between 1.390 and 1.547\,m ADE (\Tabref{vla-e2e}), and neither chain-of-thought variant outperforms direct answer prediction. The failure is interpretable. The text-CoT model repeatedly writes ``there are no objects in the current scene that I need to pay attention to'' while passing critical work zone objects. Work-zone scenarios are rare and underrepresented in the training data of these NAVSIM-trained models~\cite{navsim}. Additional reasoning alone provides limited gains, while fine-tuning on our work-zone data improves all three variants (\Tabref{vla-e2e}).

\noindent\textbf{Trace Augmentation for Text-CoT Fine-tuning.} Fine-tuning Text-CoT with a single distilled trace per frame reaches 0.879\,m ADE, while training over $\sim$5 traces per frame improves it to 0.824\,m, approaching answer-only fine-tuning (0.808\,m). \Tabref{vla-e2e} reports the augmented variant.

\noindent\textbf{Per-City Generalization.} The per-city breakdown (\figref{vla-city-dumbbell-cities}) shows that training on Los Angeles alone improves prediction in 15 of the other 16 cities.

\noindent\textbf{Broader Training Data Across Cities.} The model trained on the E2E-Only training scenes improves over zero-shot in all 17 test cities (\figref{vla-city-dumbbell}). In the aggregate comparison, the CL model achieves lower ADE than the E2E-Only model, while their union performs best (\Tabref{vla-e2e}). These results show that both training groups provide useful supervision for end-to-end learning.


\begin{figure*}[p]
\centering
\begin{minipage}{0.90\textwidth}
\input{figures/fig-vla-fails}
\end{minipage}
\par\medskip
\begin{minipage}{0.90\textwidth}
\input{figures/fig-vla-captions}
\end{minipage}
\end{figure*}

\clearpage
\subsection{Gaussian Splat Reconstruction}
\label{app:nvs-details}

\begin{table}[!htbp]
\centering
\small
\caption{\textbf{Full-split reconstruction quality} across \benchmarkname{} scenes (train + test frames).}
\label{tab:nvs-full}
\setlength{\tabcolsep}{5pt}
\renewcommand{\arraystretch}{1.15}
\begin{tabular}{lcc}
\toprule
Metric & OmniRe & OmniRe\,+\,Depth \\
\midrule
PSNR$\uparrow$ & \worstcell{26.14} & \bestcell{26.53\,{\scriptsize(+1.5\%)}} \\
SSIM$\uparrow$ & \worstcell{0.851} & \bestcell{0.870\,{\scriptsize(+2.2\%)}} \\
LPIPS$\downarrow$ & \worstcell{0.208} & \bestcell{0.170\,{\scriptsize(-18\%)}} \\
FID$\downarrow$ & \worstcell{49.3} & \bestcell{43.7\,{\scriptsize(-11\%)}} \\
cFID$\downarrow$ & \worstcell{7.2} & \bestcell{5.9\,{\scriptsize(-18\%)}} \\
DINO-S$\downarrow$ & \worstcell{0.071} & \bestcell{0.064\,{\scriptsize(-10\%)}} \\
DSim$\downarrow$ & \worstcell{0.069} & \bestcell{0.056\,{\scriptsize(-19\%)}} \\
\bottomrule
\end{tabular}
\end{table}

\begin{table*}[p]
\centering
\small
\setlength{\tabcolsep}{5pt}
\renewcommand{\arraystretch}{1.1}
\caption{\textbf{Rendering quality on long-tailed work-zone objects}
  across \benchmarkname{} scenes. Percentage improvement in parentheses.}
\label{tab:nvs-perclass}
\begin{tabular}{lrcccccc}
\toprule
  & & \multicolumn{2}{c}{PSNR$\uparrow$}
  & \multicolumn{2}{c}{SSIM$\uparrow$}
  & \multicolumn{2}{c}{LPIPS$\downarrow$} \\
\cmidrule(lr){3-4} \cmidrule(lr){5-6} \cmidrule(lr){7-8}
Class & $n$
  & OmniRe & +\,Depth
  & OmniRe & +\,Depth
  & OmniRe & +\,Depth \\
\midrule
Arrow board         &  229 & \worstcell{19.22} & \bestcell{19.47} & \worstcell{0.523} & \bestcell{0.534} & \worstcell{0.322} & \bestcell{0.299\,{\scriptsize(-7\%)}} \\
Barricade           &  942 & \worstcell{20.67} & \bestcell{21.12} & \worstcell{0.598} & \bestcell{0.620} & \worstcell{0.239} & \bestcell{0.210\,{\scriptsize(-12\%)}} \\
Barrier             &  828 & \worstcell{22.09} & \bestcell{22.45} & \worstcell{0.632} & \bestcell{0.645} & \worstcell{0.254} & \bestcell{0.224\,{\scriptsize(-12\%)}} \\
Cone                &  717 & \worstcell{21.97} & \bestcell{23.10} & \worstcell{0.659} & \bestcell{0.706} & \worstcell{0.242} & \bestcell{0.193\,{\scriptsize(-20\%)}} \\
Drum                &  718 & \worstcell{21.31} & \bestcell{21.84} & \worstcell{0.597} & \bestcell{0.619} & \worstcell{0.244} & \bestcell{0.211\,{\scriptsize(-14\%)}} \\
TTC message board   &  163 & \worstcell{20.25} & \bestcell{20.76} & \worstcell{0.515} & \bestcell{0.547} & \worstcell{0.292} & \bestcell{0.259\,{\scriptsize(-11\%)}} \\
TTC sign            & 1{,}063 & \worstcell{20.72} & \bestcell{21.13} & \worstcell{0.583} & \bestcell{0.609} & \worstcell{0.223} & \bestcell{0.200\,{\scriptsize(-10\%)}} \\
Tubular marker      &  793 & \worstcell{20.64} & \bestcell{21.31} & \worstcell{0.583} & \bestcell{0.615} & \worstcell{0.234} & \bestcell{0.202\,{\scriptsize(-14\%)}} \\
Vertical panel      &  381 & \worstcell{19.00} & \bestcell{19.84} & \worstcell{0.524} & \bestcell{0.567} & \worstcell{0.226} & \bestcell{0.196\,{\scriptsize(-13\%)}} \\
\midrule
\textit{Weighted mean} & 5{,}834 & \worstcell{20.94} & \bestcell{21.51} & \worstcell{0.595} & \bestcell{0.622} & \worstcell{0.242} & \bestcell{0.211\,{\scriptsize(-13\%)}} \\
\bottomrule
\end{tabular}
\end{table*}

\noindent\textbf{Overview.} Directly evaluating depth recovered from in-the-wild imagery is difficult because ground-truth depth is unavailable. Following prior work~\cite{vuong2025aerialmegadepth, chen2025omnire}, we evaluate the depth produced by \projectname{} through novel-view synthesis, isolating its contribution to rendering held-out views. \secref{nvs} showed that depth supervision from \projectname{} improves held-out novel-view quality on every metric. This section provides the full-split and per-class breakdown.

\paragraph{Details of our setup.} We randomly sample 100 scenes from the 512-scene evaluation set, limiting the sample size for computational reasons. We use per-frame depth supervision to account for dynamic scene content. Every tenth frame is held out for novel-view evaluation, and both its RGB image and depth map are excluded from Gaussian-splat training. Following OmniRe~\cite{chen2025omnire}, we report PSNR and SSIM. These metrics average over the entire frame and are therefore insensitive to long-tailed objects that occupy only a small fraction of the image. Moreover, they penalize low-level misalignment that minimally affects the perceptual quality of the rendered scene~\cite{zhang2018unreasonable, ghildyal2022shift}. We additionally report LPIPS~\cite{zhang2018unreasonable}, FID~\cite{heusel2017gans}, CLIP-FID~\cite{kynkaanniemi2022role} (cFID), DINO-Struct~\cite{tumanyan2022splicing} and DreamSim~\cite{fu2023dreamsim} (DSim), which are increasingly adopted in driving novel view synthesis evaluations~\cite{wickrema2025benchmarking, khan2025autosplat, cao2025pseudo, wang2026sensor2sensor} and capture higher-level visual fidelity.

\noindent\textbf{Full-split reconstruction quality.}
\Tabref{nvs-full} reports metrics on the full split. Depth supervision improves all metrics, including on training views where the photometric loss already provides direct pixel-level supervision: LPIPS improves by 18\% and DreamSim by 19\%. This shows that the depth regularizes the Gaussian representation itself. As splat positions and shapes are better constrained by the available depth, the representation generalizes to both seen and unseen views. Monocular video is the most difficult setting for OmniRe~\cite{chen2025omnire}, which is primarily evaluated on multi-modal fleet data. This partially explains the drop in absolute scores from full split to held-out split.

While a direct comparison is difficult due to differences in data source and sensor setup, the PSNR from monocular logs is comparable to that reported by recent fleet-based sensor simulation methods~\cite{zhou2025hugsim, cao2025pseudo, chen2025omnire}, suggesting that dashcam-derived driving logs may become sufficient in the future for realistic sensor simulation with improvements in rendering.

\noindent\textbf{Rendering of small long-tail objects.}
\Tabref{nvs-perclass} evaluates rendering quality on 9
long-tailed work-zone object categories across 5{,}834 crop-level
instances in the held-out test split.
We compute per-class PSNR and SSIM over the corresponding class masks, and LPIPS on bounding-box crops enclosing each class mask in a frame.
These objects span only a few pixels in the dashcam image
(\secref{reconstruction}) and the ego vehicle observes each one from a
narrow range of directions over a short temporal window. 

LPIPS~\cite{zhang2018unreasonable} improves for every long-tailed category, with the largest improvements on thin, vertical objects like cones ($20\%$), tubular markers ($14\%$), and vertical panels ($13\%$). This shows that the dense depth from \secref{reconstruction} improves object-level rendering of small work-zone objects, not just the overall scene.

\begin{figure*}[p]
    \centering
    \includegraphics[width=0.9\linewidth]{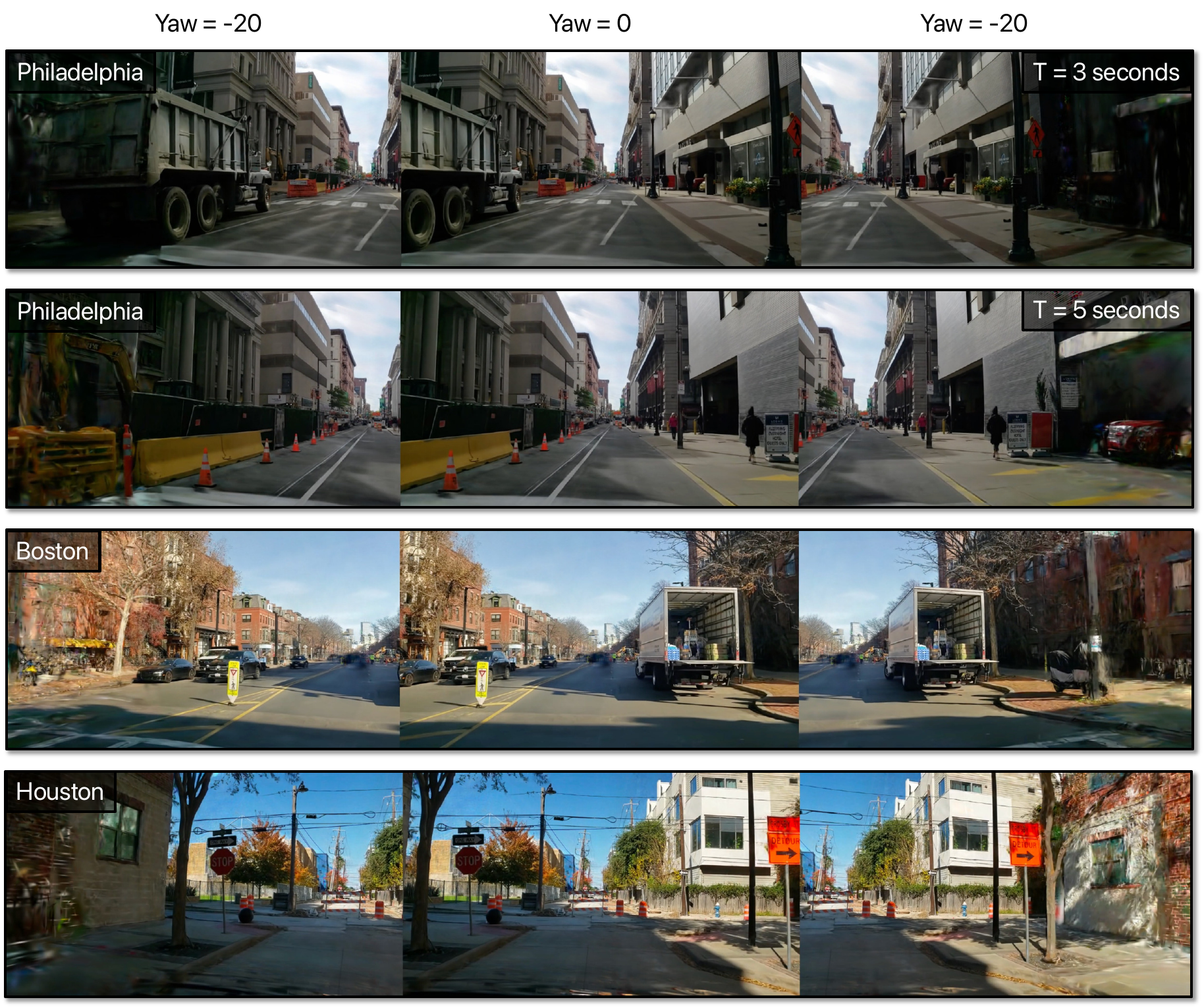}
    \caption{\textbf{Rendering Additional Camera Viewpoints.} Qualitative results of rendering additional camera viewpoints (yaw of $\pm20^\circ$) from a single monocular dashcam video, conditioned on the dense depth recovered by \projectname{}. A forward-facing dashcam leaves much of the scene unobserved; these renderings suggest that this depth could condition closed-loop sensor simulation beyond the original viewpoint.}
    \label{fig:multicam-initial}
\end{figure*}

\noindent\textbf{Rendering Additional Camera Viewpoints.} We present some initial qualitative results of rendering additional camera viewpoints from monocular videos in \figref{multicam-initial}.

\subsection{Application to Other Dashcam Datasets}
\label{app:cross-dataset}

\begin{figure*}[!t]
    \centering
    \includegraphics[width=\confweb{1.0}{1.0}\linewidth]{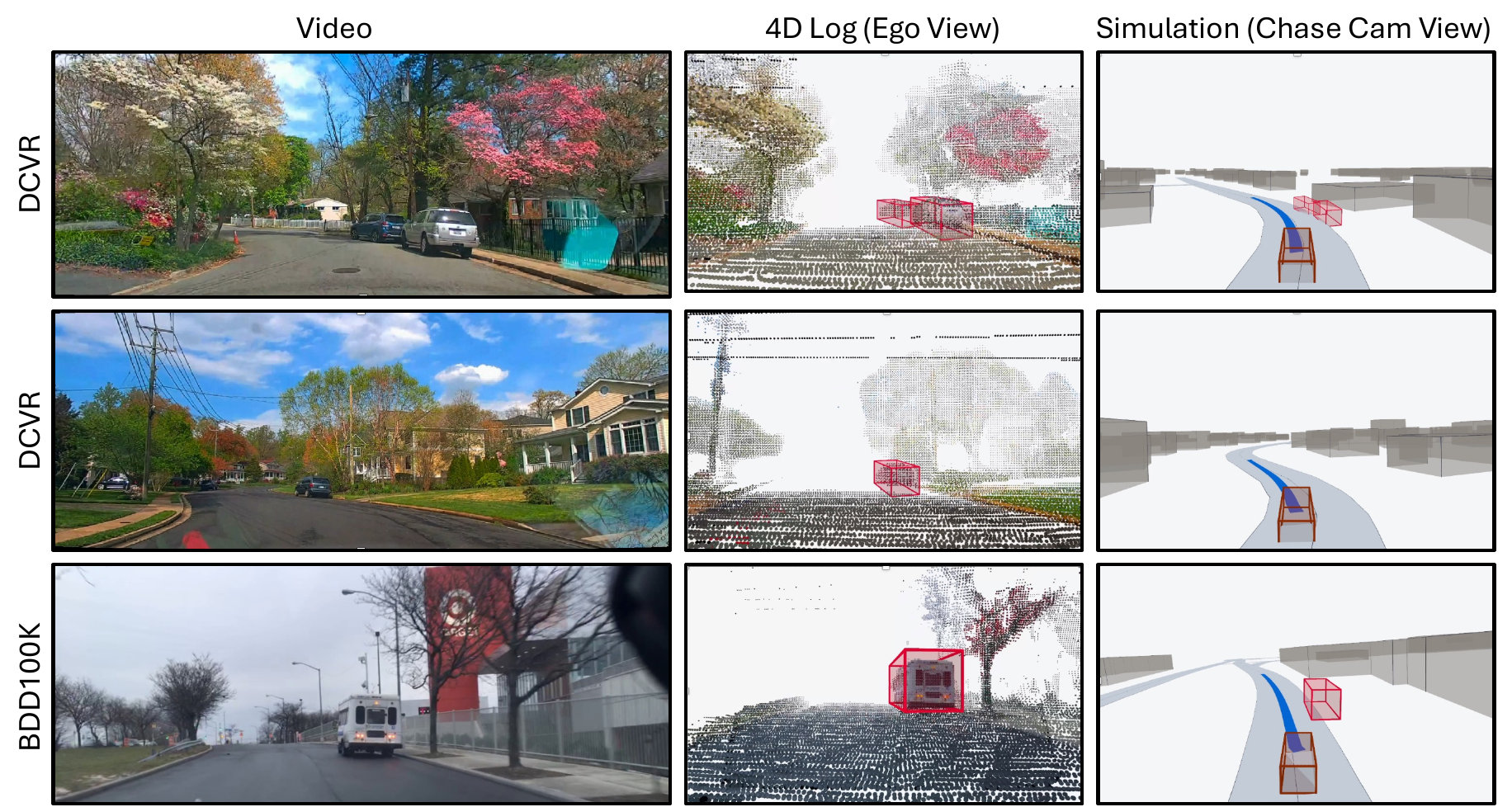}
    \vspace{-1em}
    \caption{{\textbf{Applying \projectname{} on other dashcam video datasets.} We apply \projectname{} zero-shot to videos from DCVR~\cite{wang2024dc} and BDD100K~\cite{yu2020bdd100k} datasets. This suggests the potential of employing dashcam videos for scaling long-tailed data.}}
    \vspace{-1em}
    \label{fig:zero-shot-dcvr-bdd100k}
\end{figure*}

We apply \projectname{} zero-shot to videos from DCVR~\cite{wang2024dc} and BDD100K~\cite{yu2020bdd100k}, recovering 4D logs for simulation. \Figref{zero-shot-dcvr-bdd100k} shows the source videos, recovered logs, and corresponding simulation scenes.

\clearpage
\subsection{Camera-Controlled Generative Synthesis}
\label{app:generative-conditioning}

\begin{figure}[t]
    \centering
    \IfFileExists{assets/fig-generative-conditioning-how-it-works.pdf}{%
        \includegraphics[width=\linewidth]{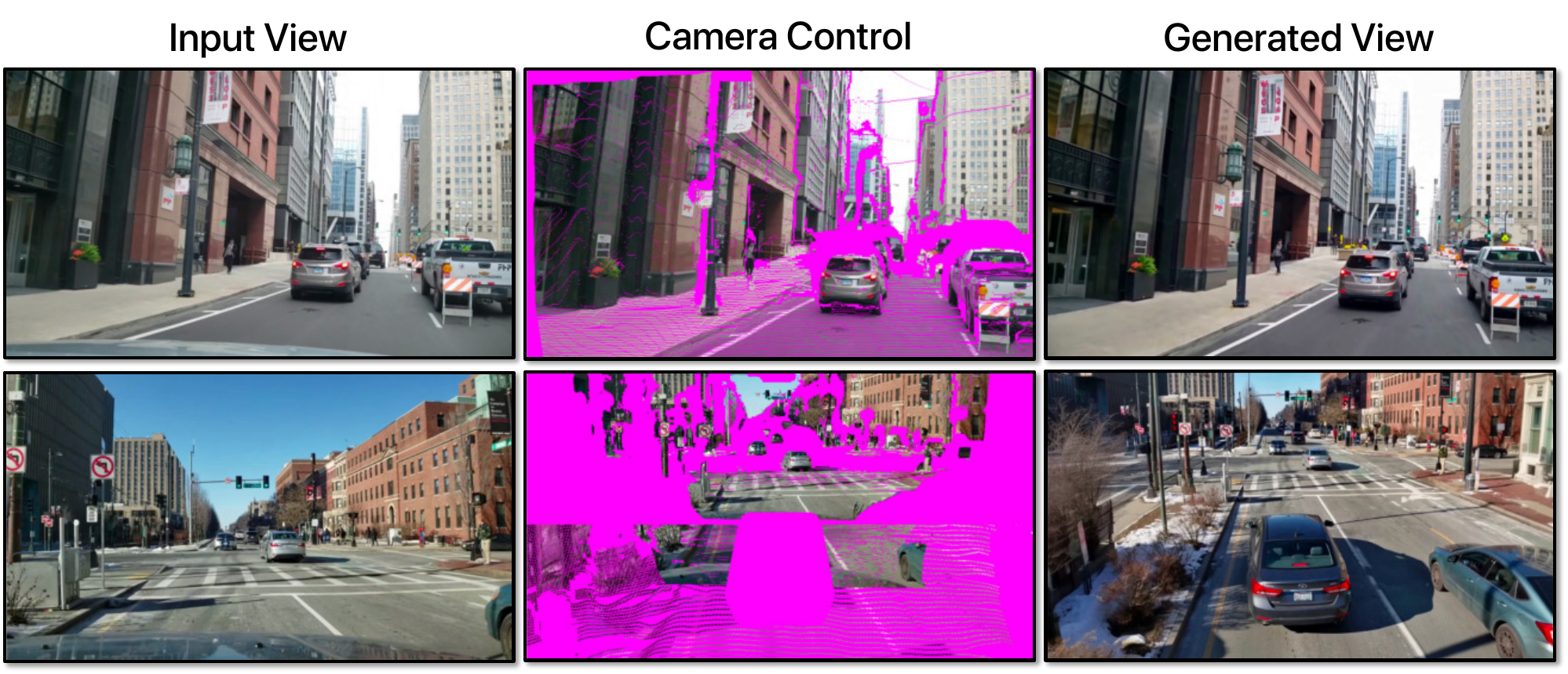}%
    }{%
        \fbox{\parbox[c][0.22\textheight][c]{0.95\linewidth}{\centering
            Conditioning overview figure\\[1em]
            \texttt{fig-generative-conditioning-how-it-works.pdf}}}%
    }
    \caption{\textbf{Conditioning for Camera-Controlled Generative Synthesis.} \textbf{(Left)} Input dashcam frames. \textbf{(Middle)} Recovered point clouds rendered at the requested camera poses, with missing regions in magenta. \textbf{(Right)} Generated views conditioned on the source videos and renderings. The examples show a lateral camera shift (top) and an elevated chase-camera view (bottom).}
    \label{fig:fig-generative-conditioning-how-it-works}
\end{figure}

\begin{figure*}[t]
    \centering
    \IfFileExists{assets/fig-generative-conditioning-more-cool.pdf}{%
        \includegraphics[width=\linewidth]{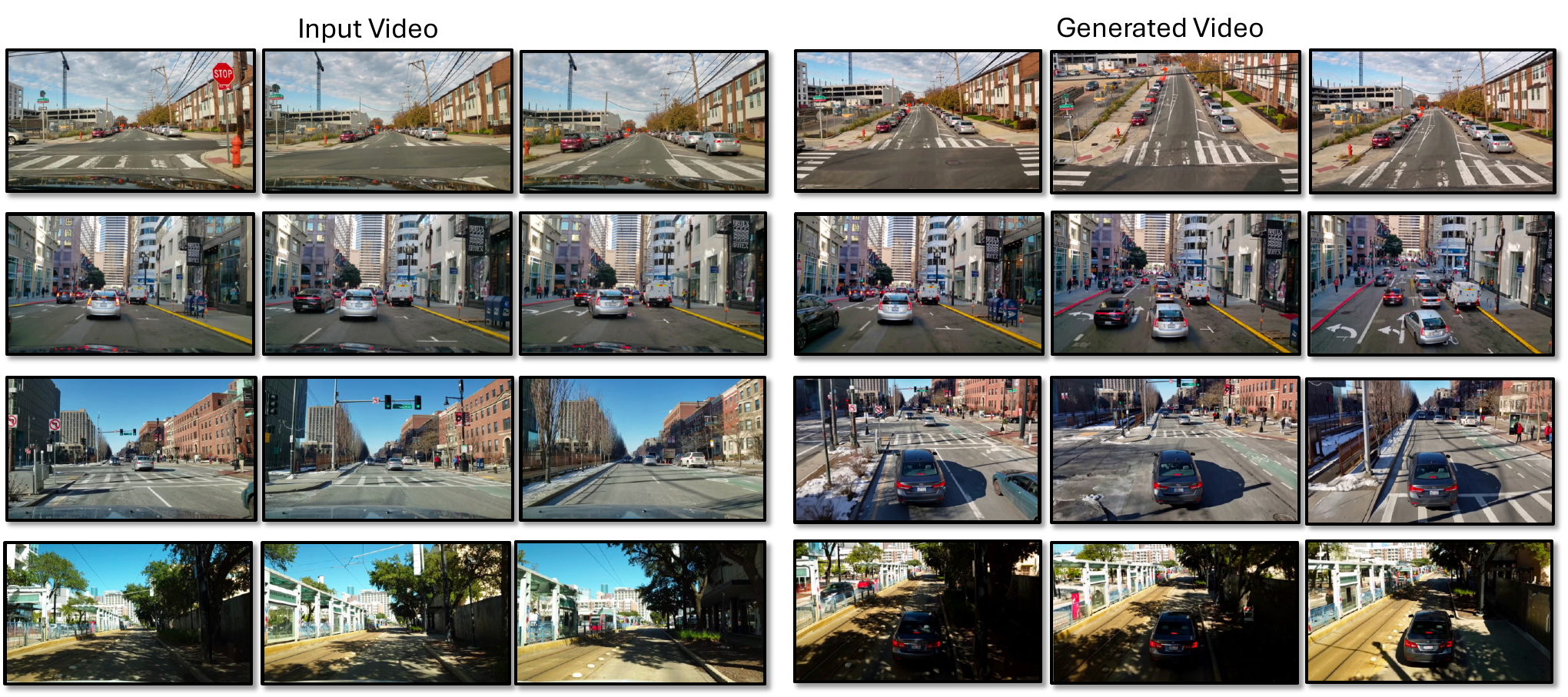}%
    }{%
        \fbox{\parbox[c][0.38\textheight][c]{0.95\linewidth}{\centering
            Additional camera-controlled synthesis examples\\[1em]
            \texttt{fig-generative-conditioning-more-cool.pdf}}}%
    }
    \caption{\textbf{Additional Qualitativate Results on Camera-Controlled Generative Synthesis.} Each row shows three frames from the input video (left) and generated video (right). The top two rows show changes to the camera trajectory, including elevation changes. The bottom two rows show chase-camera views behind a vehicle. \textbf{Best viewed zoomed in.} \textit{Supplemental material contains additional media and visualizations.}}
    \label{fig:fig-generative-conditioning-more-cool}
\end{figure*}

The depth and camera poses in \benchmarkname{} let us condition video generation on a new camera trajectory (\secref{generative-simulation}). \Figref{generative-conditioning-control} and \figref{fig-generative-conditioning-more-cool} show ego action control with modified camera trajectories, including shifts into an adjacent lane, agent view synthesis from another vehicle's perspective, and chase-camera view synthesis. For chase-camera synthesis, we use the recorded ego-camera trajectory as the target and render the source from a chase-camera viewpoint.

We render the recovered RGB point clouds along the requested trajectory and use a conditional generative video model, MiniMax H3~\cite{minimax2026h3} to complete the missing regions. The model receives the rendered video together with the synchronized source video (\figref{fig-generative-conditioning-how-it-works}).

\paragraph{Rendering the conditioning video.}
We render a partial RGB view at each target pose using the recovered depth. For an undistorted source pixel $\tilde{\mathbf{u}}=(u,v,1)^\top$ with valid depth $d_t(\mathbf{u})$ and source camera intrinsics $\mathbf{K}_s$, we back-project into the source camera frame as
\[
    \mathbf{x}_s = d_t(\mathbf{u})\,\mathbf{K}_s^{-1}\tilde{\mathbf{u}}.
\]
We transform $\mathbf{x}_s$ into the target camera frame using the relative pose at time $t$, project it with the target intrinsics, and assign it the source pixel's RGB value. When multiple points project to the same target pixel, we retain the closest point with positive depth. We color pixels without a projected observation magenta to mark regions that require completion.

\paragraph{Fine-tuning.}
We initialize a rank-32 LoRA from the MiniMax H3 Cross-View adapter\footnote{\url{https://huggingface.co/Cseti/MiniMax-H3_Ref2VA-LoRA-CrossView-Warp_v1}} and fine-tune it for 2 epochs with AdamW at a learning rate of $10^{-5}$. Training uses 200 \benchmarkname{} samples and 25 paired-view Syn4D~\cite{jiang2026syn4d} samples (sampled twice).

%


\clearpage
\section{\projectname{} Framework}
\label{app:framework-details}

\begin{figure*}[t!]
    \centering
    \includegraphics[width=1.0\linewidth]{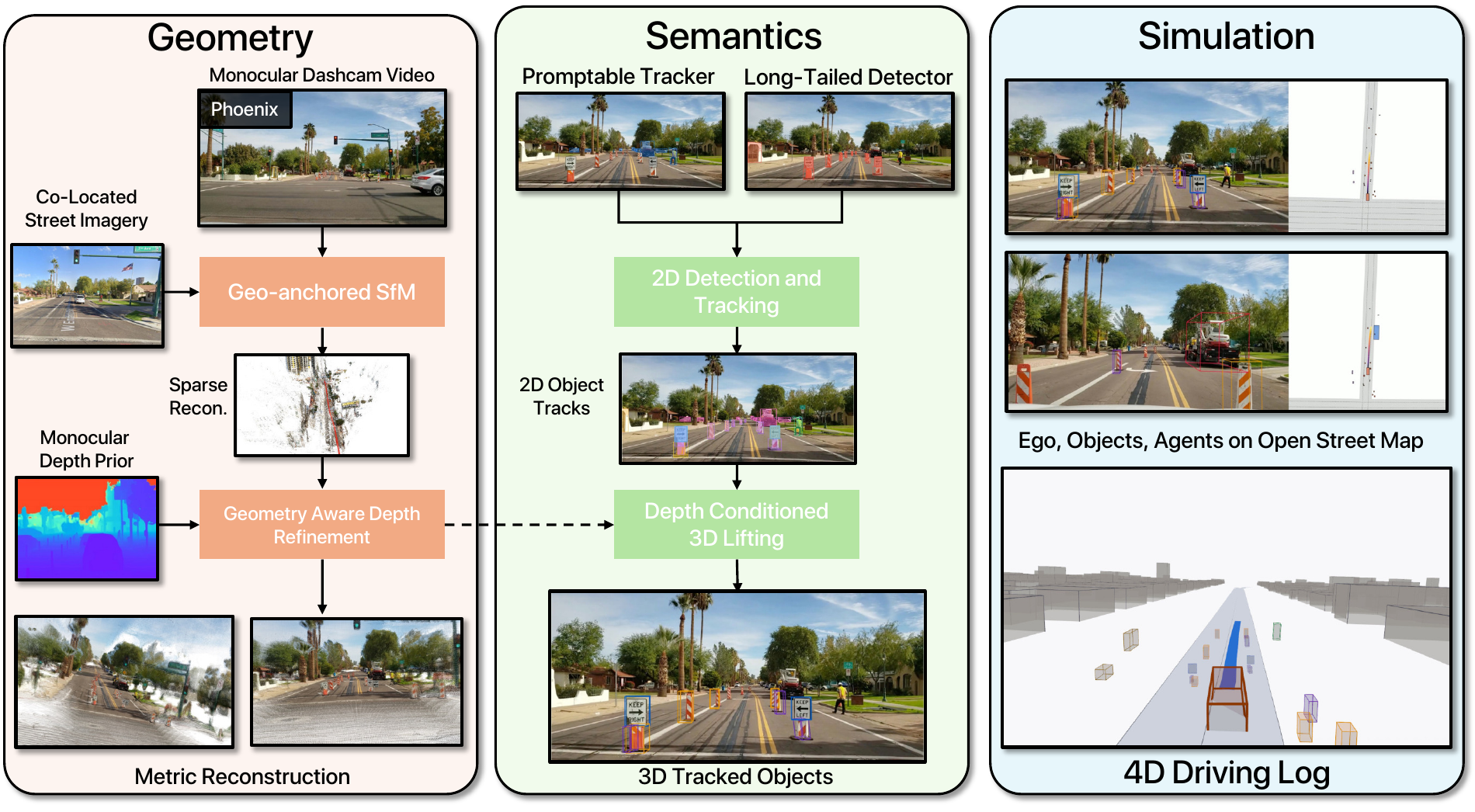}
\caption{\textbf{\projectname{} Framework.} \projectname{} produces a metric, geo-referenced 4D driving log from in-the-wild monocular dashcam video in three stages, shown left to right. \emph{Geometry} recovers a metric ego trajectory and dense per-image depth. \emph{Semantics} detects, tracks, and lifts agents and long-tailed objects to 3D using promptable foundation models~\cite{carion2025sam,huang2026wilddet3d}. For \textit{Simulation}, we retrieve the OpenStreetMap~\cite{openstreetmap} tile and place the ego, agents, and objects.}
\label{fig:pipeline}
\end{figure*}

\begin{figure*}[!tp]
    \centering
    \includegraphics[width=1\linewidth]{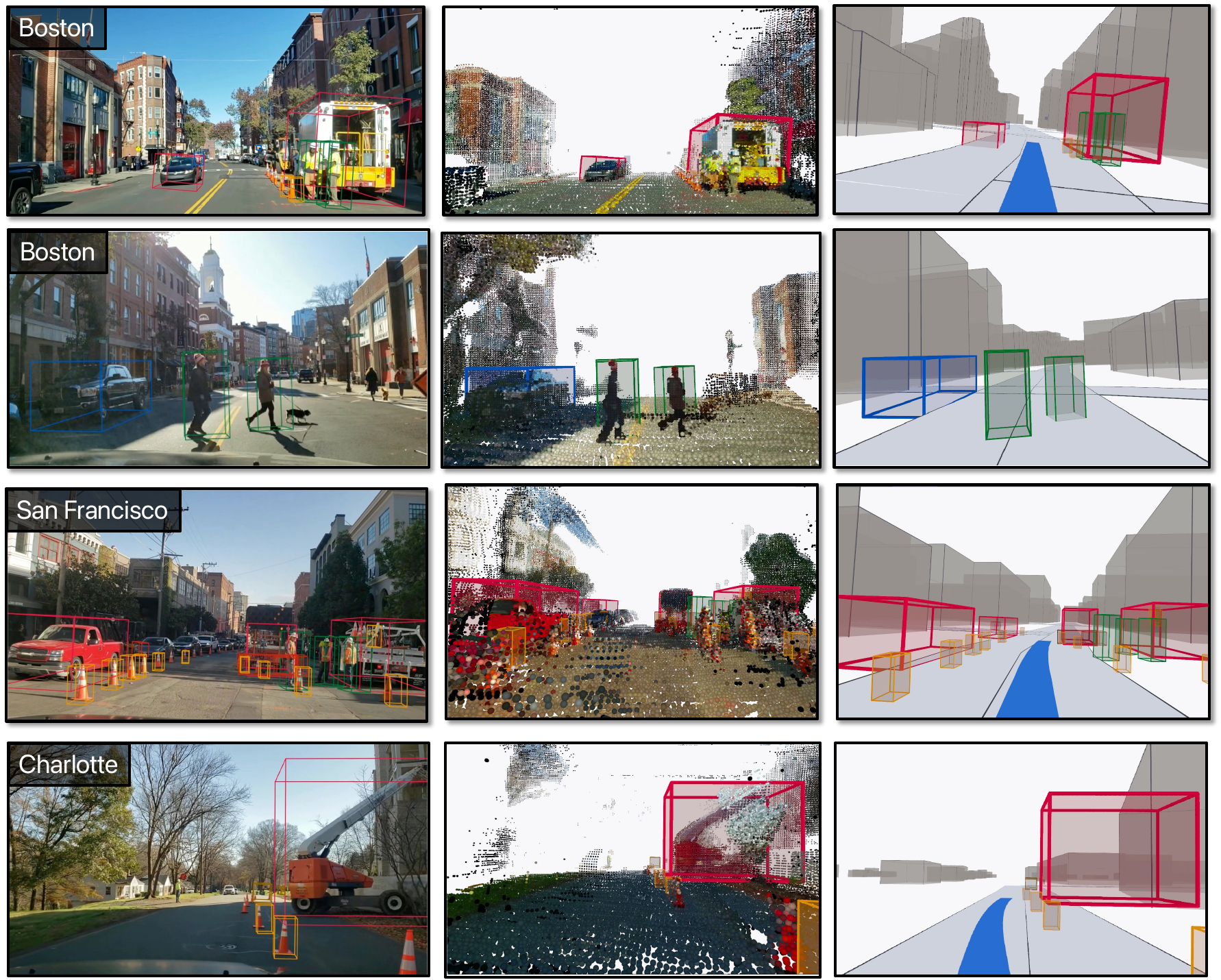}
    \caption{{\textbf{More Driving logs from \benchmarkname{} recovered by \projectname{}.} Our logs include long-tailed work zone scenarios with rare objects and dynamic agents. Our supplemental website and walkthrough video contain additional media and visualizations.}}
    \label{fig:cl-logs-cities-supp}
\end{figure*}

\subsection{Recovering Metric Geometry}
\label{app:geometry-blocks}

Several issues make metric reconstruction from dashcam video challenging. Monocular localization is scale-ambiguous, as the recovered trajectory and scene are determined only up to a global similarity transform~\cite{hartley2003multiple}. The dashcam's GPS, if available, routinely places the ego vehicle in adjacent city blocks instead of the driven lane, and most in-the-wild video lacks GPS entirely. Long-tailed objects (e.g., cones, tubular markers) also span only a few pixels in the dashcam image, so feature triangulation produces too few 3D points on them to support downstream 3D lifting and novel view synthesis, and, potentially, closed-loop sensor simulation. \projectname{}'s \emph{geometric} framework (\figref{pipeline}, left) addresses all three.

\noindent\textbf{Sparse reconstruction.} We compute image-level descriptors with EigenPlaces~\cite{berton2023eigenplaces} and retrieve nearest neighbors with FAISS~\cite{douze2024faiss} to build the scene graph, then extract local features with SuperPoint~\cite{detone2018superpoint} and match image pairs with LightGlue~\cite{lindenberger2023lightglue}. We run Global SfM~\cite{pan2024global,schonberger2016structure} jointly over the dashcam and Street View streams. The resulting reconstruction is aligned to global coordinates via a similarity transform fit to the Street View poses. We discard the noisy dashcam GPS (\figref{ego-trajectories}, red) during this alignment to avoid incorporating its biases. We fit a ground plane using a road-surface mask from a semantic segmentation model~\cite{cheng2022masked} for downstream simulator compatibility. Sparse reconstruction did not converge for 131 of the 4,375 source videos, leaving 4,244 recovered scenes. Further tuning of the reconstruction pipeline may recover some of these cases.

\noindent\textbf{Geometry-aware depth refinement.} The joint SfM above is sparse and produces few 3D points. As long-tailed objects are small, sparse depth is insufficient for accurate localization~\cite{sural2026workzone3d} and view synthesis. We refine per-image depth with MP-SfM~\cite{pataki2025mpsfm}, which fuses a monocular depth-and-normal prior~\cite{hu2024metric3d} with multi-view geometric constraints. We run this stage on the dashcam frames alone, excluding co-located street imagery since the two captures occurred at different times and disagree on dynamic and transient scene content. The output is per-image, metric, geometrically consistent depth. Feedforward depth estimators alone~\cite{hu2024metric3d, wang2025vggt,keethamapanything} did not produce sufficient accuracy in our experiments, but might improve performance when combined with the sparse reconstruction step~\cite{pan2026gluemap}, which we leave to future work.

\subsection{Geometry Evaluation: Setup and Limitations}
\label{app:pose-eval}

\noindent\textbf{Overview.} Our dashcam videos lack ground-truth ego poses and LiDAR. We therefore use daytime nuScenes-mini~\cite{nuscenes} to evaluate recovered trajectories against reference poses and sparse geometry against LiDAR. Several evaluated sequences are captured in Singapore.

\noindent\textbf{Pose Evaluation Setup.}
For each scene, \projectname{} retrieves nearby Google Street View
panoramas using the ego vehicle's coarse GPS, projects them into perspective crops, and reconstructs a sparse model anchored to the panoramas' georeferenced coordinates. We evaluate the resulting dashcam camera poses against nuScenes ego-pose ground truth following the standard SLAM or odometry benchmark
protocol~\cite{grupp2017evo}: we estimate the $\mathrm{Sim}(3)$ transform that best aligns the predicted trajectory to the ground truth via Umeyama alignment~\cite{umeyama1991least} and report the recovered scale ratio and absolute trajectory error (ATE) after alignment. Relative pose error (RPE) is computed without correcting the recovered metric scale, so it remains sensitive to scale errors removed by the ATE alignment.

\noindent\textbf{Role of Street Imagery.} We isolate the contributions of metric anchoring and additional views with two reconstruction variants. \emph{Coarse GPS} keeps Street View imagery as part of the SfM but replaces the panorama alignment anchors with simulated consumer-grade GPS positions along the ego trajectory, perturbed with iid Gaussian noise ($\sigma_h{=}10$\,m, $\sigma_v{=}15$\,m) modeling multipath-dominated error in urban areas. \emph{No street imagery} removes Street View imagery entirely, running SfM on monocular frames alone with the same coarse GPS anchors. \Tabref{pose-eval} in the main paper reports median results. For the variant without street imagery, we compute metrics over reconstructed scenes only.

\begin{table}[t]
\centering
\caption{\textbf{Sparse geometry agreement with LiDAR.} After trajectory alignment, our sparse reconstruction has a median surface error of 26.2\,cm for points within 20\,m. We report medians of per-scene statistics over six daytime nuScenes-mini scenes, using points with LiDAR support.}
\label{tab:sparse-lidar}
\footnotesize
\setlength{\tabcolsep}{4pt}
\renewcommand{\arraystretch}{1.12}
\begin{tabularx}{\linewidth}{l*{3}{>{\centering\arraybackslash}X}}
\toprule
\shortstack{Maximum\\viewing depth} &
\shortstack{Median error\\(cm)} &
\shortstack{Accuracy\\$\leq 50$\,cm} &
\shortstack{Accuracy\\$\leq 1$\,m} \\
\midrule
10\,m & 25.4 & 89.2\% & 92.4\% \\
20\,m & 26.2 & 75.2\% & 90.0\% \\
30\,m & 31.4 & 67.7\% & 89.4\% \\
40\,m & 34.1 & 67.2\% & 88.4\% \\
\bottomrule
\end{tabularx}
\end{table}

\paragraph{LiDAR Agreement with Sparse Geometry.}
\label{app:sparse-lidar}
We compare our sparse reconstruction depths with LiDAR depth on the nuScenes-mini dataset. We align the reconstructed trajectory to the ground-truth ego poses using a $\mathrm{Sim}(3)$ transform and apply it to the reconstructed points, without further alignment to LiDAR. We transform LiDAR returns into each camera frame using sensor calibration and ego poses at the sensor timestamps. We use Mask2Former~\cite{cheng2022masked} to exclude sky and a five-pixel margin around semantic boundaries. We evaluate reconstructed points observed in the monocular sequence only when a LiDAR return projects within two pixels and lies in the same connected semantic region. For each point, we measure the 3D distance to the nearest LiDAR return of the same semantic class, considering only predicted and reference depths below 40\,m.

\Tabref{sparse-lidar} reports one-way nearest-neighbor distances from the aligned sparse reconstruction to LiDAR. We count each point once and define its viewing depth by the median matched LiDAR depth across its observations. Within 20\,m, the median per-scene surface error is 26.2\,cm and the median per-scene accuracy within 50\,cm is 75.2\%. These results show sub-meter agreement with LiDAR for short-range, supported sparse geometry.

\noindent\textbf{Limitations of the LiDAR Evaluation.} This evaluation covers the LiDAR-supported subset and does not measure reconstruction completeness or sensor-equivalent accuracy. Nearest-neighbor surface distance differs from viewing-ray depth error and includes residual global alignment error and LiDAR sampling effects. Semantic filtering reduces boundary mismatches but does not eliminate occlusions or distinguish different surfaces within the same class.

\begin{figure*}[!tp]
    \centering
    \includegraphics[width=\linewidth]{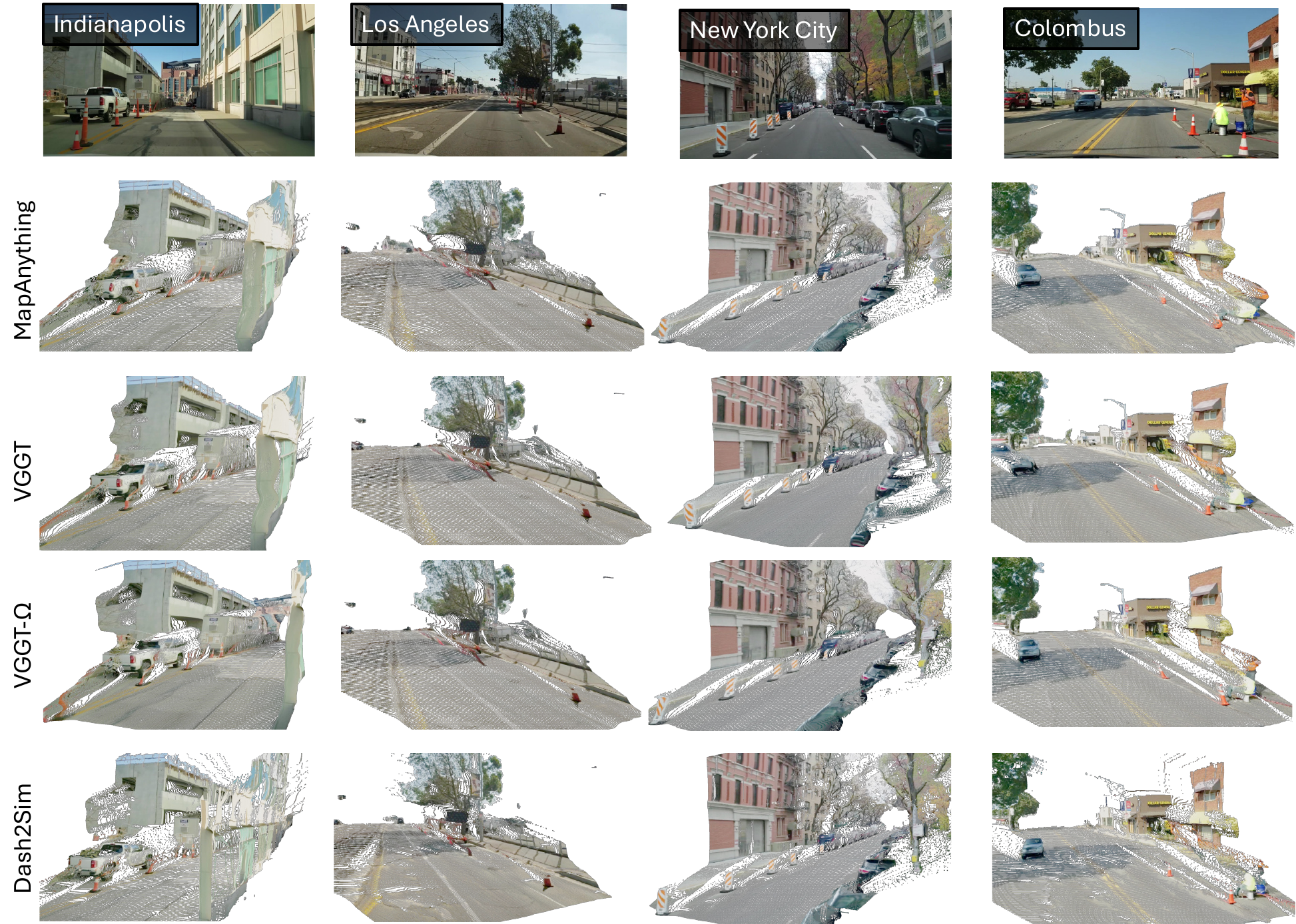}
    \caption{\textbf{Qualitative comparison of recovered point clouds.} Top row: dashcam frames from Indianapolis, Los Angeles, New York City, and Columbus. Rows below: point clouds from MapAnything~\cite{keethamapanything}, VGGT~\cite{wang2025vggt}, VGGT-$\Omega$~\cite{vggtomega}, and \projectname{}. Examples show road surfaces, vehicles, and small work-zone objects such as cones and vertical panels.}
    \label{fig:qualitative-pointclouds}
\end{figure*}

\paragraph{Dense Depth Agreement with Sparse Geometry.}
\label{app:depth-agreement}
We additionally evaluate dense depth on the 512-scene \benchmarkname{} test set to compare methods on work-zone dashcam videos. These videos lack measured depth, so we use their recovered sparse SfM geometry as a reference. We compare depth from \projectname{}, VGGT~\cite{wang2025vggt}, VGGT-$\Omega$~\cite{vggtomega}, and MapAnything~\cite{keethamapanything}, excluding sky regions and aligning scale per sequence. For $M$ valid depth pairs, let $d_j$ and $z_j$ denote the aligned predicted and reference depths. We report absolute relative error (AbsRel) and root mean squared error (RMSE),
\begin{align*}
    \mathrm{AbsRel} &= \frac{1}{M}\sum_{j=1}^{M}\frac{|d_j-z_j|}{z_j}, \\
    \mathrm{RMSE} &= \sqrt{\frac{1}{M}\sum_{j=1}^{M}(d_j-z_j)^2}.
\end{align*}
AbsRel is dimensionless and RMSE is in meters. We also report the percentage of depth ratios below $1.25^t$,
\[
    \delta_t = \frac{100}{M}\sum_{j=1}^{M}\mathbf{1}\!\left[\max\!\left(\frac{d_j}{z_j},\frac{z_j}{d_j}\right)<1.25^t\right],
\]
where $\mathbf{1}[\cdot]$ is the indicator function. Lower AbsRel and RMSE and higher $\delta_t$ indicate better agreement.

Prior depth evaluations, including Metric3D v2~\cite{hu2024metric3d}, use $t\in\{1,2,3\}$. VGGT-$\Omega$~\cite{vggtomega} uses $\delta_1$ but denotes it as $\delta_{1.25}$. These thresholds are less sensitive to the remaining geometric errors as depth estimates improve. We therefore also report $\delta_{0.5}$ and $\delta_{0.25}$, which require depth ratios below $1.25^{0.5}\approx1.118$ and $1.25^{0.25}\approx1.057$, respectively. In \Tabref{colmap_agreement}, the gap between \projectname{} and VGGT-$\Omega$ increases from 5.02 percentage points at $\delta_1$ to 15.53 at $\delta_{0.5}$ and 28.00 at $\delta_{0.25}$. The tighter thresholds reveal differences that are less apparent at $\delta_1$. We show qualitative point-cloud comparisons in \figref{qualitative-pointclouds}.

\begin{figure}[tbp]
    \centering
    \includegraphics[width=\linewidth]{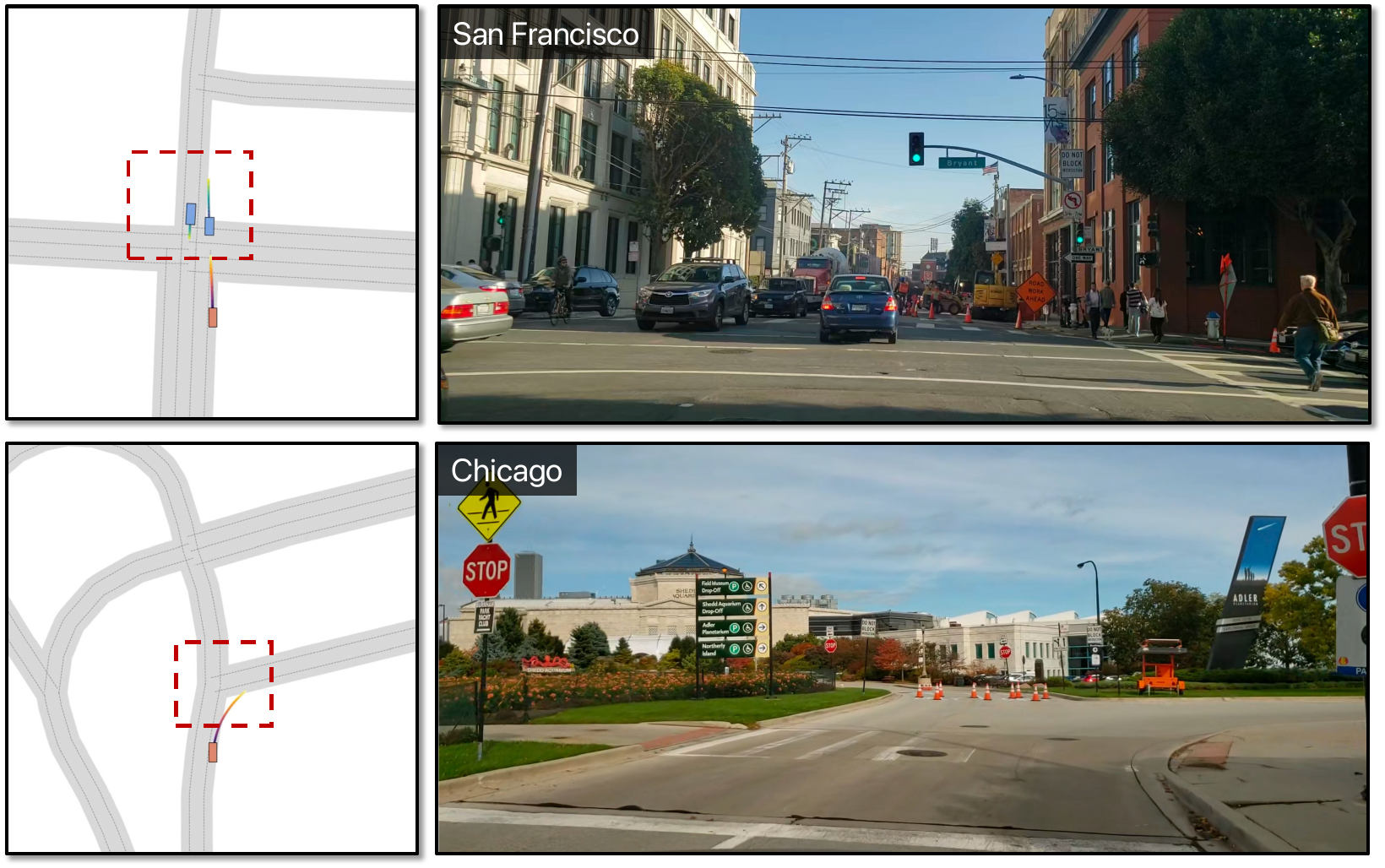}
    \caption{\textbf{Examples of Mismatch and Failures with SD Maps.} Cases where the video and the SD map disagree. \textbf{Top:} A lane-count mismatch: OSM shows 2 lanes, but the video shows 4 lanes with construction on some of them (the map may predate a transition of that street to a narrower configuration). Since OSM is crowd-sourced and updated nightly, it is difficult to retrieve a map that matches the state of the road at the time of video capture. \textbf{Bottom:} A missing slip lane: OSM models the junction as two centerlines meeting at a node, omitting the curved connector and its traffic island, so trajectories following the actual curb appear to leave the drivable area.}
    \label{fig:osm-failures}
\end{figure}

\noindent\textbf{Limitations in Reconstruction and Map Sources.} Reporting trajectory error after local rigid-similarity alignment is the standard benchmarking choice~\cite{kitti,grupp2017evo,sturm2012benchmark} (used in Table~\ref{tab:pose-eval}), but these metrics do not capture many different types of errors and limitations, for example, in global coordinate alignment across different sources. Recent work on city-scale visual localization~\cite{krishnan2025lamaria} confirms this limitation\footnote{\url{https://www.youtube.com/watch?v=gLoMiIcCiCk&t=1098s}; Quote: \textit{We're essentially within a centimeter based on those ground control points... But Google Earth isn't [as accurate]. Google Earth is actually several meters off in many places.}}, showing that consumer-grade GNSS and associated products (Google Street View, Virtual 3D Assets in Google Earth, etc.) have minor errors in dense urban environments due to many potential factors~\cite{spilker1996global, klingner2013street, hsu2018analysis}.

Our anchors derive from Street View panorama coordinates, which are not survey-grade. We measured the residual absolute-frame errors between our reconstructions and nuScenes' map-anchored ground truth at $3$ to $9$~meters horizontally, among other sources of error. Any system anchored to Street View or consumer GNSS inherits this class of errors. ATE does not measure this global alignment error because it is computed after alignment to ground truth, while RPE measures relative motion. We indirectly measure it because physically grounded verification reacts to its consequence: a globally misaligned log fails drivable-area compliance against the independently maintained map (\secref{simulation}).

Our map source itself (OpenStreetMap) has further limitations, owing to its Standard Definition (SD) quality compared to the true HD maps provided with fleet-collected benchmarks. For example, being crowd-sourced, OpenStreetMap does not always provide lane width, nor does it provide slip-lane or curb information at turns (i.e., many streets appear to intersect as road lines, with no drivable area around curbs), which is a problem for tight turns. We partly mitigate this during closed-loop evaluation by being more lenient when computing what constitutes a drivable area violation, but this is an unavoidable problem with SD Maps. Moreover, prior work~\cite{sriram2020smart, cai2020summit} has shown that studying autonomous driving with these limitations of SD Maps is still valuable.

We partially address some of these issues via non-reactive log-replay simulation (\secref{simulation}), which selects 4D driving logs for closed-loop evaluation based on consistency with the simulation setup. Other approaches operate without a global map: they either operate in a local ENU frame~\cite{wod_e2e, lu2026onevl} or build fully synthetic simulations~\cite{cai2020summit} from real logs. Addressing these issues would expand the benchmark for privileged closed-loop planning~\cite{nuplan}, but we leave them outside the scope of this work.


\subsection{Retrieving Street Imagery Sans Source GPS}
\label{app:vpr}

\noindent\textbf{Overview.} \projectname{} uses coarse GPS to retrieve nearby street view imagery, an assumption that holds for most dashcam data. To test whether GPS can be removed entirely, we evaluate visual place recognition as an alternative.

\noindent\textbf{Setup.} We take the 227 \emph{annotated} San Francisco images
in the ROADWork~\cite{ghosh2025roadwork} dataset as queries and match them with
MegaLoc~\cite{berton2025megaloc} embeddings against $\approx$106k
Mapillary~\cite{mapillary} street-level references covering
$\approx$20\,km² of the city.

\noindent\textbf{Findings.} \Tabref{vpr-recall} shows that 60\% of queries
retrieve a reference within 50\,m of ground truth at top-1, and that,
conditioned on coverage, MegaLoc~\cite{berton2025megaloc} reaches 76\% at top-1 and 85\% at top-5, well within what \projectname{} needs as a spatial location prior.
Most of the unconditional error therefore comes from sparse Mapillary
coverage rather than the retrieval model. Note that R@$\infty$ is a
purely spatial bound: a reference within $d$ meters need not visually
match the query, since viewing direction, time of day, weather, and
occlusions can all differ between dashcam captures and crowdsourced
Mapillary images; thus, achievable recall is likely below R@$\infty$. For video queries, redundancy across nearby frames opens an orthogonal opportunity to filter and reweight street view retrievals across the sequence~\cite{torii2011visual,ghosh2016dynamic}.

\begin{figure}[tbp]
    \centering
    \includegraphics[width=\linewidth]{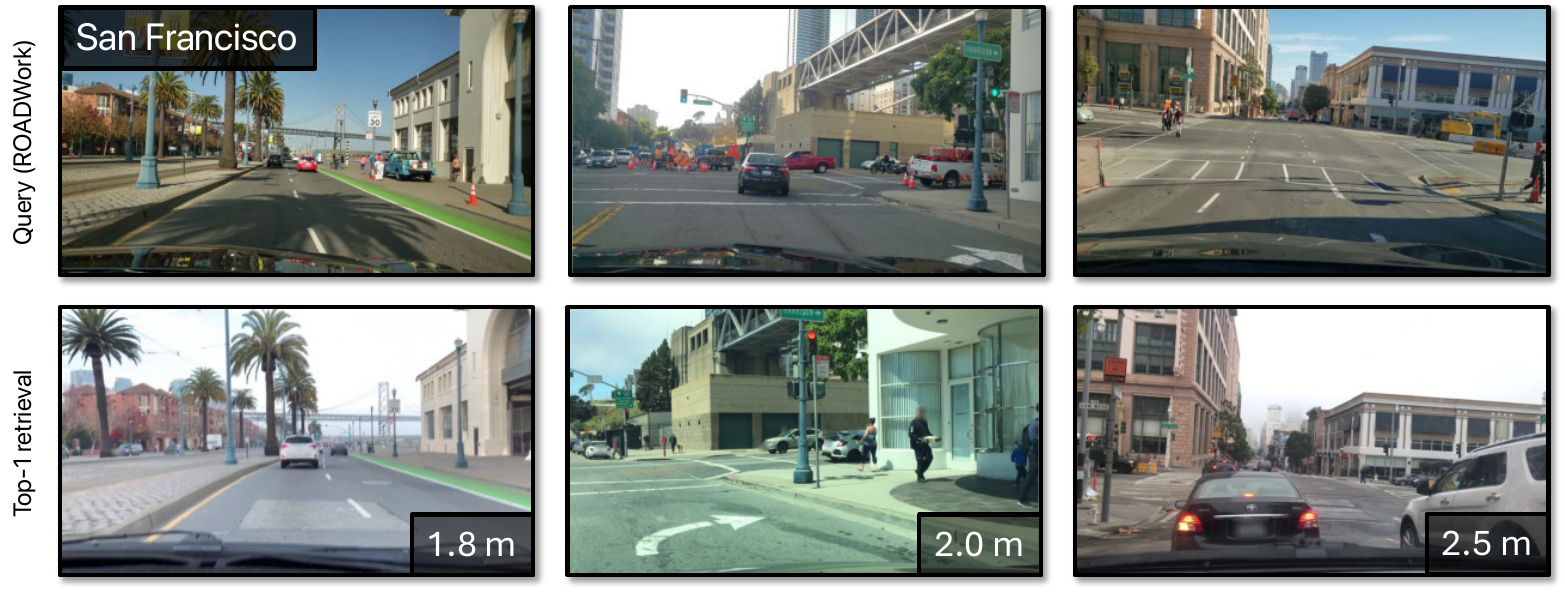}
\caption{\textbf{Street View Imagery recovery via Visual Place Recognition.} Top-1 retrievals from a Mapillary~\cite{mapillary} reference database for ROADWork~\cite{ghosh2025roadwork} queries, with haversine distance between the query's ground-truth GPS and the retrieved image's GPS. Despite significant differences, retrieved images localize within a few meters of ground truth given coverage.}
    \label{fig:vpr-retrievals}
\end{figure}

\begin{table}[t]
\centering
\caption{\textbf{Street View Imagery recovery via Visual Place Recognition.} 227 San Francisco
ROADWork~\cite{ghosh2025roadwork} queries against ${\sim}$106k
Mapillary~\cite{mapillary} references ($\approx$20\,km²), retrieved with
MegaLoc~\cite{berton2025megaloc}. R@$k$ is the fraction of queries with a
top-$k$ retrieval within $d$ meters of ground truth; R@$\infty$ is the
coverage ceiling (any reference within $d$). Bracketed percentage scores are
R@$k$/R@$\infty$, i.e. the share of geometrically solvable queries correctly
retrieved.}
\label{tab:vpr-recall}
\resizebox{\linewidth}{!}{%
\begin{tabular}{lccc}
  \toprule
   & $d = 25\,\text{m}$ & $d = 50\,\text{m}$ & $d = 100\,\text{m}$ \\
  \midrule
  R@1                            & 0.44 (65\%) & 0.60 (76\%) & 0.71 (77\%) \\
  R@5                            & 0.50 (74\%) & 0.67 (85\%) & 0.76 (83\%) \\
  \midrule
  \rowcolor{tablehighlight}
  Coverage (R@$\infty$)  & 0.68        & 0.79        & 0.92        \\
  \bottomrule
\end{tabular}}
\end{table}

\Figref{vpr-retrievals} shows a handful of retrievals. MegaLoc~\cite{berton2025megaloc} retrieves nearby references despite substantial appearance variation between query and reference captures: different traffic, lighting, time of year, occlusions, and presence of long-tail objects such as work vehicles or blocked-off streets. We present this simple method as proof-of-concept for applying VPR at scale. We use Mapillary because it is openly queryable and allows retrieving city-scale imagery. Street View coverage is denser and more uniform, so we expect comparable or better retrieval. For further improvements in accuracy and size of the reference set, hybrid methods are applicable~\cite{lindenberger2025scaling}.

\subsection{Detecting and Lifting Objects into 3D}
\label{app:detection-tracking}

\begin{algorithm*}[!tp]
\caption{Iterative Exemplar Mining for Long-Tailed Object Tracking}
\label{alg:iter-exemplar}
\small
\begin{algorithmic}[1]
\Require Video $V$; class-agnostic video propagator $\mathcal{P}$; image-level detections $\mathcal{D} = \{d_i\}$ with class $c_i$, score $s_i$, box $b_i$, mask $m_i$, frame $f_i$; long-tailed class set $\mathcal{C}$; thresholds $\tau_{\text{ex}}, \tau_{\text{co}}, \tau_{\text{iou}}$; per-class iteration budget $K_c$.
\Ensure Set of object tracks $\mathcal{T}$.
\State $\mathcal{T} \gets \emptyset$
\For{each class $c \in \mathcal{C}$}
    \State $\mathcal{A}_c \gets \{\, d \in \mathcal{D} \mid c_i = c \,\}$ \Comment{active detections of class $c$}
    \State $\mathcal{T}_c \gets \emptyset$
    \For{$k = 1, \dots, K_c$}
        \State $e \gets \arg\max_{d \in \mathcal{A}_c} s_d$ \Comment{pick highest-confidence exemplar}
        \If{$e = \varnothing$ \textbf{or} $s_e < \tau_{\text{ex}}$}
            \State \textbf{break}
        \EndIf
        \State $\mathcal{O} \gets \{\, d \in \mathcal{A}_c \setminus \{e\} \mid f_d = f_e,\; s_d \geq \tau_{\text{co}} \,\}$ \Comment{co-class detections in same frame}
        \State $\mathbf{p}_e \gets \textsc{SamplePoints}(m_e)$ \Comment{positive/negative point hints from exemplar mask}
        \State $\mathcal{Q} \gets \{(b_e, \mathbf{p}_e)\} \cup \{(b_d, \textsc{SamplePoints}(m_d))\}_{d \in \mathcal{O}}$ \Comment{multi-instance prompt set}
        \State $\mathcal{X} \gets \mathcal{P}.\textsc{Propagate}(V, f_e, \mathcal{Q})$ \Comment{tubelet $\mathcal{X} = \{(f, \text{id}, \hat{m}_{f,\text{id}})\}$}
        \If{$\mathcal{X} = \varnothing$}
            \State $\mathcal{A}_c \gets \mathcal{A}_c \setminus (\{e\} \cup \mathcal{O})$ \Comment{retire exemplar; do not retry}
            \State \textbf{continue}
        \EndIf
        \For{each $(f, \text{id}, \hat{m}) \in \mathcal{X}$} \Comment{suppress detections this tubelet explains}
            \For{each $d \in \mathcal{A}_c$ with $f_d = f$}
                \If{$\textsc{IoU}(b_d, \textsc{Bbox}(\hat{m})) > \tau_{\text{iou}}$}
                    \State $\mathcal{A}_c \gets \mathcal{A}_c \setminus \{d\}$
                \EndIf
            \EndFor
        \EndFor
        \State $\mathcal{T}_c \gets \mathcal{T}_c \cup \textsc{Reid}(\mathcal{X})$ \Comment{shift IDs for global uniqueness}
        \State $\mathcal{A}_c \gets \mathcal{A}_c \setminus (\{e\} \cup \mathcal{O})$
    \EndFor
    \State $\mathcal{T} \gets \mathcal{T} \cup \mathcal{T}_c$
\EndFor
\State \Return $\mathcal{T}$
\end{algorithmic}
\end{algorithm*}

\begin{figure*}[!tp]
    \centering
    \includegraphics[width=\linewidth]{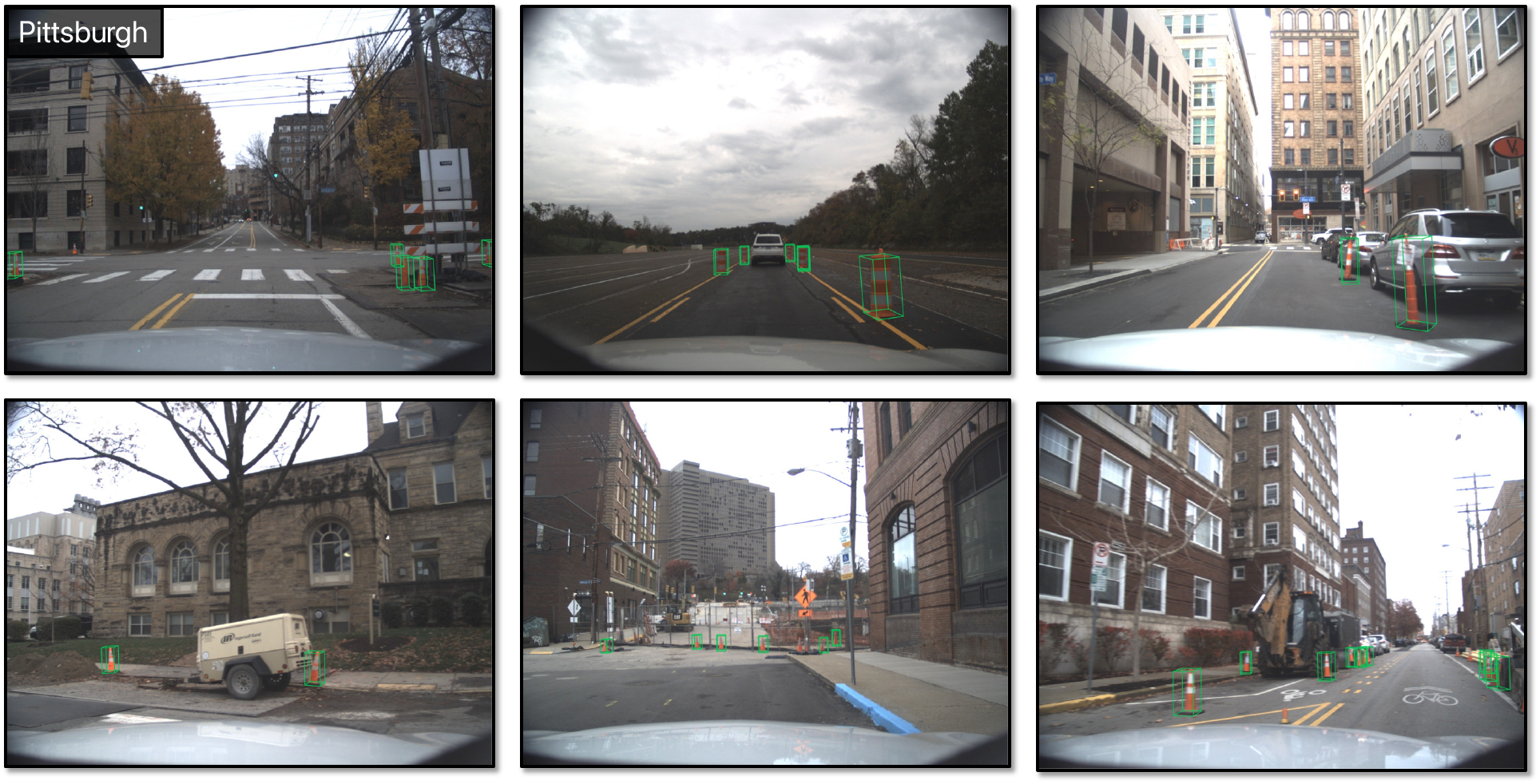}
\caption{\textbf{Qualitative 3D Lifting Results on WorkZone3D~\cite{sural2026workzone3d}.} Predicted 3D boxes from WildDet3D~\cite{huang2026wilddet3d} for long-tailed work-zone categories (cones, barrels, channelizers). Boxes are well-localized when a depth prior is available. Image-only prompting produces more misses and depth errors on small distant instances.}
\label{fig:appendix-workzone3d}
\end{figure*}

\noindent\textbf{Promptable detection and tracking.} Our observation is that the bottleneck for detecting long-tailed classes is generalization to the ``long-tailed'' text prompt, not the object tracker. SAM3~\cite{carion2025sam} accepts text and mask prompts, so we prompt with text for common objects and with exemplar masks from a domain-specific detector~\cite{ghosh2025roadwork} for long-tailed objects. A claim-and-suppress strategy then uses the detector in tandem with the tracker to recover each instance, which we detail next.

\begin{figure*}[p]
    \centering
    \includegraphics[width=\linewidth,height=0.82\textheight,keepaspectratio]{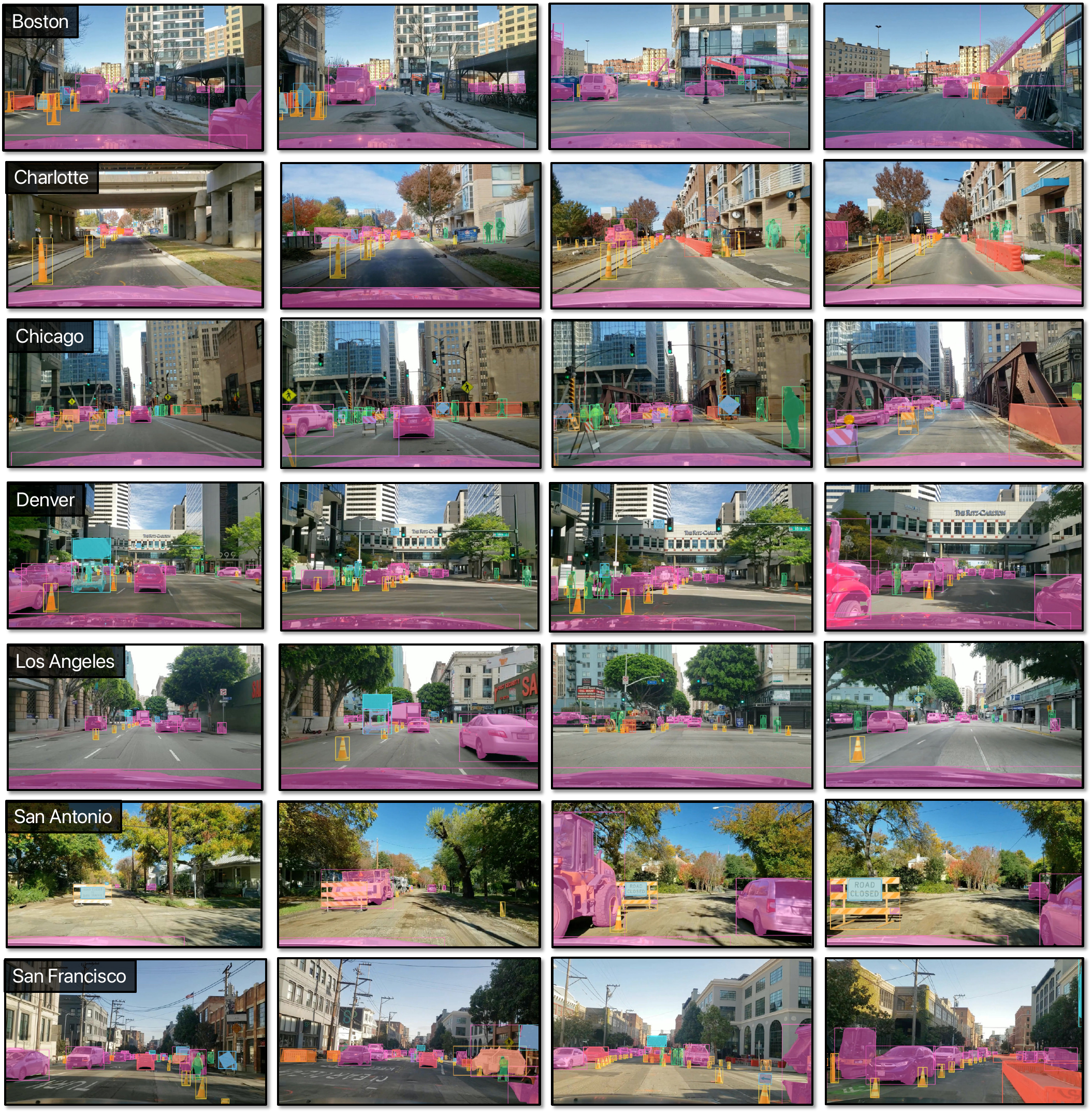}
    \caption{\textbf{2D Object Tracks recovered by Dash2Sim.} We show examples of 2D object tracks. Our approach accurately recovers both common and long-tailed objects (examples include \textit{``Arrow Board''}, \textit{``Vertical Panels''}, \textit{``Cones''}, \textit{``TTC Sign Board''}) with consistent IDs.}
    \label{fig:tracks2d-examples}
\end{figure*}

\noindent\textbf{Iterative Detection and Tracking of Long-Tailed Objects.} Segmentation Foundation Models~\cite{carion2025sam} are already comparable to domain-specific detectors when prompted appropriately~\cite{ghosh2025roadwork}. For long-tailed work-zone categories, prior work~\cite{ghosh2025roadwork} shows that SAM models recover 2D segmentations at up to 90\% relative accuracy when paired with a domain-specific detector. We adopt a strategy to efficiently prompt SAM3~\cite{carion2025sam} across all instances, composing an image-level detector~\cite{ghosh2025roadwork} in a detect-then-track loop. Tracking-by-detection via IoU-based matching is a common strategy for 2D multi-object tracking~\cite{bewley2016simple, zhang2022bytetrack}, which we adapt for our setting. Please see our recovered 2D tracks in \figref{tracks2d-examples}.

\Algoref{iter-exemplar} provides the sketch of the algorithm. For each long-tailed class $c$, we maintain a pool of active image-level detections, select the highest-confidence surviving detection as the exemplar, and prompt SAM3~\cite{carion2025sam} with its box together with positive and negative point hints sampled from the exemplar's mask. To help SAM3~\cite{carion2025sam} disambiguate the exemplar from nearby same-class instances at prompt time, other same-frame detections of class $c$ above a co-prompt confidence threshold $\tau_{\text{co}}$ are additionally passed as auxiliary prompts. SAM3~\cite{carion2025sam} propagates a tubelet across the video, and every detection whose 2D box overlaps the tubelet at any frame by IoU greater than $\tau_{\text{iou}}$ is considered ``explained'' and retired from the pool, together with the exemplar and the auxiliary detections that contributed to its prompt. The loop terminates when no detection remains above the exemplar threshold $\tau_{\text{ex}}$ or a per-class iteration budget $K_c$ is reached. In practice, a small $K_c$ suffices because a single template typically recovers most instances of its class within a scene.

\begin{figure*}[p]
    \centering
    \includegraphics[width=\linewidth,height=0.82\textheight,keepaspectratio]{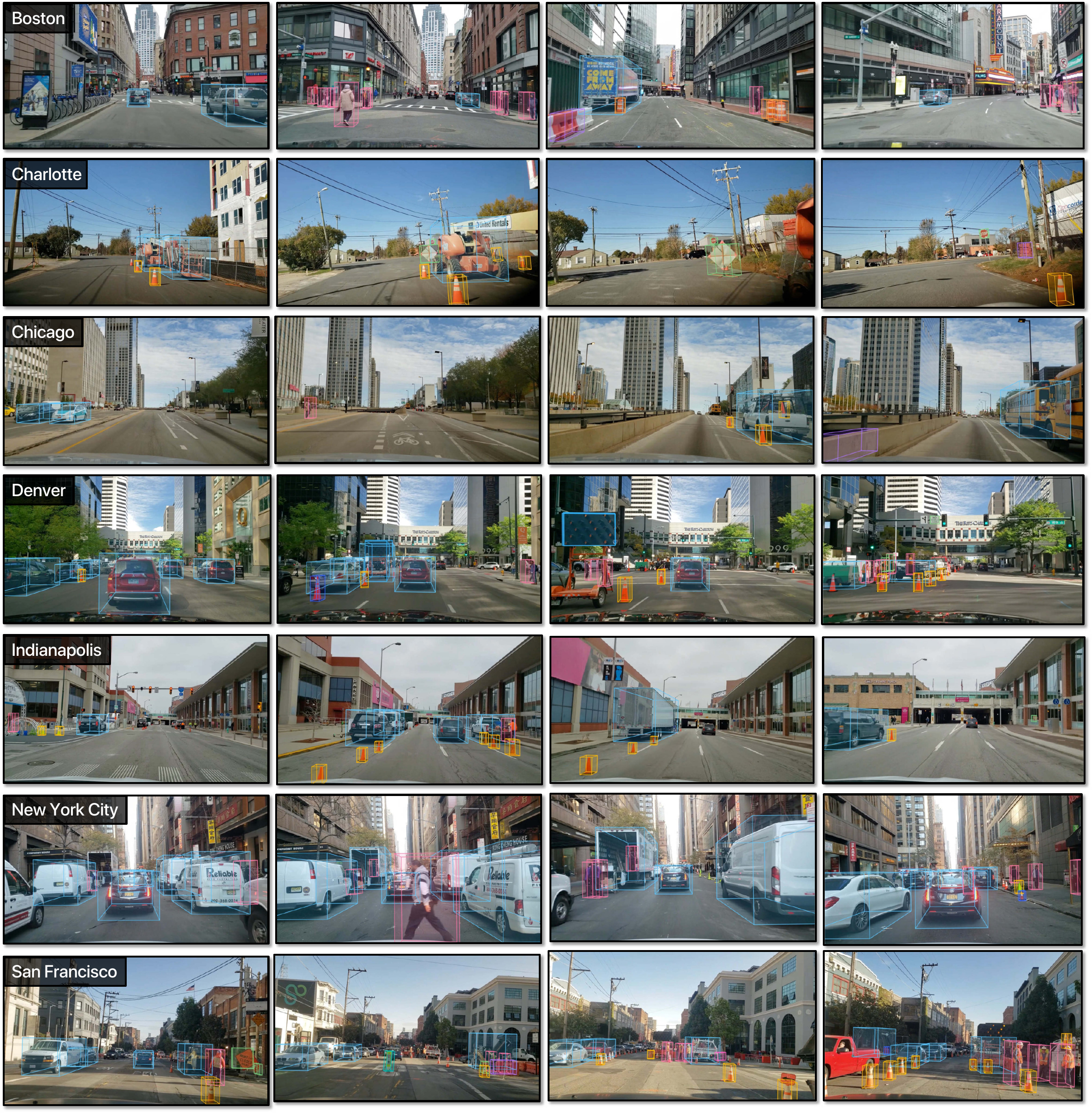}
    \caption{\textbf{3D Object Tracks recovered by Dash2Sim.} We show examples of 3D object tracks on ROADWork4D logs. Our approach recovers common and long-tailed objects when given depth supervision (examples include \textit{``Arrow Board''}, \textit{``Vertical Panels''}, \textit{``Cones''}, \textit{``TTC Sign Board''}) with consistent IDs.}
    \label{fig:tracks3d-examples}
\end{figure*}

\noindent\textbf{Depth-conditioned 4D lifting.} We observe that the per-image depth recovered in \secref{reconstruction} can act as the conditioning signal an open-vocabulary 3D detector needs for lifting small long-tailed objects. We prompt WildDet3D~\cite{huang2026wilddet3d} with 2D tracks and condition it on this depth. Following prior work that uses novel-view synthesis as a proxy for depth quality~\cite{vuong2025aerialmegadepth, tung2024megascenes}, we evaluate the recovered depth through its effect on rendering quality in \secref{nvs}. We validate 3D localization of long-tailed objects in \appref{wilddet-eval}. Please see our recovered 3D tracks in \figref{tracks3d-examples}.

\noindent\textbf{3D Track Smoothing and Class-Prior Rejection.} We smooth each lifted 3D track with an Extended Kalman Filter~\cite{smith1962application} under a constant-velocity motion model to reduce per-frame jitter in WildDet3D~\cite{huang2026wilddet3d}'s center and yaw estimates. We then reject detections that violate class-prior heuristics, namely predicted 3D dimensions outside class-specific bounds (e.g., a person is roughly 1.7 meters tall) and tracks that are too far from the fitted ground plane (\appref{geometry-blocks}).

\subsection{Evaluating Depth for 3D Object Lifting}
\label{app:wilddet-eval}

\begin{table}[t]
\centering
\tablesize
\caption{\textbf{Evaluating 3D-detection on long-tailed categories}. mAP is averaged over center-distance thresholds $\{0.5, 1, 2, 4\}$\,m. Depth conditioning gives a consistent improvement.}
\resizebox{\linewidth}{!}{%
\begin{tabular}{lrcccc}
\toprule  &  & \multicolumn{2}{c}{AP-11} &
  \multicolumn{2}{c}{AP-40} \\
\cmidrule(lr){3-4}\cmidrule(lr){5-6}
Class       & \#GT  & Img   & Img+LiDAR        & Img & Img+LiDAR \\
\midrule
Barrel      & 1{,}735 & \bestcell{0.325} & \bestcell{0.325} & \worstcell{0.293} & \bestcell{0.294} \\
Channelizer & 5{,}544 & \worstcell{0.171} & \bestcell{0.180} & \bestcell{0.127} & \bestcell{0.127} \\
Cone        & 1{,}113 & \worstcell{0.129} & \bestcell{0.130} & \worstcell{0.124} & \bestcell{0.126} \\
\midrule
Mean & ---  & \worstcell{0.208} & \bestcell{0.212} & \worstcell{0.181} & \bestcell{0.182} \\
\bottomrule
\end{tabular}}
\label{tab:workzone3d_eval}
\end{table}

\noindent\textbf{Overview.} WildDet3D~\cite{huang2026wilddet3d} is already state-of-the-art when prompted on common driving objects~\cite{argoverse}. The remaining question is whether depth conditioning improves 3D lifting of work-zone objects, which span only a few pixels. ROADWork~\cite{ghosh2025roadwork} provides only 2D annotations, so 3D detection cannot be quantified on it. We evaluate on WorkZone3D~\cite{sural2026workzone3d} instead, which provides ${\sim}$20{,}000 images with LiDAR and 3D ground truth over a similar set of long-tailed categories.

\noindent\textbf{Setup.} To isolate 3D localization accuracy, we follow a prompt-conditioned protocol that matches the lifting procedure described earlier. For every class present in a frame, a frozen 2D detector~\cite{sural2026workzone3d} produces an instance mask from which we sample point prompts and query WildDet3D~\cite{huang2026wilddet3d} in its geometric mode to return exactly one 3D box per prompt. We compare two regimes: \emph{Img} (image only) and \emph{Img+LiDAR}, where sparse LiDAR points serve as a reference depth prior. The depth recovered in \secref{reconstruction} is dense but noisy, whereas LiDAR is sparser but accurate. We report AP@11 and AP@40 averaged over center-distance thresholds $\{0.5, 1, 2, 4\}$\,m.

\noindent\textbf{Findings.} \Tabref{workzone3d_eval} shows that adding a depth prior improves mean AP@11 and AP@40, with no category-level degradation, with the largest gain on \emph{Channelizer} (AP@11: $0.171 \to 0.180$). This validates the design choice of conditioning on a depth prior when lifting long-tailed objects in \projectname{}. Qualitative results in \Figref{appendix-workzone3d} show that WildDet3D~\cite{huang2026wilddet3d} can produce well-localized boxes for the long-tailed categories when provided with depth conditioning.

\subsection{Integrating Logs into Closed-Loop Simulation}
\label{app:simulation}

Closed-loop planning needs three components on top of a 4D driving log: a simulator interface, a routable map, and a reactive-agent model. The nuPlan~\cite{nuplan} privileged-planning simulator accepts metric, map-aligned logs, provides rule-based IDM reactive agents, and lets existing planners run on \clbenchmarkname{}. For the routable map, the nuPlan dataset assumes HD maps collected by the same fleet that captured the logs, which we do not have. We retrieve an OpenStreetMap~\cite{openstreetmap} tile using the geo-referenced ego trajectory, following prior work~\cite{sriram2020smart, cai2020summit} that used OpenStreetMap lane graphs in place of HD maps. OpenStreetMap is maintained independently, so the map provides an external reference that the recovered log must be consistent with.

%


\clearpage
\section{\benchmarkname{} Benchmark Details}
\label{app:benchmark-details}

\begin{figure*}[!tp]
    \centering
    \includegraphics[width=\linewidth]{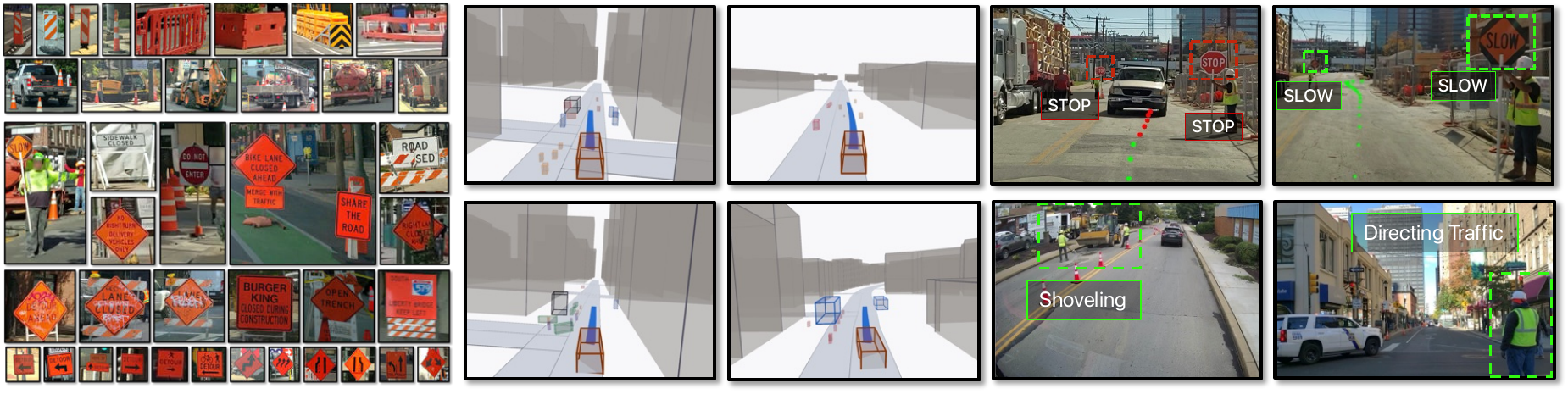}
    \captionof{figure}[Work zones involve long-tailed objects, layouts and behaviors]{\textbf{Work zones involve long-tailed objects, layouts and behaviors.} Examples along the three long-tail axes present in work zones. (Left) Rare \emph{objects} and \emph{entities}: cones, barrels, arrow boards, construction vehicles, and a wide vocabulary of temporary signage (e.g., \emph{Road Closed}, \emph{Burger King Closed During Construction}, \emph{Delivery Vehicles Only}). (Middle) Rare \emph{layouts}: temporary lane closures and detours that contradict the map, forcing the ego vehicle off expected lanes. (Right) Rare \emph{behaviors}: human flaggers holding \emph{Stop}/\emph{Slow} paddles, construction workers shoveling in active traffic, and traffic officials directing traffic with gestures. Left and right images are courtesy of the ROADWork~\cite{ghosh2025roadwork} dataset.}
    \label{fig:workzones-trifecta}
\end{figure*}
\begin{figure*}[!tp]
    \centering
    \includegraphics[width=\linewidth]{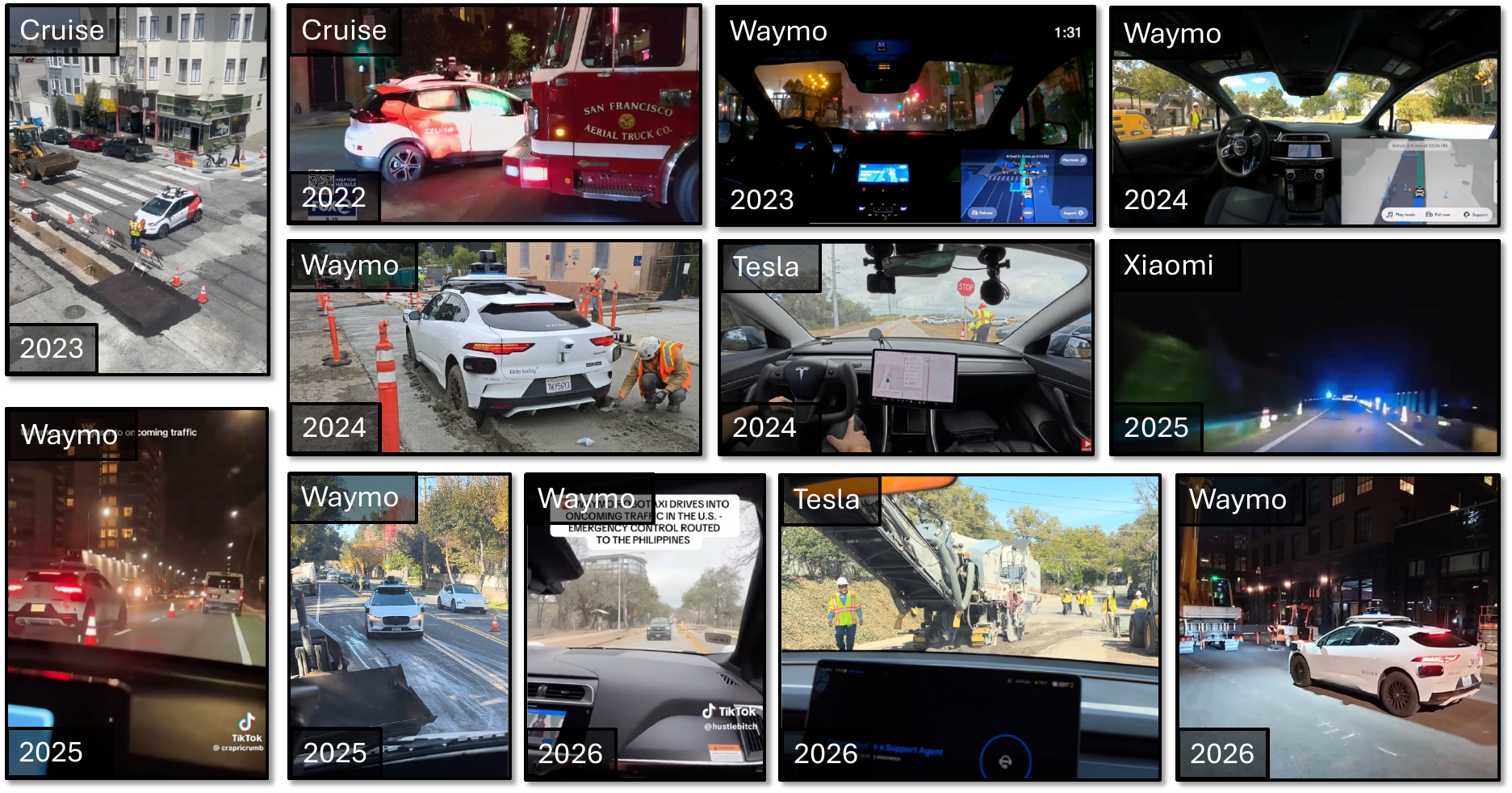}
    \captionof{figure}[Work Zones are a persistent hurdle.]{\textbf{Work Zones are a persistent hurdle.} A few work zone failures by commercial autonomous driving and driver-assisted vehicles, drawn from publicly available information (news reports, social media videos, etc.) between 2022 and 2026. Work zones remain a challenge for reliable self-driving. Failure modes include colliding with emergency vehicles, driving into wet concrete, not stopping for human flaggers, getting boxed in by construction equipment, merging into oncoming traffic, and fatal accidents that prompted regulatory action~\cite{reuters2025chinaadban}. Recent incidents~\cite{techcrunch2026waymoworkzone, cnbc2026waymorecall} suggest that solving autonomous driving in work zones is a current, practical, and critical challenge. \textit{Do note the visualization is not exhaustive or necessarily representative of the abilities of the commercial systems, as some commercial operators have a much larger fleet compared to others and thus a larger error surface area.}}
    \label{fig:workzone-fails}
\end{figure*}

\subsection{Long-tail Driving}
\label{app:longtail-definition}

Long-tail driving refers to scenarios that occur rarely but determine system safety and deployment readiness. Within a fixed operational design domain (ODD), the scenario distribution is heavy-tailed: each additional fleet hour draws predominantly from common scenarios, making rare events increasingly expensive to collect.

Rare cases span multiple axes that often co-occur~\cite{ghosh2025roadwork}: rare objects (debris, animals), rare layouts (temporary lane closures, washed-out lane markings), rare behaviors (vehicles driving on the wrong side of the road, ``illegal''-but-correct maneuvers), and rare environmental conditions (snow, fog, rain, glare). Prior work generally targets each axis in isolation, for instance through dedicated adverse-weather collection~\cite{sakaridis2021acdc, bijelic2020seeing}.

\subsection{Work Zones as a Testbed}
\label{app:workzones}

Work zones as a category span three long-tail axes. A typical work zone may contain rare objects (cones, barrels, arrow boards, and a vocabulary of temporary signage defined by federal standards~\cite{fhwa_mutcd_2026}), rare layouts (lane closures that contradict the static map, contraflow setups, contradictory markings), and rare behaviors (flaggers holding stop and slow paddles, workers in active traffic, officers directing traffic with gestures). \figref{workzones-trifecta} shows examples along each axis.

Work zones are a significant source of traffic fatalities: NHTSA reports 898 fatalities in U.S. construction or maintenance zones in 2023 alone~\cite{nhtsa2026farsworkzone}. They remain a persistent challenge for commercial self-driving systems. \figref{workzone-fails} compiles publicly documented failures from commercial autonomous vehicles between 2022 and 2026, ranging from collisions and merges into oncoming traffic to regulatory action following fatalities~\cite{reuters2025chinaadban}, operational halts~\cite{techcrunch2026waymoworkzone, cnbc2026waymorecall}. These failures span commercial platforms across years.

Work zones are also tractable to study. They occur frequently in routine driving, appear in disengagement and crash filings as a measurable category, and the ROADWork~\cite{ghosh2025roadwork} dashcam corpus provides large-scale 2D annotations for work-zone perception, complementing \benchmarkname{}. Overall, \benchmarkname{} offers an additional setting for long-tail 4D perception and planning that complements existing autonomous driving benchmarks.

\subsection{Dashcams as a complementary data source}
\label{app:why-dashcams}

Fleet data covers the head of the scenario distribution well. Instrumented vehicles carry high-quality sensors but operate in a small number of cities, and each additional fleet hour is increasingly unlikely to encounter a novel rare event. Curating long-tail subsets from fleet logs~\cite{wod_e2e, wagner2026longtail} helps, but can only surface what the fleet already drove through. Synthetic simulators~\cite{carla, bench2drive, fail2drive} can construct rare scenarios by hand, but the range of rare objects, layouts, and behaviors they produce is determined by the available assets and scenario definitions, and both visual and behavioral sim-to-real gaps remain.

Dashcam video covers a different part of the distribution. Consumer dashcams are owned by an estimated 30\% of U.S. drivers~\cite{autoinsurance2026dashcams}, span far more cities and road conditions than any single fleet, and civilian uploads tend to over-represent unusual driving events. The cost is noisier scene reconstruction: monocular video lacks the multi-modal sensing of fleet vehicles. \projectname{} works within this constraint to produce metric 4D logs that complement fleet-derived benchmarks for studying the long tail of autonomous driving. \figref{city-distribution} shows the geographic distribution of \benchmarkname{} scenarios.

\benchmarkname{} spans 17 US cities, exceeding existing fleet-derived benchmarks (nuPlan covers 4 cities, nuScenes 2, Argoverse 6). Many of these cities, including Chicago, Columbus, Indianapolis, Jacksonville, and Charlotte, are not represented in any existing planning benchmark and do not yet have a commercial autonomous ride-hailing service collecting fleet data. This geographic breadth matters because, although U.S. regulations~\cite{fhwa_mutcd_2026} establish national minimum standards for temporary traffic control, work zone regulations are implemented and supplemented at the state and local level~\cite{usdot2024mutcdadoption, usdot2024workzonesafety}, inducing domain gaps in both perception and planning for autonomous driving~\cite{ghosh2025roadwork}. The closed-loop and end-to-end benchmarks have similar city distributions. The end-to-end protocol uses the full 4{,}244-scene corpus, with 3{,}732 scenes for training and 512 for held-out evaluation (\secref{e2e-planning}). \appref{benchmark-comparison} compares \benchmarkname{} against existing benchmarks.

\subsection{Existing benchmarks and simulators}
\label{app:benchmark-comparison}

\paragraph{Column definitions.} \textbf{Source}: data origin (\emph{Fleet} for calibrated autonomous-vehicle fleets, \emph{Synthetic} for procedural generation in CARLA or its derivatives, \emph{Dashcam} for in-the-wild consumer-grade video sources). \textbf{Long-Tail}: whether the benchmark is explicitly curated for rare scenarios. \textbf{Locales}: distinct cities or synthetic towns. \textbf{Maps}: whether routable map data is provided. \textbf{Hours}/\textbf{Scenes}: total recorded driving time and number of scenes, segments, or routes (unit varies by benchmark and is clarified per row below). \textbf{Closed-Loop}: whether a policy's actions change future observations, as opposed to scoring a predicted trajectory against a fixed log. \textbf{Reactive}: whether non-ego agents respond to the ego vehicle's actions rather than replaying logged trajectories.

\paragraph{Comments on planning benchmarks.}

\textbf{nuScenes}~\cite{nuscenes} provides 1{,}000 20-second scenes (5.5 hours annotated). \textbf{Argoverse 2}~\cite{argoverse2} provides 1{,}000 15-second sensor sequences (4.2 hours), which we count in \Tabref{benchmarks}. Its 250{,}000 motion-forecasting scenarios form a separate collection. Many multi-modal perception datasets can be repurposed as planning benchmarks~\cite{sriram2020smart}.

\textbf{WOMD}~\cite{waymo_motion} contains $\sim$104{,}000 20-second segments. \textbf{WOD-E2E}~\cite{wod_e2e} curates 4{,}021 long-tail segments at an occurrence frequency less than 0.03\%. \textbf{Navsim}~\cite{navsim} builds on OpenScene~\cite{openscene2023, sima2023occnet} and nuPlan~\cite{nuplan} data, with 12{,}000 evaluation samples and a short-horizon non-reactive rollout that we count as pseudo closed-loop. \textbf{Navsim V2}~\cite{cao2025pseudo} uses the same 103k training scenes and extends this with sensor-grounded pseudo closed-loop evaluation via 3D Gaussian Splatting on 450 source scenes. \textbf{nuPlan}~\cite{nuplan} provides an open-source closed-loop reactive simulator, with IDM-based reactive agents for privileged planning evaluation. Learned agent extensions~\cite{hagedorn2025planners} to nuPlan~\cite{nuplan} have been proposed but are not yet open-sourced.

\textbf{InterPlan}~\cite{interplan} augments nuPlan logs by spawning additional traffic agents and modifying routes to force lane changes. The maps and ego logs are realistic (inherited from nuPlan), but the added interactions are synthetic, so we group it with synthetic benchmarks. The full release contains 335 scenarios. \textbf{Bench2Drive}~\cite{bench2drive} provides 220 evaluation routes spanning 44 interactive scenarios. \textbf{Fail2Drive}~\cite{fail2drive} provides 200 paired routes across 17 scenario classes.


\section{Extended Limitations and Future Work}
\label{app:limitations}

\noindent\textbf{Scope of the Work.} We show that dashcam videos are a useful data source for end-to-end learning and privileged closed-loop simulation, and provide a framework that recovers 4D logs from such videos at scale. Numerous adjacent opportunities remain, such as building a truly closed-loop sensor simulation from dashcam videos, which we leave as future work, having shown the potential of the \projectname{} framework and the corresponding \benchmarkname{} benchmark for such tasks. We hope our work encourages the broader community to make progress on the difficult, ongoing challenge of long-tailed driving.

For privileged closed-loop evaluation, the \clbenchmarkname{} benchmark can be extended in several useful ways for better planning evaluation in long-tailed situations: \textit{(a)} We do not recover traffic light or signaling information, which would be useful for more realistic simulation. \textit{(b)} Considered planners do not exploit the ``open-set'' nature of our data which may yield additional benefits, i.e. we map all the object categories to existing nuPlan categories~\cite{nuplan} for compatibility with existing planners. \textit{(c)} The dynamics of the recovered agent tracks could be improved beyond an EKF~\cite{smith1962application} with a constant-velocity motion model, which rarely holds in the real world (see \appref{detection-tracking}). \textit{(d)} We use IDM-based reactive agents provided by the simulator. These agents follow road-centerline paths and respond to the ego vehicle with car-following dynamics~\cite{treiber2000idm}. They do not capture complex interactive behaviors such as yielding negotiations, aggressive merges, or flagger compliance. Incorporating learned traffic models~\cite{hagedorn2025planners} would further improve fidelity of closed-loop evaluation. \textit{(e)} Additionally, our verification mechanism can be employed w.r.t.\ each recovered agent by treating that agent as the ego agent, but we omit that step for simplicity.

\noindent\textbf{Dashcam video captures a single viewpoint.} Fleet vehicles carry surround-view LiDAR and multiple cameras. A single dashcam observes a forward-facing view and cannot match that density. This sensing modality leaves many regions unobserved, introducing an implicit domain gap even for perfectly reconstructed logs. Sensor simulation methods can potentially synthesize plausible observations for unobserved viewpoints (see \figref{multicam-initial} for some initial results). \projectname{} provides the 4D driving logs that would act as \textit{physically grounded} conditioning for such methods.

\noindent\textbf{Noise in \benchmarkname{}.} Our framework, \projectname{}, composes monocular reconstruction, detection, tracking, and 3D-object lifting. Each stage contributes to downstream error. We quantify component errors wherever reference data are available and discuss mitigation strategies wherever possible (see \appref{framework-details}). A few limitations are beyond our scope, for example, extending \projectname{} beyond the daytime, good-weather captures in the ROADWork~\cite{ghosh2025roadwork} dashcam dataset, since bad weather is a known failure mode for many of the underlying methods in the \projectname{} framework. We also expect that advances in 3D and 4D reconstruction~\cite{pan2024global, pan2026gluemap, wang2025vggt, keethamapanything} and vision foundation models~\cite{carion2025sam} will directly improve recovered log quality without architectural changes to the framework. Some error sources, notably mismatched or incomplete map topology (see \figref{osm-failures}), lie outside \projectname{} and are difficult to correct automatically.

\noindent\textbf{Scale and Scope of \benchmarkname{}.} We focus our demonstrations of \projectname{} on a broad class of long-tailed situations, work zones. Navigating work zones is a persistent hurdle for commercial operators (\figref{workzone-fails}), and we believe this focus is a practical way to study a problem that commercial autonomous vehicles currently face. \benchmarkname{} complements existing long-tailed driving benchmarks~\cite{wod_e2e, bench2drive, fail2drive, wagner2026longtail}. We expect that further scaling with dashcam videos would offer additional insights and research directions for the field. Our work shows that in-the-wild data sources can complement expensive curated datasets.

\noindent\textbf{Correlation with real-world deployment.} Our evaluation shows that planners exhibit measurable performance differences on \benchmarkname{} scenarios relative to fleet-derived benchmarks, but whether these differences predict on-road deployment outcomes remains an open problem. Validation frameworks that augment real-world testing with simulation~\cite{luo2025sim2val} are necessary for bridging this gap.

\ifARXIV\else
\clearpage
\ifSUBMISSION
  \deferredwrites
  \linenumbers
  \setcounter{page}{\value{bodylastpage}}
  \DISCARDPAGESfalse
\fi
\ifSUPPLEMENTAL\else
  \paragraph{LLM Usage Statement}
We have used AI assistance from OpenAI Codex and Claude Code for writing code, running experiments, and polishing the text in the manuscript. We have reviewed all the outputs from these LLMs (including any code artifacts and experimental scripts), and have manually reviewed and referenced the text, references, code and the associated data. We take full responsibility for all results and content included in the manuscript.

\fi
{\small\bibliography{refs}}
\fi

\end{document}